\documentclass[10pt,twocolumn,letterpaper]{article}
\makeatletter
\let\@fnsymbol\@arabic
\makeatother
\usepackage{cvpr}
\usepackage{times}
\usepackage{epsfig}
\usepackage{graphicx}
\usepackage{amsmath}
\usepackage{amssymb}
\usepackage{subcaption}
\usepackage{booktabs}
\usepackage[outline]{contour}
\usepackage[shortlabels]{enumitem}

\newcommand{\M}[1]{\mathtt{#1}}
\newcommand{\V}[1]{\M{#1}}

\newcommand{\arr}[2]{\begin{array}{#1} #2\end{array}}
\newcommand{\mat}[2]{\left[\!\!\arr{#1}{#2}\!\!\right]}

\newcommand{\norm}[1]{\left\lVert#1\right\rVert}

\cvprfinalcopy % *** Uncomment this line for the final submission

\def\cvprPaperID{6849} % *** Enter the CVPR Paper ID here

\def\squad{\hskip0.15em\relax}

\ifcvprfinal\fi
\begin{document}
\setcounter{footnote}{1}
%%%%%%%%% TITLE
\title{From two rolling shutters to one global shutter}
\author{Cenek Albl$^\textrm{1}$ $\squad$
Zuzana Kukelova$^\textrm{2}$ $\squad$ 
Viktor Larsson$^\textrm{1}$ $\squad$
Michal Polic$^\textrm{3}$ $\squad$
Tomas Pajdla$^\textrm{3}$ $\squad$
Konrad Schindler$^\textrm{1}$  
\and
$^\textrm{1}$ETH Zurich$\qquad\qquad$
%$^\textrm{2}$
\thanks{FEE - Faculty of Electrical Engineering, Czech Technical University in Prague, $^3$CIIRC - Czech Institute of Informatics, Robotics and Cybernetics, Czech Technical University in Prague. TP and MP were supported by the European Regional Development Fund (IMPACT CZ.02.1.01/0.0/0.0/15 003/0000468) and Horizon 2020 projects 856994 and 871245. ZK was supported by OP RDE project International Mobility of Researchers MSCA-IF at CTU Reg. No. CZ.02.2.69/0.0/0.0/$17\_050$/0008025 and OP VVV project Research Center for Informatics Reg. No. CZ.02.1.01/0.0/0.0/$16\_019$/0000765. VL was supported by an ETH Postdoctoral Fellowship. We thank Alexander\ Wolf for building the camera rig and Nico Lang for help with data acquisition.}~\,VRG, FEE, CTU in Prague. $\qquad\qquad$
$^\textrm{3}$CIIRC, CTU in Prague\\
}
\maketitle
\thispagestyle{empty}

%%%%%%%%% ABSTRACT
\begin{abstract}
\noindent
 Most consumer cameras are equipped with electronic rolling shutter, leading to image distortions when the camera moves during image capture. We explore a surprisingly simple camera configuration that makes it possible to undo the rolling shutter distortion: two cameras mounted to have different rolling shutter directions. Such a setup is easy and cheap to build and it possesses the geometric constraints needed to correct rolling shutter distortion using only a sparse set of point correspondences between the two images. We derive equations that describe the underlying geometry for general and special motions and present an efficient method for finding their solutions. Our synthetic and real experiments demonstrate that our approach is able to remove large rolling shutter distortions of all types without relying on any specific scene structure.
\end{abstract}
\vspace{-1.2em}
\section{Introduction}
\noindent Thanks to low price, superior resolution and higher frame rate, CMOS cameras equipped with rolling shutter (RS) dominate the market for consumer cameras, smartphones, and many other applications.
Unlike global shutter (GS) cameras, RS cameras read out the sensor line by line~\cite{meingast_geometric_2005}. Every image line is captured at a different time, causing distortions when the camera  moves during the image capture. The distorted images not only look unnatural, but are also unsuitable for conventional vision algorithms developed for synchronous perspective projection~\cite{hedborg_rolling_2012,albl_rolling_2019,saurer_rolling_2013}. 

There are two main approaches to remove RS distortion. The first is to estimate the distortion and remove it, i.e., synthesize an image with global shutter geometry that can be fed to standard vision algorithms~\cite{grundmann_calibration-free_2012,rengarajan_bows_2016,rengarajan_unrolling_2017,vasu_occlusion-aware_2018,lao_robust_2018}. The second approach is to keep the original images and adapt the algorithms to include the RS in the camera model~\cite{hedborg_structure_2011,hedborg_rolling_2012,albl_rolling_2019,saurer_rolling_2013,dai_rolling_2016}. 
The latter approach has recently lead to RS-aware algorithms for many parts of the 3D vision pipeline, including RS camera calibration~\cite{oth_rolling_2013}, RS structure-from-motion reconstruction~\cite{hedborg_rolling_2012}, dense multi-view RS stereo~\cite{saurer_rolling_2013}, and RS absolute camera pose~\cite{ait-aider_simultaneous_2006,magerand_global_2012,albl_rolling_2019,albl_rolling_2016,saurer_minimal_2015,albl_linear_2018}. Two-view geometry of RS cameras has been studied in~\cite{dai_rolling_2016} and triangulation with a RS stereo rig is discussed in~\cite{ait-aider_structure_2009}.

\begin{figure}[t]
\centering
\renewcommand{\tabcolsep}{0.1pt}
    \begin{subfigure}[b]{.26\columnwidth}
        \centering
        \includegraphics[width=\columnwidth]{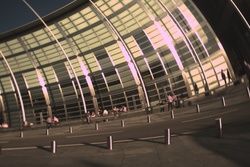} \\
        \includegraphics[width=\columnwidth,angle=-180,origin=c]{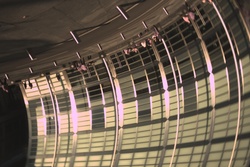} 
    \end{subfigure}
    \begin{subfigure}[c]{.15\columnwidth}
    \centering
    \raisebox{8.9em}{\footnotesize
    \begin{tabular}{c}
    $\boldsymbol\omega$ solver\\\contour{black}{$\Longrightarrow$}\\
    1.5 point\\
    matches
    \end{tabular}
    }
    \end{subfigure}
    \begin{subfigure}[b]{.525\columnwidth}
        \centering
        \includegraphics[width=\columnwidth]{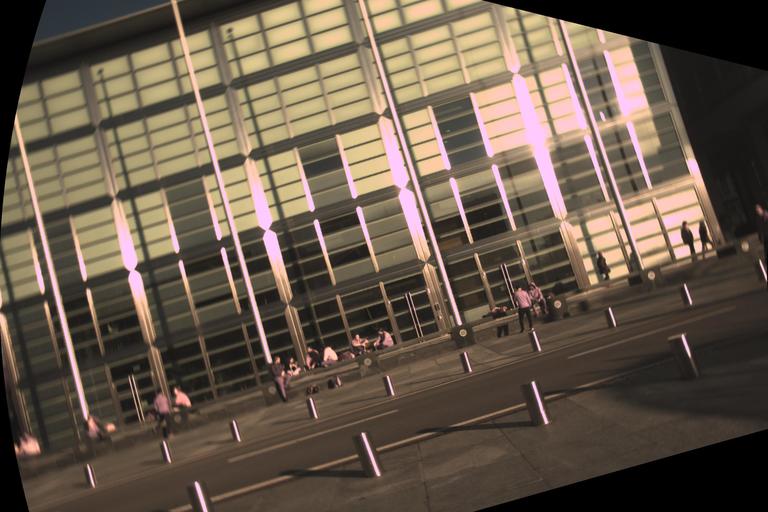} \\
    \end{subfigure}
    \vspace{-5.5em}
    \caption{When two images are recorded with different rolling shutter directions, their motion-induced distortion is different, and a few point correspondences are enough to recover the motion as well as an undistorted image.
    %The right image is unwarped from the top left image using rotation parameters estimated using our solver from one and a half 2D point correspondence.
    }
    \label{fig:eyecatcher}
\end{figure}

% \begin{figure}
%     \centering
%     \renewcommand{\tabcolsep}{0.2pt}
%     \renewcommand{\arraystretch}{0.2}
%      \includegraphics[width=0.45\columnwidth]{figs/rot4_1.jpg} 
%      \includegraphics[width=0.45\columnwidth,angle=-180,origin=c]{figs/rot4_2.jpg} \\
%      \includegraphics[width=0.91\columnwidth]{figs/re_rot4.jpg} \\
%     %  \includegraphics[width=0.35\columnwidth]{figs/rot4_gt.jpg}
%     %  \includegraphics[width=0.353\columnwidth]{figs/rot12_1.jpg} &
%     %   \includegraphics[width=0.353\columnwidth,angle=-180,origin=c]{figs/rot12_2.jpg} &
%     %   \includegraphics[width=0.33\columnwidth]{figs/rot12.jpg} \\
%     % %   \includegraphics[width=0.3\columnwidth]{figs/rot12_gt.jpg}\\
%     \caption{ When two images (top) are recorded with different rolling shutter directions, the motion-induced distortion is different and can be used to recover the motion as well as an undistorted image (bottom).
%     %The bottom image is unwarped from the top left image using rotation parameters estimated using our solver from one and a half 2D point correspondence.
%     }
%     \label{fig:eyecatcher}
% \end{figure}
%

Recently, more emphasis has been put on explicitly removing RS distortion from the images: in this way, one not only obtains visually appealing images, but can also continue to use the whole ensemble of existing, efficient vision algorithms.
For the case of pure camera rotation, the distortion has been modelled as a mixture of homographies~\cite{grundmann_calibration-free_2012}. If only a single image is available, some external constraint is needed to recover the distortion, for instance~\cite{rengarajan_bows_2016} assume a Manhattan world and search for lines, respectively vanishing points. 
And~\cite{lao_robust_2018} relaxes the Manhattan assumption and only requires the (curved) images of straight 3D lines to undistort a single RS image, including a RANSAC scheme to filter out false line matches.
In~\cite{vasu_occlusion-aware_2018} an occlusion-aware undistortion method is developed for the specific setting of $\geq$3 RS images with continuous motion and known time delay, assuming a piece-wise planar 3D scene.
Others propose learning-based methods where a CNN is trained to warp single RS images to their perspective counterparts~\cite{rengarajan_unrolling_2017,Zhuang_2019_CVPR}.

\vspace{0.2em}
\noindent \textbf{Motivation.} Despite more than a decade of research, RS distortion remains a challenge. In fact, when presented with a single RS image, it is impossible to remove the distortion unless one makes fairly restrictive assumptions about the scene~\cite{rengarajan_bows_2016,lao_robust_2018}, or about the camera motion~\cite{purkait_ackermann_2018}, (the best-understood case being pure rotation~\cite{grundmann_calibration-free_2012,rengarajan_unrolling_2017}). The same holds true for learned undistortion~\cite{rengarajan_unrolling_2017,Zhuang_2019_CVPR}, which only works for the types of scenes it has been trained on. Moreover, it does not guarantee a geometrically correct output that downstream processing steps can digest.

Equations for the generalized epipolar geometry of RS cameras have been derived in~\cite{dai_rolling_2016}, however, due to the complexity of the resulting systems there is no practical solution for the RS epipolar geometry. The method of~\cite{ait-aider_structure_2009} utilizes triangulation and therefore requires non-negligible baseline. Furthermore, their solution is iterative, non-minimal and therefore not suitable for RANSAC-style robust estimation. 

Even with multiple views, removing RS distortion either requires strong assumptions, like a piece-wise planar scene observed at a high frame-rate with known shutter timings~\cite{lao_robust_2018}; or it amounts to full SfM reconstruction~\cite{hedborg_rolling_2012}, thus requiring sufficient camera translation.
Moreover, SfM with rolling shutter suffers from a number of degeneracies, in particular it has long been known that constant translational motion along the baseline (e.g., side-looking camera on a vehicle) does not admit a solution~\cite{ait-aider_structure_2009}.
More recently, it has been shown~\cite{albl_degeneracies_2016} that (nearly) parallel RS read-out directions are in general degenerate, and can only be solved with additional constraints on the camera motion~\cite{ito_self-calibration-based_2017}.

RS cameras are nowadays often combined into multi-camera systems. Even in  mid-range smartphones it is common to have two forward-facing cameras, usually mounted very close to each other. It seems natural to ask whether such an assembly allows one to remove RS distortion.

\vspace{0.5em}
\noindent \textbf{Contribution.} We show that a simple modification of the two-camera setup is sufficient to facilitate removal of RS distortion: the cameras must have known baseline -- ideally of negligible length, as in the typical case of smartphones (for scene depth $>$1$\,$m); and their RS read-out directions must be significantly different -- ideally opposite to each other. Finally, the cameras must be synchronized (the offset between their triggers must be known). If those conditions are met, the motion of the cameras can be recovered and a perspective (global shutter) image can be approximated from the two RS images, regardless of image content. The only requirement is enough texture to extract interest points. 

We also show that if the cameras undergo translational motion, depth maps can be computed given optical flow between the images, and that the undistorted sparse features can be used in an SfM pipeline to obtain a reconstruction similar to one from GS images.

In the following, we investigate the geometry of this configuration and propose algorithms to compute the motion parameters needed to remove the RS distortions. We first discuss the solution for the general case of unconstrained 6DOF motion, and develop an efficient solver that can be used in RANSAC-type outlier rejection schemes.
We then go on to derive simpler solutions for special motion patterns that often arise in practice, namely pure rotation, pure translation, and translation orthogonal to the viewing direction. These cases are important because in real applications the motion is often restricted or known, e.g., when the camera is mounted on the side of a vehicle.
The proposed methods require only point-to-point correspondences between the images to compute the camera motion parameters.

% \vspace{0.5em}
% \noindent
\section{Problem formulation}
\label{sec:problem_formulation}
\noindent Throughout, we assume two images taken by (perspective) cameras with rolling shutter. In those images, %2D~$\!\leftrightarrow\!$~2D 
corresponding points have been found such that $\V{u}_i = [u_{i}\;v_{i}\;1]^\top$ in the first camera corresponds to $\V{u}'_i = [u^\prime_{i}\;v^\prime_{i}\;1]^\top$ in the second camera, and both are projections of the same 3D point $\V{X}_i = [X_{i1},\;X_{i2},\;X_{i3},\;1]^\top$.
If neither the cameras nor the objects in the scene move, we can model the projections in both cameras as if they had a global shutter, with projection matrices $\M{P}_g$ and $\M{P}^\prime_g$~\cite{hartley_multiple_2003} such that 
\begin{equation}
    \begin{array}{rcl}
       %\lambda_g\V{u}_{gi}=\M{P}_g\V{X}_i  & = & \M{K}\left[\M{R} \mid \V{C}\right]\V{X}_i \\
       \lambda_g\V{u}_{gi}=\M{P}_g\V{X}_i  & = & \M{K}\left[\M{R} \mid {-\M{R}}\V{C}\right]\V{X}_i \\
       \lambda^\prime_g\V{u}^\prime_{gi}=\M{P}^\prime_g\V{X}_i  & = & \M{K}^\prime\left[\M{R}^\prime \mid {-\M{R}^\prime}\V{C}^\prime\right]\V{X}_i.
    \end{array}
     \label{eq:persp_proj_2cam}
\end{equation}
where $\M{K}$, $\M{K}^\prime$ are the camera intrinsics, $\M{R}$, $\M{R}^\prime$ are the camera rotations, $\V{C}$, $\V{C}^\prime$ are the camera centers, and $\lambda_g$, $\lambda^\prime_g$ are the scalar perspective depths. 

If the cameras move during image capture, we have to generalize the projection function. Several motion models were used to describe the RS geometry~\cite{meingast_geometric_2005,ait-aider_structure_2009,magerand_global_2012,albl_r6p_2015,saurer_minimal_2015}. These works have shown that, in most applications, assuming constant rotational and translational velocities\footnote{For constant rotational velocity we have $R_\omega(\alpha) = \exp(\alpha[\omega]_\times)$.} during RS read-out is sufficient to obtain the initial motion estimate, which can be further improved using a more complex  model~\cite{ringaby_efficient_2012} in the refinement step.
The projection matrices $\M{P}(v_{i})$ and $\M{P}^\prime(v^\prime_{i})$ are now functions of the image row, because each row is taken at a different time and hence a different camera pose. We can  write the projections as
%\begin{equation}
%    \begin{array}{rcccl}
       %\lambda\V{u}_{i} & = &\M{P}(v_{i})\V{X}_i  & = & \M{K}\left[\M{R}_\omega(v_{i})\M{R} \mid \V{C}+\V{t}\,v_{i}\right]\V{X}_i \\
%       \lambda\V{u}_{i} & = &\M{P}(v_{i})\V{X}_i  & = & \M{K}\left[\M{R}_\omega(v_{i})\M{R} \mid {-\M{R}\V{C}}+\V{t}\,v_{i}\right]\V{X}_i \\
%       \lambda^\prime\V{u}^\prime_{i} & = & \M{P}^\prime(v^\prime_{i})\V{X}_i & = & \M{K}^\prime\left[\M{R}^\prime_\omega(v^\prime_{i})\M{R}^\prime \mid {-\M{R}\V{C}^\prime}+\V{t^\prime}v^\prime_{i}\right]\V{X}_i.
%    \end{array}
%\end{equation}
\begin{align}
       %\lambda\V{u}_{i} & = &\M{P}(v_{i})\V{X}_i  & = & \M{K}\left[\M{R}_\omega(v_{i})\M{R} \mid \V{C}+\V{t}\,v_{i}\right]\V{X}_i \\
       \lambda\V{u}_{i}  = &\M{P}(v_{i})\V{X}_i   = \M{K}\left[\M{R}_\omega(v_{i})\M{R} \mid {-\M{R}\V{C}}+\V{t}\,v_{i}\right]\V{X}_i \label{eq:rs_proj_2cam} \\ \nonumber
       \lambda^\prime\V{u}^\prime_{i}  = & \M{P}^\prime(v^\prime_{i})\V{X}_i  =  \M{K}^\prime\left[\M{R}^\prime_{\omega^\prime}(v^\prime_{i})\M{R}^\prime \mid {-\M{R}^\prime\V{C}^\prime}+\V{t^\prime}v^\prime_{i}\right]\V{X}_i\;,
\end{align}
where $\M{R}_\omega(v_i)$, $\M{R}^\prime_{\omega^\prime}(v_i)$ are the rotational and $\V{t}$,$\V{t}^\prime$ the translational motions during image acquisition. 

Let us now consider the case when the relative pose of the cameras is known and the baseline is negligibly small (relative to the scene depth). This situation cannot be handled by existing methods~\cite{ait-aider_structure_2009,dai_rolling_2016}, but is the one approximately realized in multi-camera smartphones:
the typical distance between the cameras is $\approx$1$\,$cm, which means that already at 1$\,$m distance the base-to-height ratio is 100:1 
and we can safely assume $\V{C}=\V{C}'$ (analogous to using a homography between global shutter images). We point out that the approximation is independent of the focal length, \ie, it holds equally well between a wide-angle view and a zoomed view.

For simplicity, we consider the cameras to be calibrated,  $\M{K}\!=\!\M{K}^\prime\!=\!\M{I}$, and attach the world coordinate system to the first camera, $\M{R}\!=\!\M{I}$ and $\V{C}\!=\!\V{C}^\prime\!=\!\V{0}$, to obtain
\begin{equation}
    \begin{array}{rcl}
       \lambda\V{u}_{i}  & = & \left[\M{R}_\omega(v_{i}) \mid \V{t}\,v_{i}\right]\V{X}_i \\
       \lambda^\prime\V{u}^\prime_{i} & = & \left[\M{R}^\prime_{\omega^\prime}(v^\prime_{i})\M{R}_r \mid \V{t^\prime}v^\prime_{i}\right]\V{X}_i\;,
    \end{array}
     \label{eq:rs_proj_2cam_simple}
\end{equation}
where $\M{R}_r$ is now the relative orientation of the second camera w.r.t.\ the first one. Since the cameras are assembled in a rigid rig, their motions are always identical and we have $\V{t}^\prime\!=\!\M{R}_r\V{t}$ and $\M{R}^\prime_{\omega^\prime}(v^\prime_i)\!=\!\M{R}_r\M{R}_{\omega}(v^\prime_i)\M{R}_r^\top$. This yields even simpler equations in terms of the number of unknowns,
\begin{equation}
    \begin{array}{rcl}
       \lambda\V{u}_{i}  & = & \left[\M{R}_\omega(v_{i}) \mid \V{t}\,v_{i}\right]\V{X}_i \\
       \lambda^\prime\V{u}^\prime_{i} & = & \left[\M{R}_r\M{R}_\omega(v^\prime_{i}) \mid \M{R}_r\V{t}\,v^\prime_{i}\right]\V{X}_i.
    \end{array}
     \label{eq:rs_proj_2cam_simplest}
\end{equation}
From eq.~\eqref{eq:rs_proj_2cam_simplest} one immediately sees that the further $\M{R}_r$ is from identity, the bigger is the difference between the images and between their RS distortions. Since these differences are our source of information, we want them to be as large as possible and focus on the case of 180$^\circ$ rotation around the $z$-axis, $\M{R}_r\!=\!\mbox{diag}([-1,\,-1,\,1,])$.
%\begin{equation}
%\label{eq:Rr}
%    \M{R}_r = \mat{rrr}{-1 & 0 & 0 \\ 0 & -1 & 0 \\ 0 & 0 & %1}\;.
%\end{equation}
In this setting the second camera is upside down and its RS read-out direction is opposite to that of the first one. Note that this is equivalent to $\M{R}_r\!=\!\M{I}$ with the shutter direction reversed. If we flip the second image along the $x$- and $y$-axes, it will be identical to the first one in the absence of motion, but if the rig moves the RS distortions will be different.
This setup is easy to construct in practice: the only change to the standard layout is to reverse the read-out direction of one camera.

Also note that it is fairly straightforward to extend the algorithms derived below to shutters that are not 180$^\circ$ opposite, as well as to 
non-zero baseline with known relative orientation, as in~\cite{ait-aider_structure_2009} (e.g.\ stereo cameras on autonomous robots). In all these scenarios different RS directions make it possible to remove the distortion.
\section{Solutions}
\label{sec:solutions}
\noindent In this section we will describe how to solve the problem for typical types of motions.
\subsection{General 6DOF motion}
\label{sec:6dof_problem}
\noindent To solve for the general motion case, we start from eq.~\eqref{eq:rs_proj_2cam_simplest}. Without loss of generality, we can choose the first camera as the origin of the world coordinate system, such that
\begin{equation}
            \begin{array}{rcl}
        \M{P}&=&\left[\M{I} \mid \V{0}\right] \\
        \M{P}^\prime\left(v_{i},v^\prime_{i}\right) &=&[\M{R}(v_i,v^\prime_{i}) 
        \mid v^\prime_{i}\M{R}_r\V{t} - v_{i}\M{R}(v_{i},v^\prime_{i})\V{t}]
   \end{array}
    \label{eq:rs_2cam_Pmatrices_I}
\end{equation}
where $\M{R}(v_{i},v^\prime_{i})=\M{R}_r\M{R}_\omega(v^\prime_{i})\M{R}_\omega(v_{i})^\top$. We can consider  $\M{R}(v_{i},v^\prime_{i})$ and $\V{t}(v_i,v^\prime_i) = v^\prime_{i}\M{R}_r\V{t} - v_{i}\M{R}(v_{i},v^\prime_{i})\V{t}$ as the relative camera orientation and translation for each pair of lines. This yields one essential matrix 
\begin{equation}
    \M{E}(v_i,v_i^\prime) = \left[\V{t}(v_i,v_i^\prime)\right]_\times \M{R}(v_{i},v^\prime_{i})\;,
    \label{eq:rs_2cam_essential}
\end{equation}
for each pair of lines, with six unknowns. The translation can only be determined up to a unknown scale, using 5 correspondences. We next describe how to simplify the equations further and produce an efficient minimal solver.
\subsection{Minimal solver for the 6DOF motion}
\label{sec:6dof_solver}
\noindent Since both cameras form a rig, the two rotations $\M{R}_\omega(v_i')$, $\M{R}_\omega(v_i)$ have the same  axis and we have 
\begin{equation}
    \M{R}_\omega(v_i') \M{R}_\omega(v_i)^\top = \M{R}_\omega(v_i'-v_i)\;.
\end{equation}
For convenience, let $\M{R}_i = \M{R}_r\M{R}_\omega(v_i'-v_i)$. Then the instantaneous essential matrix for rows $v_i$ and $v_i'$ can be written
\begin{align}
    \M{E}(v_i,v_i') = [v_i'\M{R}_r\V{t} - v_i\M{R}_i \V{t}]_{\times} \M{R}_i = \hspace*{0.33\linewidth} \label{eq:simplfied_E} \\
     = v_i' [\M{R}_r\V{t}]_{\times}\M{R}_i - v_i[\M{R}_i \V{t}]_{\times} \M{R}_i
    = v_i' [\M{R}_r\V{t}]_{\times}\M{R}_i - v_i \M{R}_i [\V{t}]_{\times} \nonumber 
\end{align}
using the identity $(\M{R}\V{u}\times \M{R}\V{v}) = \M{R}(\V{u}\times\V{v})$. As  the motion due to the RS effect is small, we linearise it, as often done for RS processing, \eg, \cite{ait-aider_simultaneous_2006,magerand_global_2012,albl_rolling_2019,albl_linear_2018}. 
For constant angular velocity we get
\begin{equation}
    \M{R}_i \approx \M{R}_r(\M{I}_{3\times 3} + (v_i'-v_i) [\V{\omega}]_{\times}),
\end{equation}
where the direction of $\V{\omega}$ encodes the axis of rotation, and the angular velocity determines its magnitude. Inserting this into \eqref{eq:simplfied_E} we get
\begin{equation}
\begin{split}
     \M{E}(v_i,v_i')=&
     v_i' \M{R}_r[\V{t}]_{\times}(\M{I}+(v_i'-v_i) [\V{\omega}]_{\times}\!)\\&-v_i \M{R}_r(\M{I}+(v_i'-v_i) [\V{\omega}]_{\times}) [\V{t}]_{\times}
     \end{split}
\end{equation}
Each pair of corresponding points $(\V{u}_i,\V{u}'_i)$ now yields a single equation from the epipolar constraint,
\begin{equation} 
\V{u}_i^{\prime\top} \M{E}(v_i,v_i') \V{u}_i = 0.
\label{eq:approx_epipolar}
\end{equation}
This is a quadratic equation in the unknowns $\V{\omega}$ and $\V{t}$. Since the scale of the translation is unobservable (eq.~\eqref{eq:approx_epipolar} is homogeneous in $\V{t}$), we add a linear
constraint $\V{t}_1 + \V{t}_2 = 1$, leading  to the parameterisation
%\begin{equation}
$\V{t} = (1-x, x, y)^\top$.
%\end{equation}
Note that this constraint is degenerate for pure forward motion.
%(i.e., $\V{t}_1=\V{t}_2=0$). 

From 5 point correspondences we get 5 quadratic equations in 5 unknowns $x,y$ and $\V{\omega}$, which we solve with the hidden variable technique \cite{cox_using_2005}. We rewrite the equations as
\begin{equation}
    \M{M}(\V{\omega}) \begin{bmatrix}x &  y& 1\end{bmatrix}^\top = 0\;,
\end{equation}
where $\M{M}(\V{\omega})$ is a $5\times 3$ matrix with elements that depend linearly on $\V{\omega}$. This matrix must be rank deficient, thus all $3\times 3$ sub-determinants must vanish, which gives $10$ cubic equations in $\V{\omega}$, with up to, in general, 10 solutions. Interestingly, the equations have the same structure as the classic determinant and trace constraints for the essential matrix. To solve them one can thus employ any of the known solutions for that case \cite{Nister-5pt-PAMI-2004,Hartley-PAMI-2012}. We use the solver generator \cite{larsson_efficient_2017}.

In a similar fashion, we can produce a solution for the case of a fixed, known baseline between the cameras. Please see the Appendix section~\ref{sec:a_baseline_solver} for details.

\subsection{Pure rotation}
\noindent Next, let us consider the case where the cameras only rotate around the center of projection. We now have $\V{t}=\V{0}$ and $\M{R}_\omega(\alpha) \neq \M{I}$ for $x\in \mathbb{R}\setminus\V{0}$. Equations~\eqref{eq:rs_proj_2cam_simplest} become
\begin{equation}
    \begin{array}{rcl}
       \lambda\V{u}_{i}  & = & \left[\M{R}_\omega(v_i) \mid \V{0}\right]\V{X}_i \\
       \lambda^\prime\V{u}^\prime_{i} & = & \left[\M{R}_r\M{R}_\omega(v^\prime_{i}) \mid  \V{0}\right]\V{X}_i.
    \end{array}
\end{equation}
and we can express the relationship between $\V{u}_i$ and $\V{u}^\prime_i$ as
\begin{equation}
    \begin{array}{rcl}
       \lambda^\prime\V{u}^\prime_{i} & = & \M{R}_\omega(v^\prime_{i})\M{R}_r\M{R}^\top_\omega(v_i)\lambda\V{u}_i.
    \end{array}
\label{eq:rs_2cam_homography}
\end{equation}
This resembles a homography between GS images, except that the matrix $\M{H}=\M{R}_\omega(v^\prime_{i})\M{R}_r\M{R}^\top_\omega(v_i)$ changes for every correspondence. To get rid of $\lambda,\lambda^\prime$ we divide~\eqref{eq:rs_2cam_homography} by $\lambda$,
%
%to obtain
%\begin{equation}
%    \begin{array}{rcccl}
%       \frac{\lambda^\prime}{\lambda}\V{u}^\prime_{i} & = & %\hat{\lambda}\V{u}^\prime_{i} & = & %\M{R}_\omega(v^\prime_{i})\M{R}_r\M{R}^\top_\omega(v_i)\V{u}_i.
%    \end{array}
%\end{equation}
%
then left-multiply with the skew-symmetric $[\V{u}^\prime_{i}]_\times$ to obtain
\begin{equation}
    \V{0} = [\V{u}^\prime_{i}]_\times\M{R}_\omega(v^\prime_{i})\M{R}_r\M{R}^\top_\omega(v_i)\V{u}_i
\end{equation}
For constant angular velocity we now have three unknown parameters for the rotation $\M{R}_\omega(\alpha)$. Each correspondence yields two equations, so the solution can be found from 1.5 correspondences.
If we further linearise $\M{R}_\omega(\alpha)$ via first-order Taylor expansion, as in~\cite{ait-aider_simultaneous_2006,magerand_global_2012,albl_rolling_2019}, we get
\begin{equation}
    \V{0} = [\V{u}^\prime_{i}]_\times\left(\M{I}+v^\prime_{i}[\omega]_\times\right)\M{R}_r\left(\M{I}-v_i[\omega]_\times\right)\V{u}_i\;,
\end{equation}
where $\omega$ is the angle-axis vector. This is a system of three 2$^\text{nd}$-order equations in three unknowns, which can be solved efficiently with the \emph{e3q3} solver~\cite{kukelova_efficient_2015}.

\subsection{Translation}
\label{sec:translation}
\noindent Next, let us consider a general translation with constant velocity and no rotation, $\M{R}_\omega(\alpha) = \M{I}$ . Substituting $\M{R}_\omega(\alpha) = \M{I}$ in equations~\eqref{eq:rs_proj_2cam_simplest} and subtracting the second equation from the first one we obtain (for details see the Appendix~\ref{sec:a_txyz})
\begin{equation}
    \mat{c}{t_x \\ t_y \\ t_z}v_i-\mat{c}{t_x \\ t_y \\ t_z}v^\prime_{i} = \mat{c}{u_{i} \\ v_i \\ 1}\lambda_i-\mat{c}{-u^\prime_{i} \\ -v^\prime_{i} \\ 1}\lambda^\prime_i\;.
    \label{eq:rs_twocam_linsys_txyz}
\end{equation}
Each correspondence adds three equations and two unknowns $\lambda_i,\lambda^\prime_i$. Two correspondences give us 6 linear homogeneous equations in 7 unknowns $t_x,t_y,t_z,\lambda_1,\lambda^\prime_1,\lambda_2,\lambda^\prime_2$, which allow us to find a solution up to scale, \ie, relative to one depth (e.g., $\lambda_1$) or to the magnitude of the translation.

\vspace{0.25em} \noindent \textbf{Translation in the $xy$-plane}
\noindent We also consider the case of translation orthogonal to the viewing direction, $\V{t}=\mat{ccc}{t_x & t_y & 0}$. The system~\eqref{eq:rs_twocam_linsys_txyz} now lacks $t_z$ in the 3rd equation and we find that $\lambda_i=\lambda^\prime_i$, i.e.\ the perspective depth of a point $\V{X}_i$ is the same in both cameras. By solving this system for $t_x$ and $t_y$, we can express them in terms of $\lambda_i$,
\begin{equation}
    % \begin{array}{rcl}
    t_x = \frac{u_{i}+u^\prime_{i}}{v_i-v^\prime_{i}}\lambda_i \;,
    t_y = \frac{v_i+v^\prime_{i}}{v_i-v^\prime_{i}}\lambda_i
    % \end{array}
    \;,
    \label{eq:tx_ty_alpha}
\end{equation}
and obtain an equivalent global shutter projection as 
\begin{equation}
  %  \begin{array}{rcl}
 %   \V{u}_{gi} & = & \mat{c}{\frac{u_{i}v^\prime_{i}-u^\prime_{i}v_i}{v_i-v^\prime_{i%}}\\
 %   \frac{-2v_iv^\prime_{i}}{v_i-v^\prime_{i}}
 %   \\1}\;.
 %   \end{array}
 \V{u}_{gi}  =  [\frac{u_{i}v^\prime_{i}-u^\prime_{i}v_i}{v_i-v^\prime_{i}},
    \frac{-2v_iv^\prime_{i}}{v_i-v^\prime_{i}},1]^{\top}.
\end{equation}

\vspace{0.25em} \noindent \textbf{Translation along $x$-axis}
\noindent  Finally, let us assume a translation only along the camera $x$-axis, such as for a side-looking camera on a moving vehicle, or when observing passing cars. In this case the global shutter projection satisfies 
%\begin{equation}
    $\V{u}_{gi} = [\frac{u_{i}+u^\prime_{i}}{2}, v_i, 1]^{\top}$
%\end{equation}
(see the Appendix~\ref{sec:a_tx} for a detailed derivation), which implies that for constant velocity along the $x$-axis we can obtain GS projections by simply interpolating between the $x$-coordinates of corresponding points in the two RS images. Analogously we interpolate the $y$-coordinate for translation along $y$-axis.

\section{Refinement with advanced motion model}
\noindent Our minimal solutions resort to simplified motion models to support real-time applications and RANSAC. After obtaining an initial solution with one of the minimal solvers, we can improve the motion estimates through a non-linear refinement with a more complex model of camera motion. 

For the rotation case, we can obtain the cost function from eq.~(\ref{eq:rs_2cam_homography}) and sum the residuals of all correspondences
\begin{equation}
    \sum_{i=0}^{N}{\norm{\mat{c}{u_i \\ v_i}-\mat{c}{\V{h}_1^i\V{u}^\prime_i / \V{h}^i_3\V{u}^\prime_i \\ \V{h}^i_2\V{u}^\prime_i / \V{h}^i_3\V{u}^\prime_i}}}
    \label{eq:refinement_w}\;,
\end{equation}
where $\M{H}_i =\!\mat{c}{\V{h}_i^{1\top} \V{h}_i^{2\top} \V{h}_i^{3\top}}^\top\!\!\!= \M{R}_\omega(v^\prime_{i})\M{R}_r\M{R}^\top_\omega(v_i)$, and $\M{R}_\omega(\alpha)$ is now parametrised via the Rodrigues formula. 

For the translation case, we can minimise the the Sampson error, as in~\cite{dai_rolling_2016}, which leads to the cost function
\begin{equation}
        \sum_{i=0}^{N}{\frac{\left(\V{u}_i^{\prime\top} \M{E}_i \V{u}_i\right)^2}{\left(\M{E}_i \V{u}_i\right)_1^2+\left(\M{E}_i \V{u}_i\right)_2^2+\left(\M{E}_i^\top \V{u}_i^\prime\right)_1^2 + \left(\M{E}_i \V{u}_i^\prime\right)_2^2}}
        \label{eq:refinement_wt}
\end{equation}
with $\M{E}_i=v_i' [\M{R}_r\V{t}]_{\times}\M{R}_i - v_i \M{R}_i [\V{t}]_{\times}$ as defined in equation~\ref{eq:simplfied_E}. Again $\M{R}_i$ is defined via the Rodrigues formula.

It has been shown~\cite{ringaby_efficient_2012} that a uniform motion model across the entire image may not be sufficient, instead using three different motions defined at their respective "knots" worked better for handheld footage. If desired, this extension is also applicable in our refinement step. To that end, we simply define intermediate poses for the camera system by taking the initial parameters $\omega,\V{t}$ from the minimal solver, and then minimize either~(\ref{eq:refinement_w}) or~(\ref{eq:refinement_wt}), with the instantaneous rotation and translation of the cameras interpolated between those intermediate poses. 

\section{Undistorting the image}\label{sec:undist}
\noindent Once the motion parameters have been estimated, we can chose between two approaches to undistort the images, based on whether we treat pure rotation or also translation.

\vspace{0.25em}
\noindent \textbf{Rotation and image warping.} Under pure rotation, creating an image in global shutter geometry is simple. For each pixel, we have a forward mapping $\lambda\V{u}_{gi}=\M{R}_\omega(v_i)^\top\V{u}_i$ from the first image and $\lambda\V{u}_{gi}=\M{R}_\omega(v^\prime_i)^\top\V{u}^\prime_i$ from the second image into the virtual GS image plane. This mapping depends only on the row index of the respective RS image, which defines the time and therefore the pose. No pixel-wise correspondences are needed. We can also use the backward mapping $\lambda\V{u}_i=\M{R}_\omega(v_i)\V{u}_{gi}$, in that case $v_i$ appears on both sides of the equation, possibly in non-linear form, depending on how $\M{R}_\omega(v_i)$ is parameterised. In either case, we can iteratively solve for $v_i$, starting at the initial value $v_{gi}$. One can transform both RS inputs to the same virtual GS image for a more complete result, see the Appendix~\ref{sec:a_combine}.

\vspace{0.25em}
\noindent \textbf{Translation and dense undistortion.} Translational RS distortion poses a greater challenge, because the transformation of each pixel depends on its correspondence in the other RS image. Consequently, one must recover dense correspondence to obtain a dense, visually pleasing result~\cite{rengarajan_bows_2016,zhuang_rolling-shutter-aware_2017,vasu_occlusion-aware_2018}. Empirically, we found optical flow methods to work best, particularly PWC-net~\cite{sun_pwc-net:_2017} consistently gave the best results. 
Given dense correspondences, we can convert them to a depth map and use it to transfer image content from the RS images to the virtual GS image, using some form of $z$-buffering to handle occlusions.
Note that rows near the middle of the image have been captured at (nearly) the same time and pose in both RS views, thus depth is not observable there.
Fortunately, this barely impacts the undistortion process, because also the local RS distortions are small in that narrow strip, such that they can be removed by simply interpolating between the inputs. For details, see the Appendix~\ref{sec:a_undist_tr}.

\section{Experiments}
\begin{figure}
    \captionsetup[subfigure]{labelformat=empty}
    \renewcommand{\tabcolsep}{0.5pt}
    \renewcommand{\arraystretch}{0.5}
\begin{subfigure}{.35\columnwidth}
\centering
\includegraphics[width=7em,angle=-90]{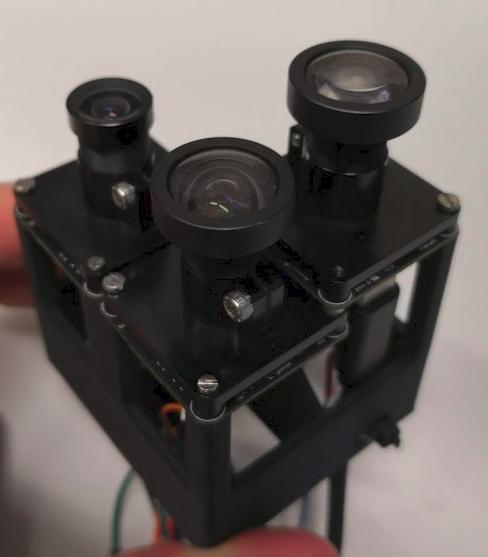}
\end{subfigure}
\begin{subfigure}{.65\columnwidth}
\centering
\begin{tabular}{cccc} \toprule
         Alg. & \# corr.& Param. & Runtime \\  \midrule
         $\mathbf{txy}$ & 1 & $t_x,t_y$ & $\approx$0\\
         $\mathbf{txyz}$ & 2 & $t_x,t_y,t_z$ & $\approx$0\\
         $\boldsymbol\omega$ & 2 & $\omega_x,\omega_y,\omega_z$ & 4$\,\mu \text{s}$\\
         $\boldsymbol\omega\mathbf{t}$ & 5 & $t_x,t_y,t_z,$ & 40$\,\mu\text{s}$\\
         && $\omega_x,\omega_y,\omega_z$ \\ \bottomrule
    \end{tabular}
\end{subfigure}
    
    \caption{Rig with two RS and one GS camera (left); Solvers used in the experiments (right).}
    \label{fig:rig}
\end{figure}

\noindent We have tested the proposed algorithms both with synthetic and with real data. Synthetic correspondences serve to quantitatively analyze performance. Real image pairs were acquired with a rig of two RS cameras, see Fig.~\ref{fig:rig}, and undistorted into global shutter geometry to visually illustrate the high quality of the corrected images across a range of motion patterns and scenes.
 
\vspace{0.25em}
\noindent \textbf{Synthetic data.} To generate synthetic data in a realistic manner, we started from the GS images of~\cite{hedborg_rolling_2012} and performed SfM reconstruction. We then placed virtual RS pairs at the reconstructed camera locations, to which we applied various simulated motions with constant translational and angular velocities, and reprojected the previously reconstructed 3D structure points. The angular velocity was gradually increased up to 30 degrees/frame which means that the camera has rotated by 30 degrees during the acquisition of one frame. The translational velocity was increased up to 1/10 of the distance of the closest part of the scene per frame. Gaussian noise with $\mu=0$ and $\sigma=0.5pix$ was added to the coordinates of the resulting correspondences. We generated around 1.4K images this way.

We test four different solvers, two for pure translations, one for pure rotation, and one for the full 6DOF case, see Fig.~\ref{fig:rig}. Additionally, we run a simple baseline that just averages the image coordinates of the two corresponding points (\emph{err-interp}). We do not consider the single axis translation solvers $\mathbf{tx}$ and $\mathbf{ty}$, since they are covered by $\mathbf{txy}$, which also requires only a single correspondence.

Three different variants are tested for each solver:
\emph{(v1)}~individually fits a motion model at each correspondence, sampling the minimal required set of additional correspondences at random and undistorting the individual coordinate. This local strategy can handle even complex global distortions with a simpler motion model, by piece-wise approximation. The downside is that there is no redundancy, hence no robustness against mismatches.
\emph{(v2)}~is a robust approach that computes a single, global motion model for the entire image and uses it to undistort all correspondences. The solver is run with randomly sampled minimal sets, embedded in a LO-RANSAC loop~\cite{chum_locally_2003} that verifies the solution against the full set of correspondences and locally optimizes the motion parameters with non-linear least squares.
\emph{(v3)}~explores a hybrid LO-RANSAC method that uses one of the simpler models to generate an initial motion estimate, but refines it with the full model with parameters $\{t_x,t_y,t_z,\omega_x,\omega_y,\omega_z\}$.

\begin{figure}[t]
    \centering
    \includegraphics[width=1\columnwidth]{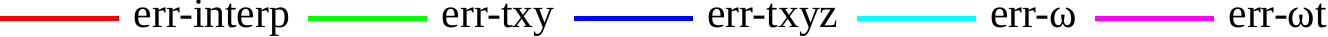}
    \includegraphics[width=0.45\columnwidth]{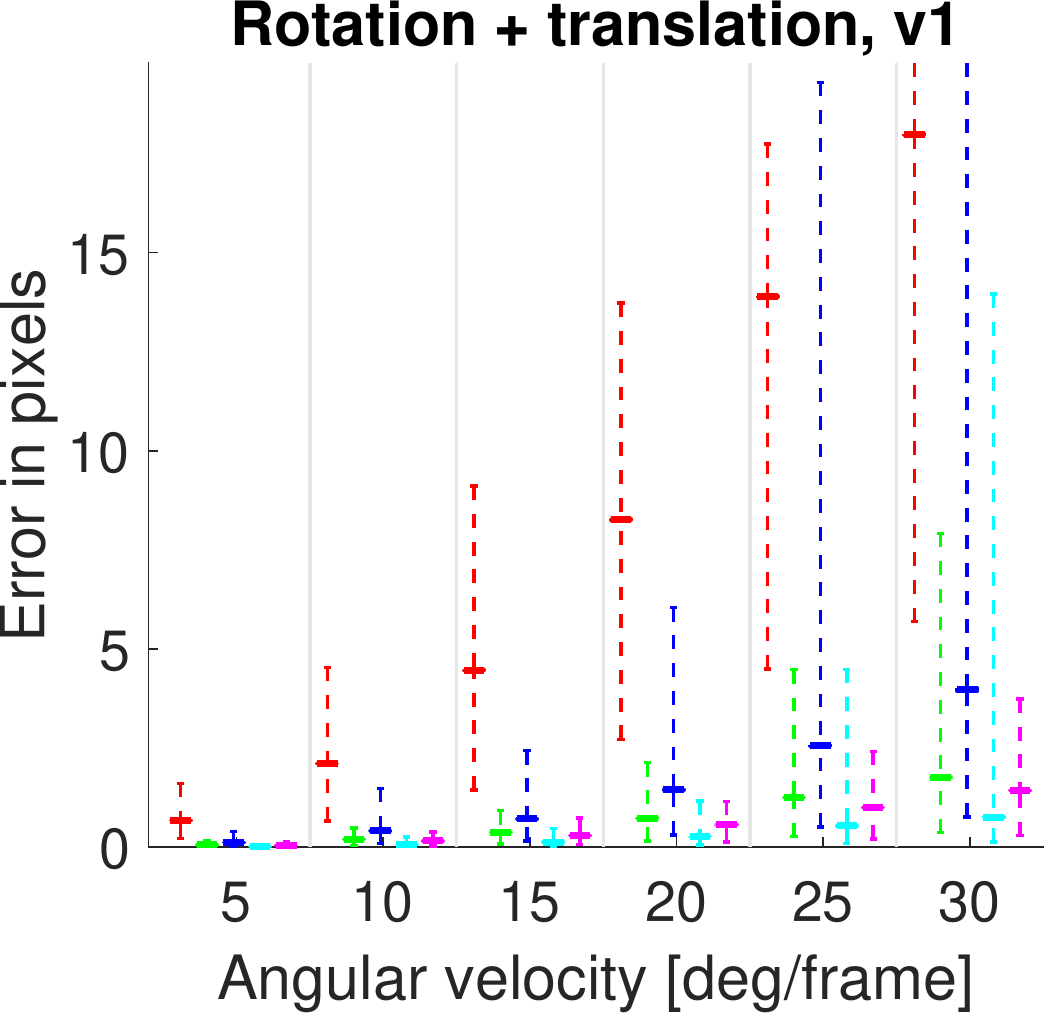}
    \includegraphics[width=0.45\columnwidth]{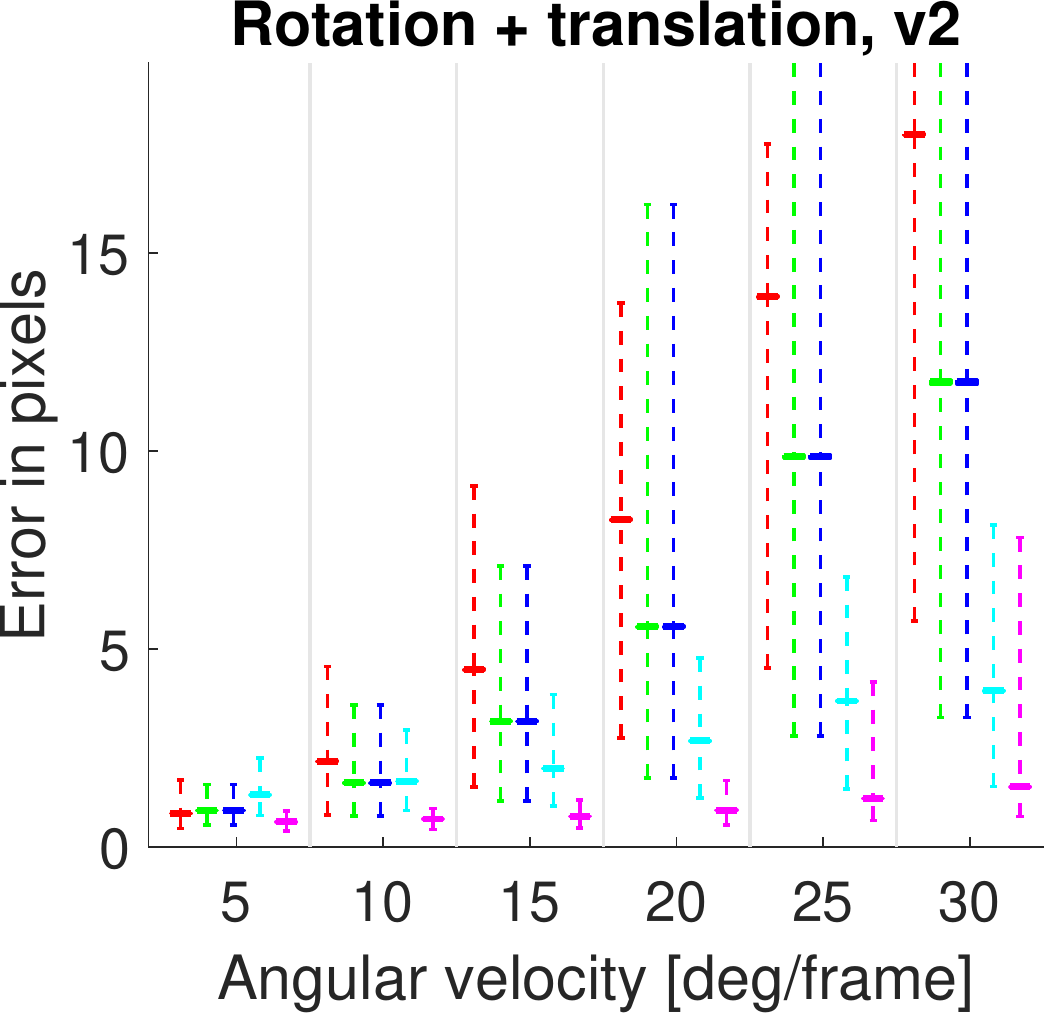}
    \caption{Results on synthetic RS pairs. Fitting a separate motion model per correspondence (left), and fitting a single model per image (right). Plots show the error of undistorted correspondences w.r.t.\ the ground truth GS image.}
    \label{fig:synth}
\end{figure}

The results provide some interesting insights, see Fig.~\ref{fig:synth}. As expected, the full $\boldsymbol\omega\mathbf{t}$ performs best, followed by the rotation-only model $\boldsymbol\omega$. The translational solvers, including the simple $\mathbf{txy}$, work well when used locally per correspondence (\emph{v1}), moreover the error has low variance. This means that the rotational distortion component can be well approximated by piece-wise translations $\mathbf{txy}$, whose estimation is more reliable than that of both $\mathbf{txyz}$ and $\boldsymbol\omega$.

With a single, global RANSAC fit (\emph{v2}) the residual errors of the undistorted points are generally higher (except for $\boldsymbol\omega\mathbf{t}$), due to the more rigid global constraint. The drop is strongest for $\mathbf{txy}$ and $\mathbf{txyz}$, i.e., a global translation model cannot fully compensate rotational distortions.
% See sample result in Fig.~\ref{fig:synth_sample}.
%
The hybrid solution (\emph{v3}) is not shown since it does not improve over the global one (\emph{v2}), suggesting that the general model gets stuck in local optima when initialised with a restricted solver.

% \begin{table}[t]
%     \centering
%     \begin{tabular}{ccc} \toprule
%          Alg. & \# corr.& Param.\\ \midrule
%          $\mathbf{txy}$ & 1 & $t_x,t_y$ \\
%          $\mathbf{txyz}$ & 2 & $t_x,t_y,t_z$\\
%          $\boldsymbol\omega$ & 2 & $\omega_x,\omega_y,\omega_z$\\
%          $\boldsymbol\omega\mathbf{t}$ & 5 & $\arr{c}{t_x,t_y,t_z, \\ \omega_x,\omega_y,\omega_z}$ \\ \bottomrule
%     \end{tabular}
%     \caption{Summary of tested solvers. }
%     \label{tab:methods}
% \end{table}

\begin{figure}[t]
    \centering
    \includegraphics[width=0.8\columnwidth]{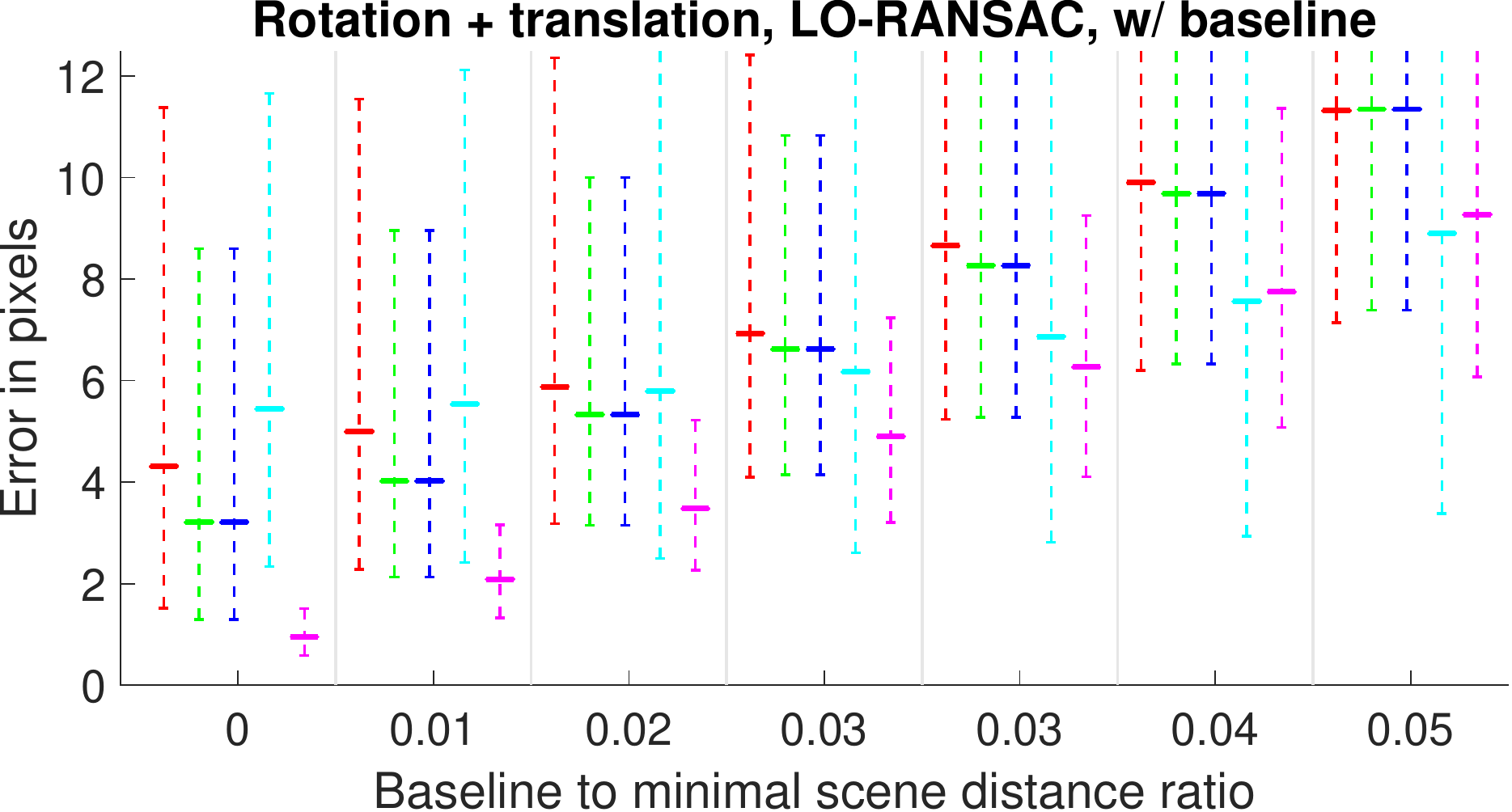}
    \caption{The impact of baselines $>$0 is negligible up to base-to-depth ratio $\approx$1:100, and remains low up to $\approx$1:30. }
    \label{fig:exp_baseline}
\end{figure}

% \begin{figure*}[t]
%     \centering
%     \includegraphics[width=0.25\columnwidth]{figs/rs_2cam_synth_rs1.eps}
%     \includegraphics[width=0.25\columnwidth]{figs/rs_2cam_synth_rs2.eps}
%     \includegraphics[width=0.25\columnwidth]{figs/rs_2cam_synth_gs.eps}
%     \includegraphics[width=0.25\columnwidth]{figs/rs_2cam_synth_interp.eps}
%     \includegraphics[width=0.25\columnwidth]{figs/rs_2cam_synth_txy.eps}
%     \includegraphics[width=0.25\columnwidth]{figs/rs_2cam_synth_txyz.eps}
%     \includegraphics[width=0.25\columnwidth]{figs/rs_2cam_synth_w.eps}
%     \includegraphics[width=0.25\columnwidth]{figs/rs_2cam_synth_wt.eps}
%     \caption{Sample result of all methods on synthetic data. Input RS distorted images, ground truth GS image and the results of our algorithms compared to it.}
%     \label{fig:synth_sample}
% \end{figure*}

\vspace{0.25em}
\noindent \textbf{Zero-baseline assumption.} Next, we evaluate the approximation error due to our assumption of negligible baseline between the two RS cameras. We start with a synthetic experiment, with the same setup as in Fig.~\ref{fig:synth}, but this time with baselines $\neq$0.
We use the global model fit \emph{(v2)}, with angular velocity 15$^\circ$/frame.
The baseline was increased from 0 to 5\% of the smallest scene depth. As shown in  Fig.~\ref{fig:exp_baseline}, the zero-baseline assumption has negligible effect for base-to-depth ratios up to 1:100, i.e., at a typical smartphone baseline of at most 2$\,$cm for objects $\geq$2$\,$m from the camera. Even at an extreme value of 1:20 (40$\,$cm from the camera) the approximation error remains below 10 pixels.

\begin{figure}[t]
    \centering
    \includegraphics[width=0.32\columnwidth]{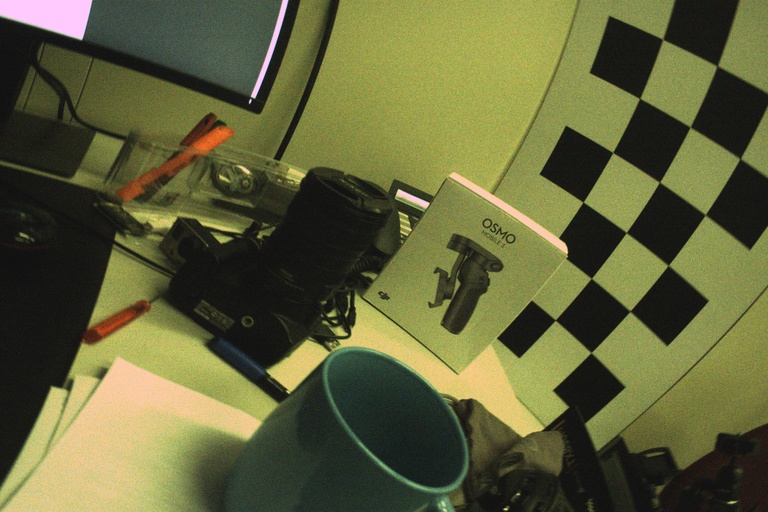}
    \includegraphics[width=0.32\columnwidth,angle=-180,origin=c]{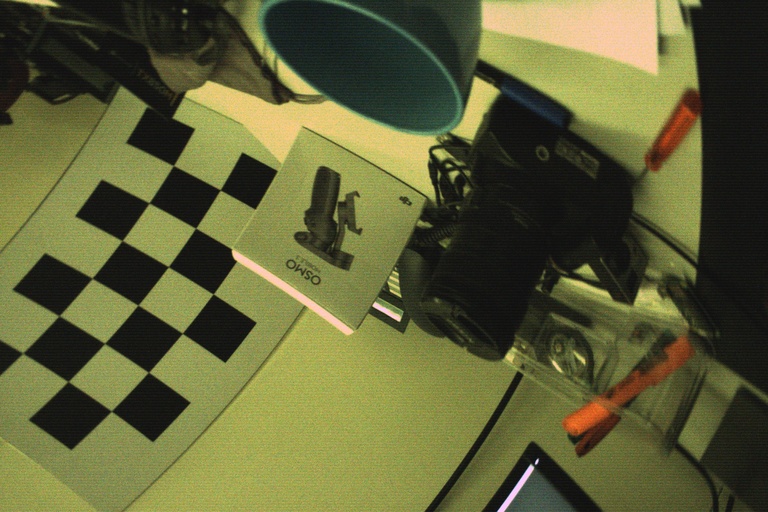}
    \includegraphics[width=0.32\columnwidth]{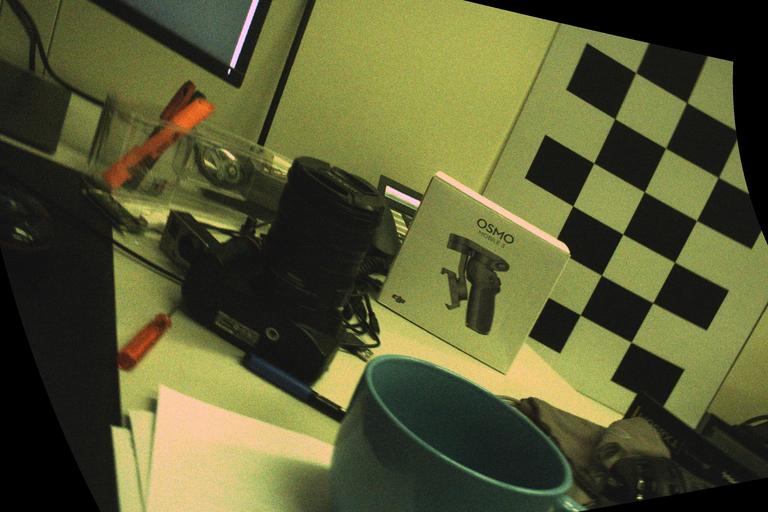}
            \vspace{-0.25em}
    \caption{Example of rotation undistortion ($\boldsymbol\omega$ solver) for close-range scenes. The distance to the nearest scene points is $\approx$10$\times$ the baseline.\vspace{0.75em}}
    \label{fig:close_range}
\end{figure}

A further experiment with real data supports the claim that the zero baseline assumption is insignificant in the case of rotational motion, unless most correspondences are on the closest scene parts.
Fig.~\ref{fig:close_range} shows rotational distortion removal with objects as close as 10$\times$ the baseline. As long as enough correspondences are found also on more distant scene parts, RANSAC chooses those and finds a mapping that is valid also for close objects. 
For translational motion, the closest correspondences carry more information about the translation than the distant ones. In our experiments the scenes always contained enough correspondences close enough to estimate the motion, but far enough to neglect the baseline. For scenes with mostly low depth (relative to the baseline) we recommend the 6pt fixed-baseline solver, described in the Appendix~\ref{sec:a_baseline_solver}.

\vspace{0.25em} \noindent \textbf{Real images.} We built a rig with two RS cameras, mounted $\approx$3$\,$cm apart and pointing in the same direction, see Fig.~\ref{fig:rig}. Their rolling shutters run in opposite directions, with $\approx$30$\,$ms readout time for a complete frame, a typical value for consumer devices. The image resolution is $3072\times2048$ pix. Additionally, we added a GS camera to the rig with resolution $1440\times1080$ pix (despite the weaker specs, that camera cost almost as much as the two RS cameras together). All cameras are triggered synchronously.

\begin{figure*}
    \centering
    \renewcommand{\tabcolsep}{0.5pt}
    \renewcommand{\arraystretch}{0.5}
    \begin{tabular}{ccccccc}
    RS input 1 & RS input 2 & OURS & GS ground truth & Undist. w.~\cite{rengarajan_bows_2016} & Undist. w.~\cite{Zhuang_2019_CVPR} & Undist. w.~\cite{lao_robust_2018}\\
         \includegraphics[width=0.295\columnwidth]{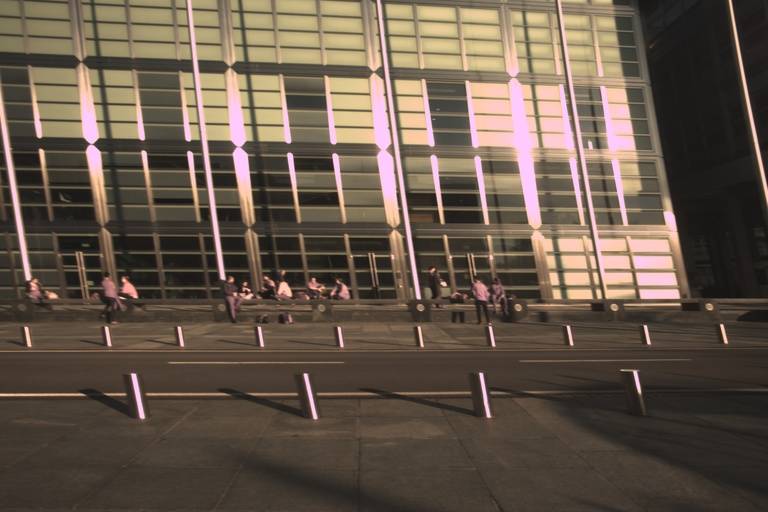} &
         \includegraphics[width=0.295\columnwidth,angle=-180,origin=c]{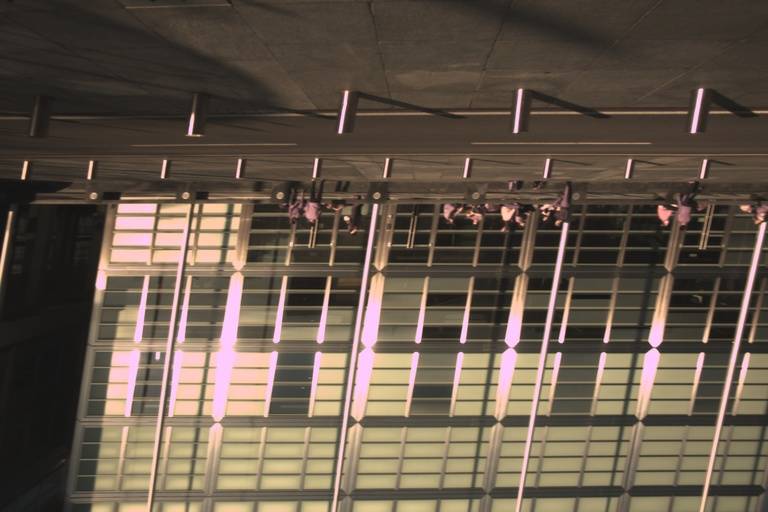} &
         \includegraphics[width=0.295\columnwidth]{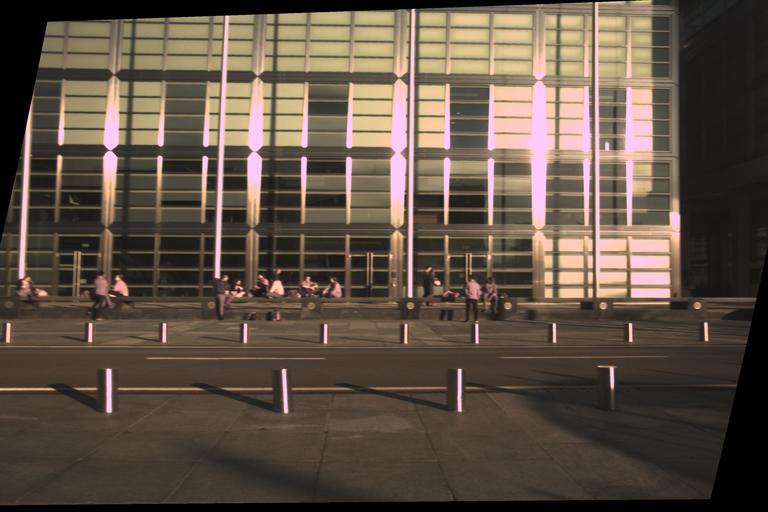} &
         \includegraphics[width=0.295\columnwidth]{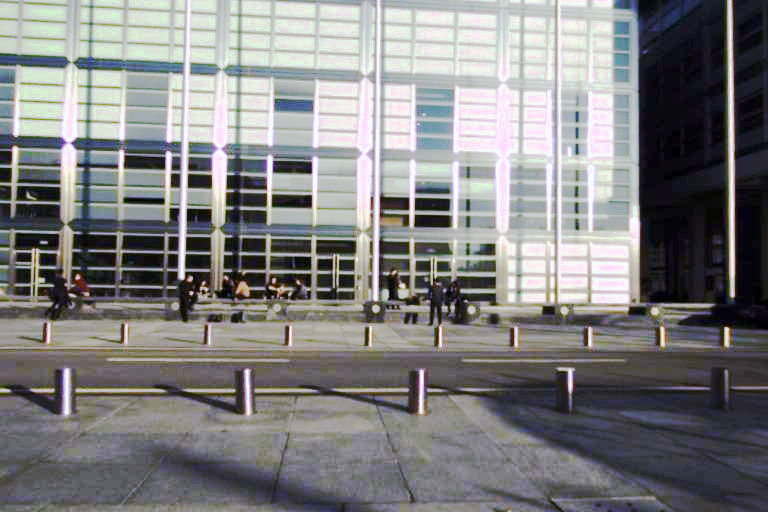} &
         \includegraphics[width=0.295\columnwidth]{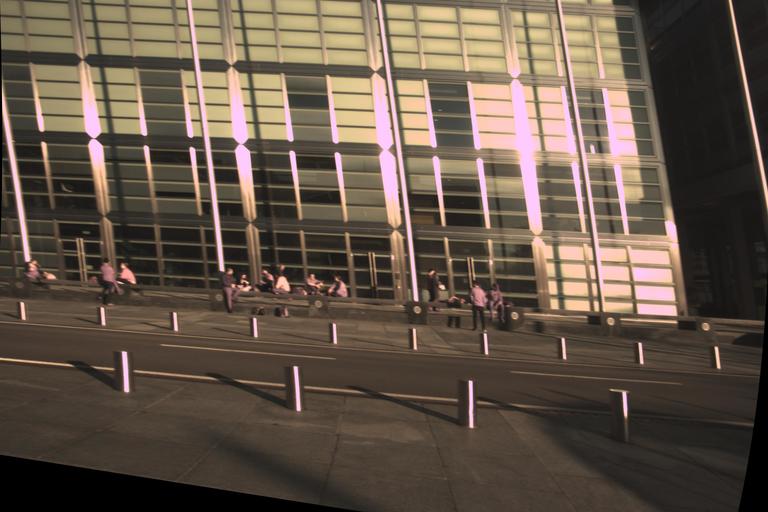} &
         \includegraphics[width=0.295\columnwidth]{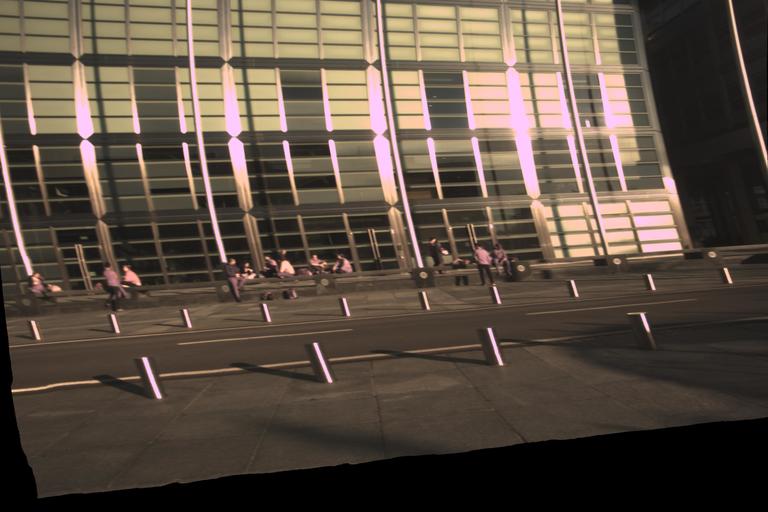} &
         \includegraphics[width=0.295\columnwidth]{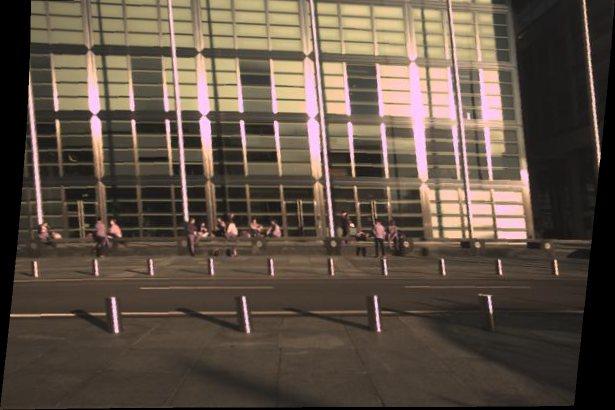} \\
         \includegraphics[width=0.295\columnwidth]{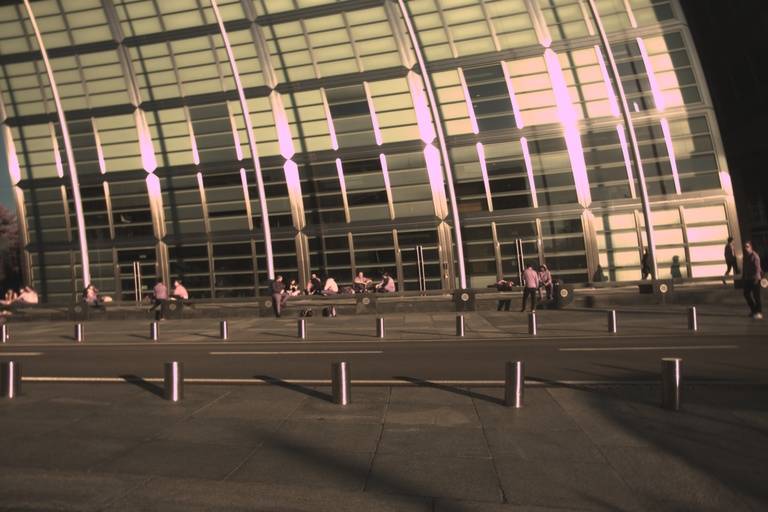} &
         \includegraphics[width=0.295\columnwidth,angle=-180,origin=c]{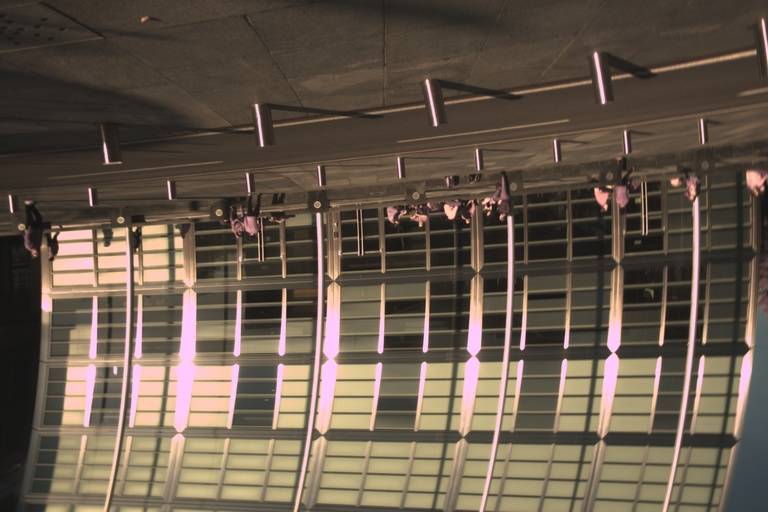} &
         \includegraphics[width=0.295\columnwidth]{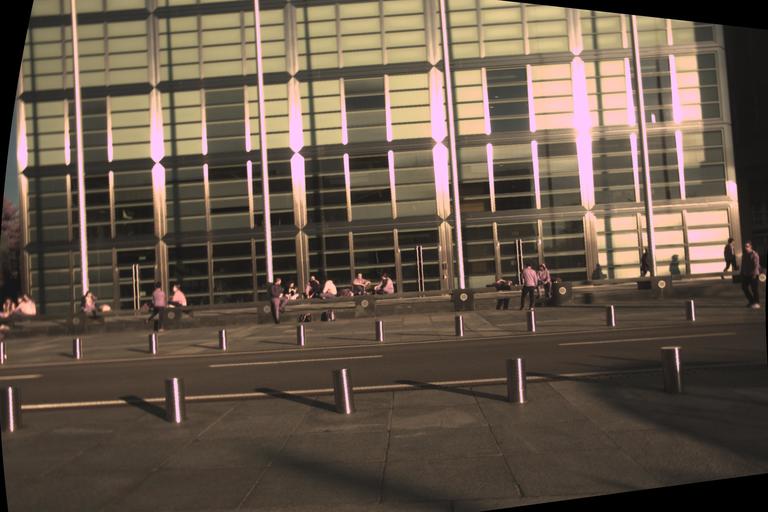} &
         \includegraphics[width=0.295\columnwidth]{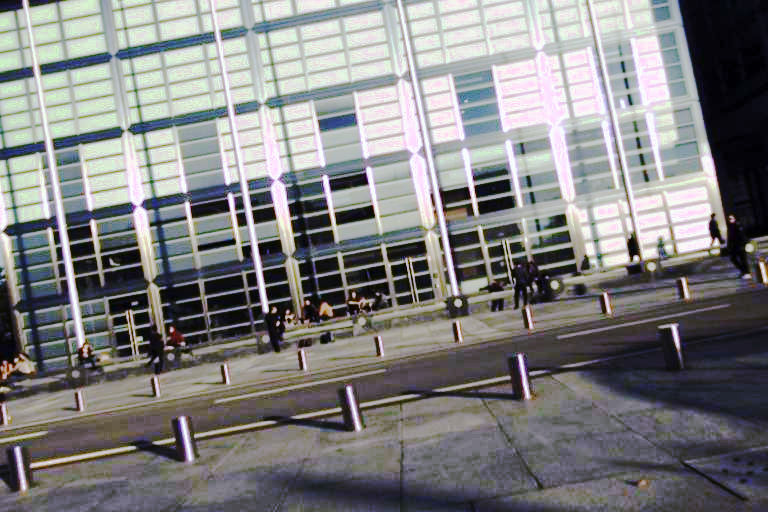} &
         \includegraphics[width=0.295\columnwidth]{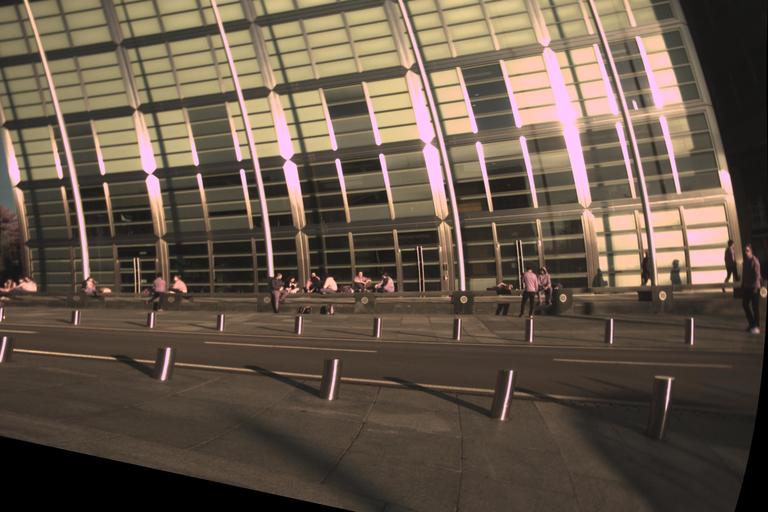} &
         \includegraphics[width=0.295\columnwidth]{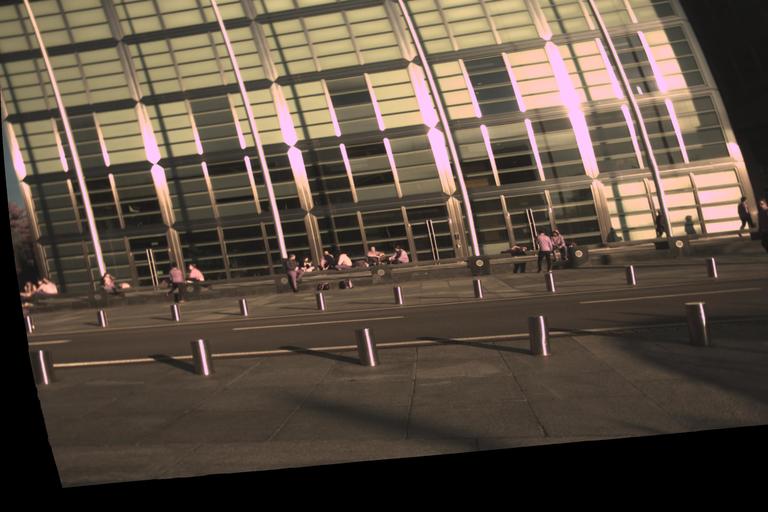} &
         \includegraphics[width=0.295\columnwidth]{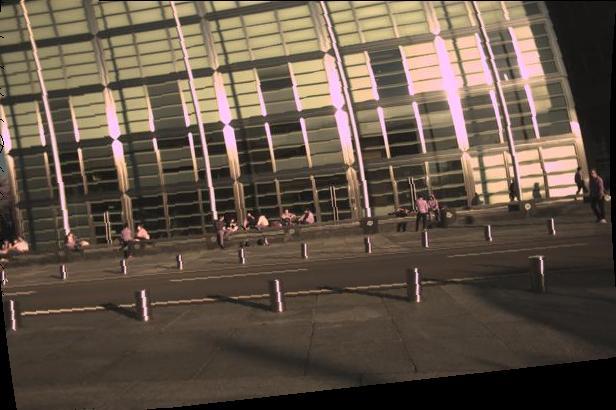} \\
         \includegraphics[width=0.295\columnwidth]{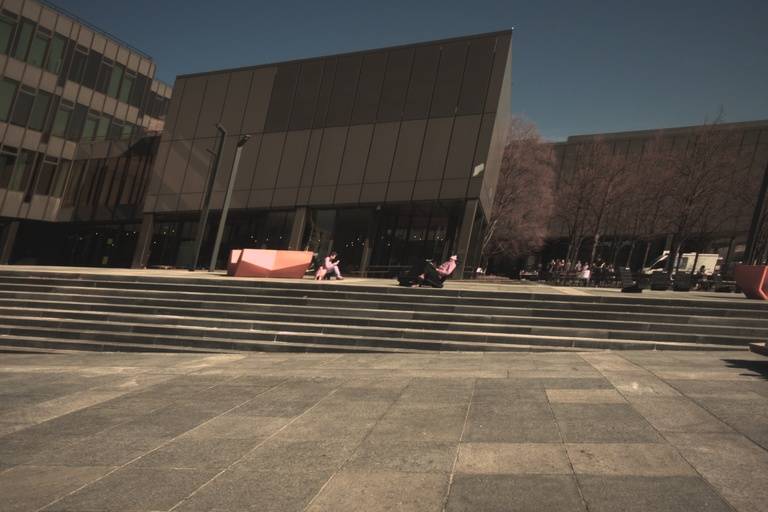} &
         \includegraphics[width=0.295\columnwidth,angle=-180,origin=c]{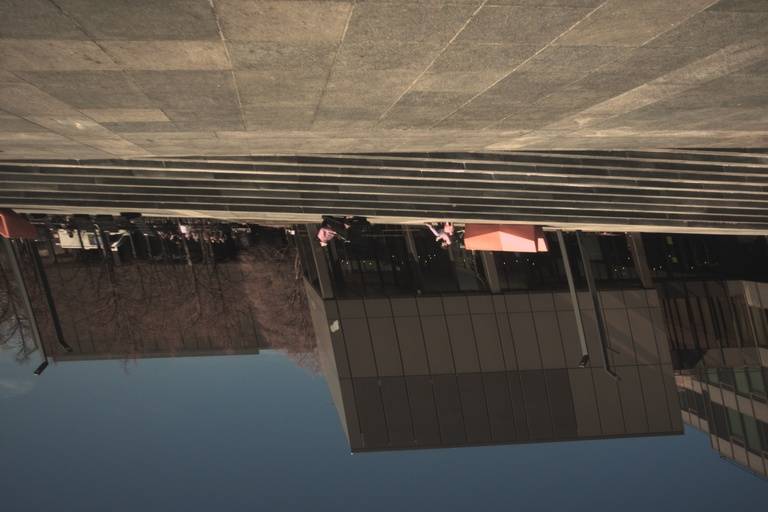} &
         \includegraphics[width=0.295\columnwidth]{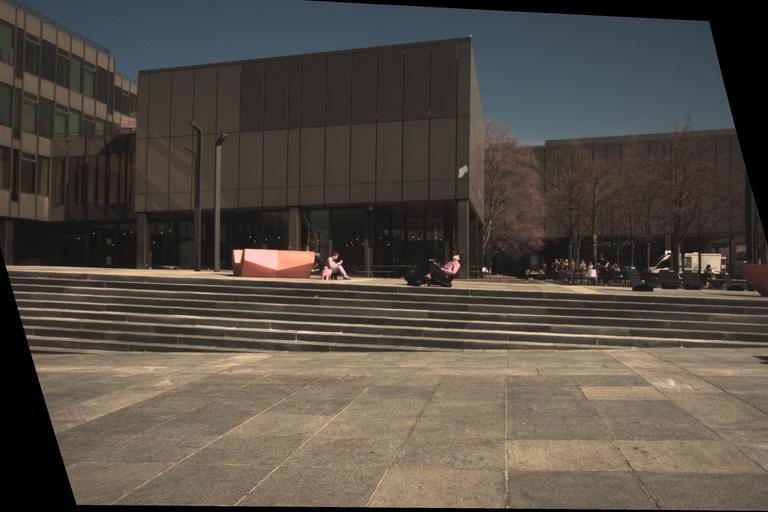} &
         \includegraphics[width=0.295\columnwidth]{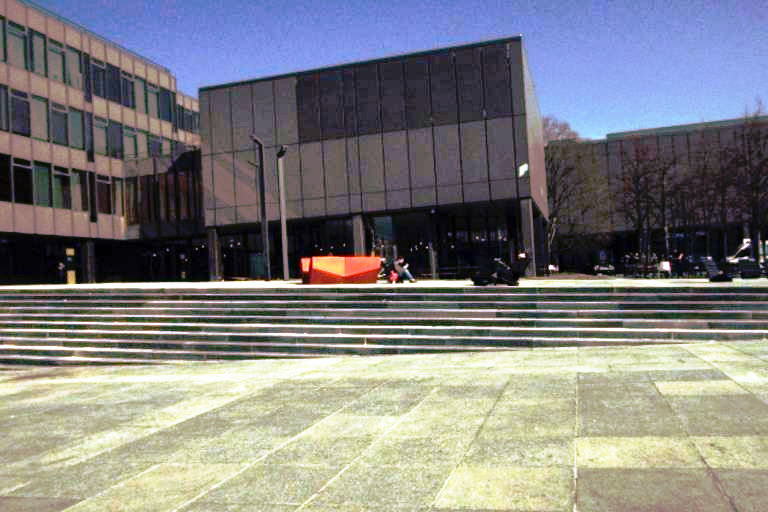} &
         \includegraphics[width=0.295\columnwidth]{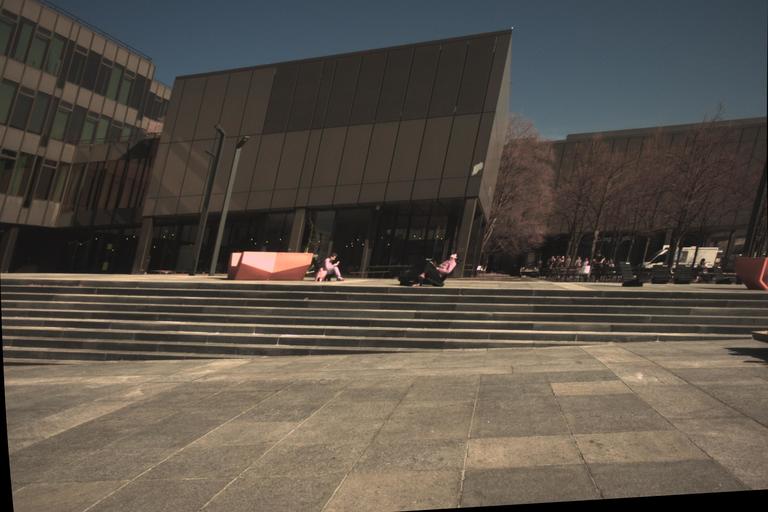} &
         \includegraphics[width=0.295\columnwidth]{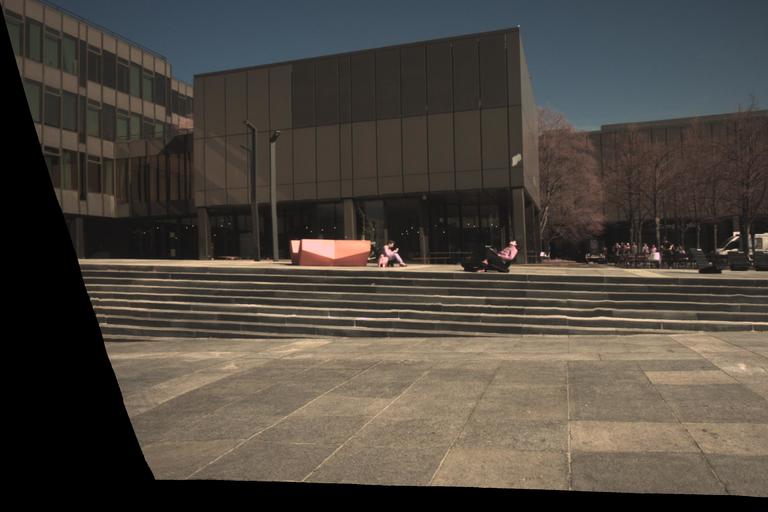} &
         \includegraphics[width=0.295\columnwidth]{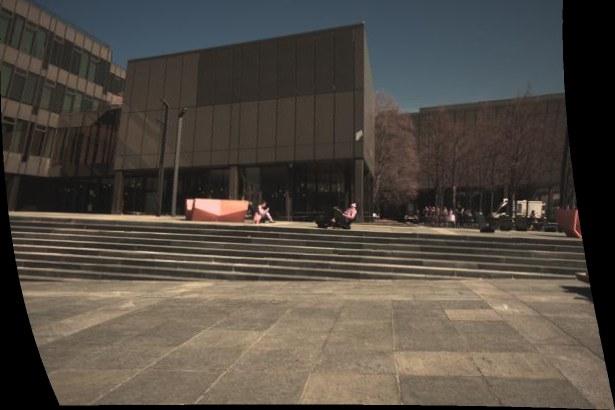} \\
         \includegraphics[width=0.295\columnwidth]{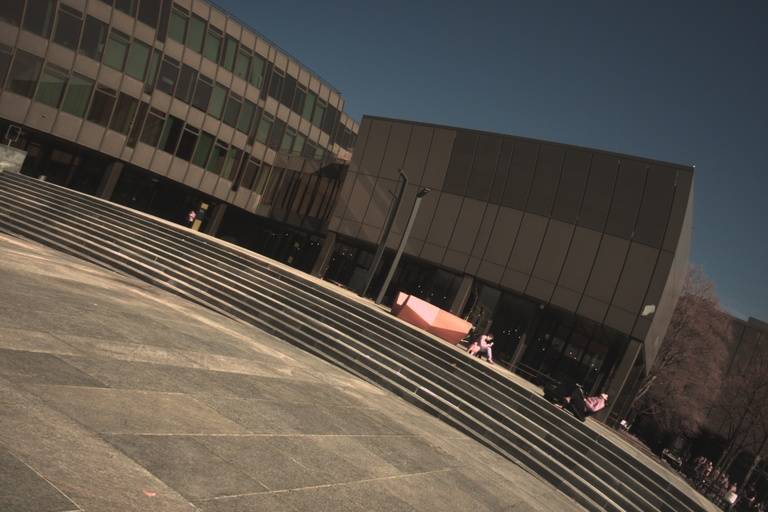} &
         \includegraphics[width=0.295\columnwidth,angle=-180,origin=c]{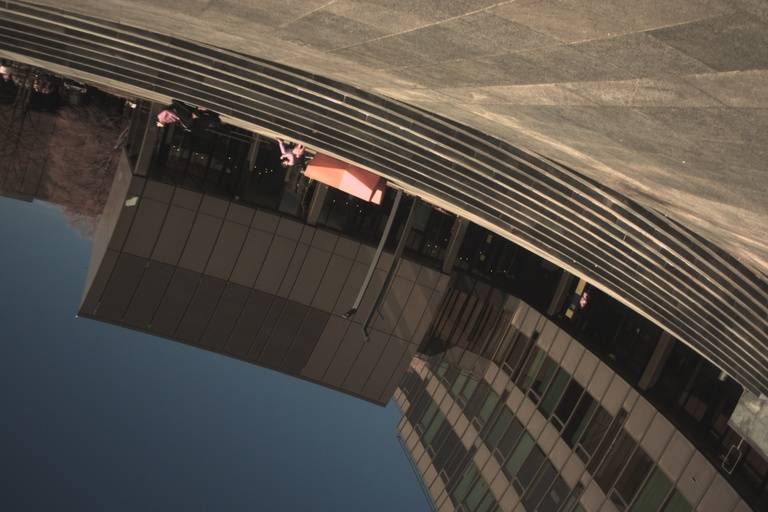} &
         \includegraphics[width=0.295\columnwidth]{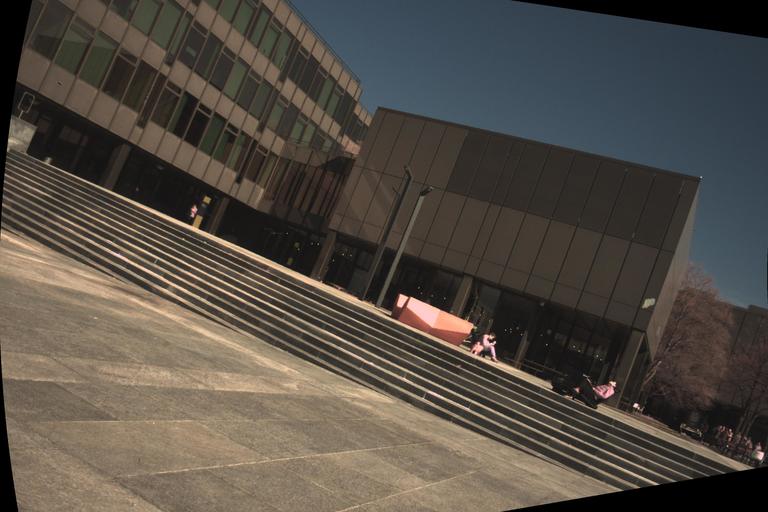} &
         \includegraphics[width=0.295\columnwidth]{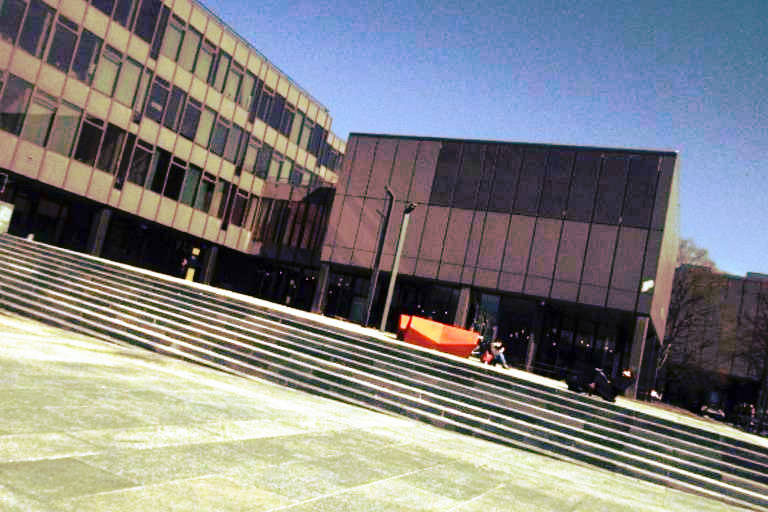} &
         \includegraphics[width=0.295\columnwidth]{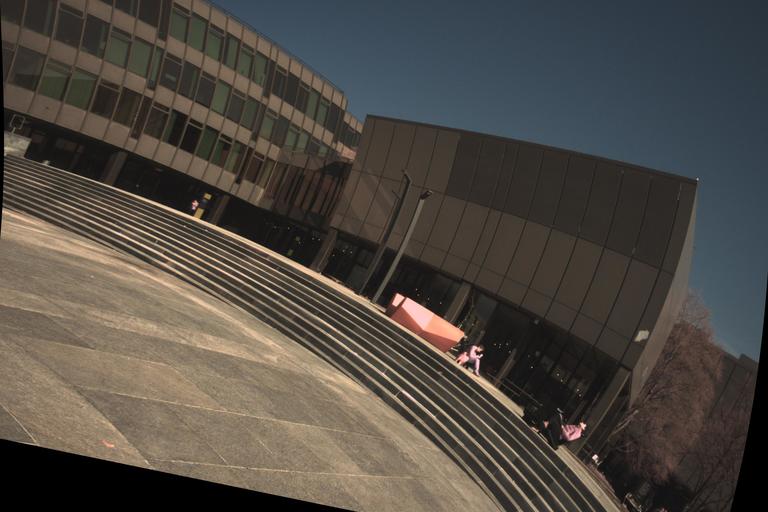} &
         \includegraphics[width=0.295\columnwidth]{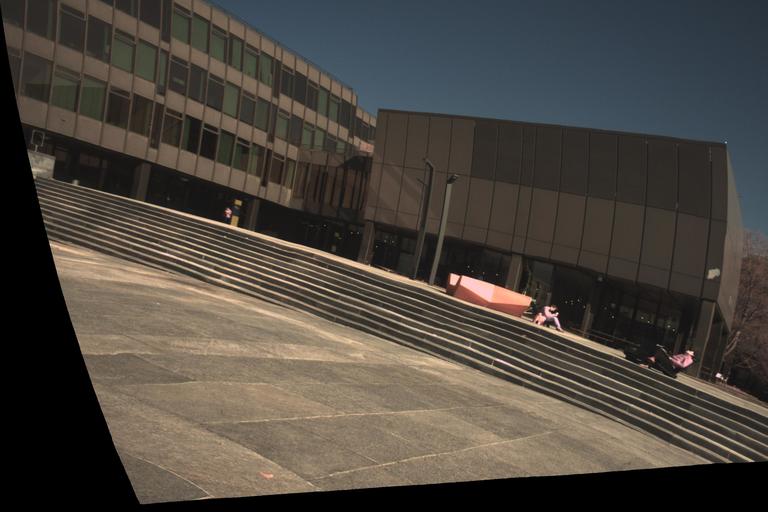} &
         \includegraphics[width=0.295\columnwidth]{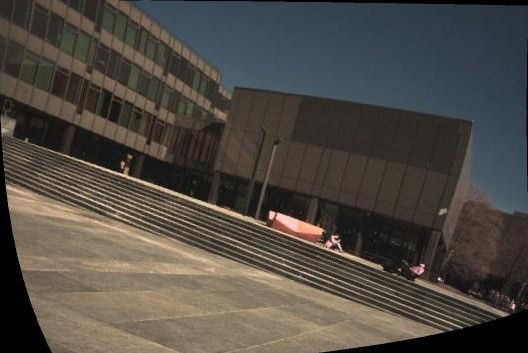} \\
         \includegraphics[width=0.295\columnwidth]{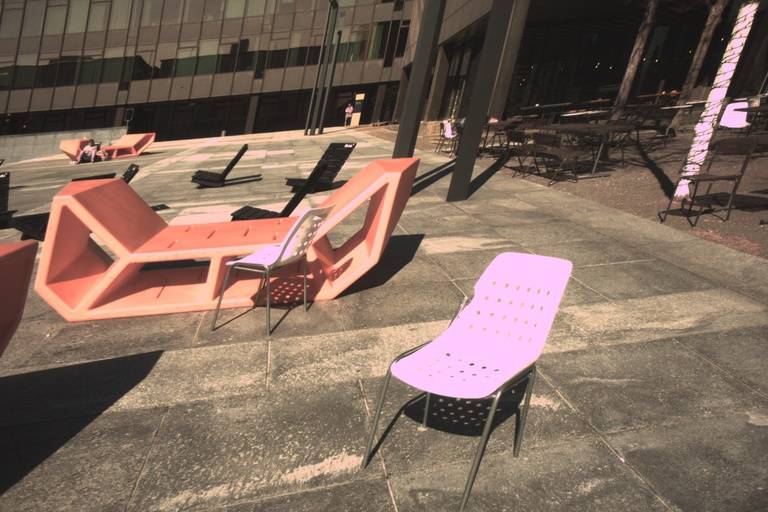} &
         \includegraphics[width=0.295\columnwidth,angle=-180,origin=c]{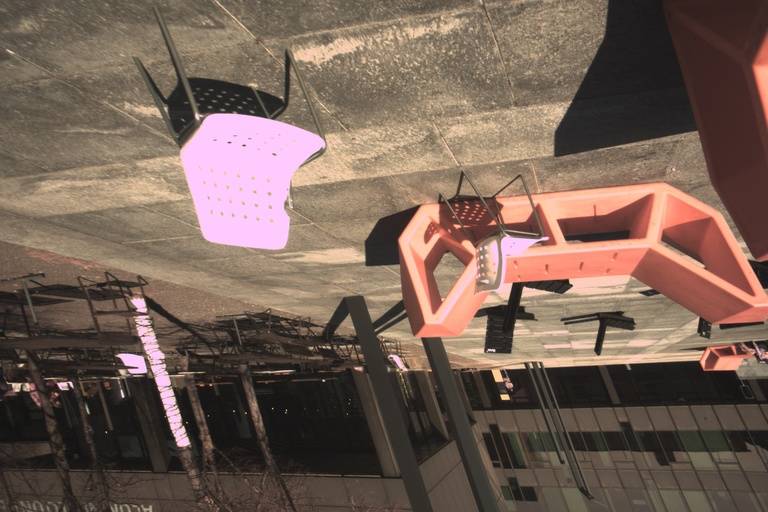} &
         \includegraphics[width=0.295\columnwidth]{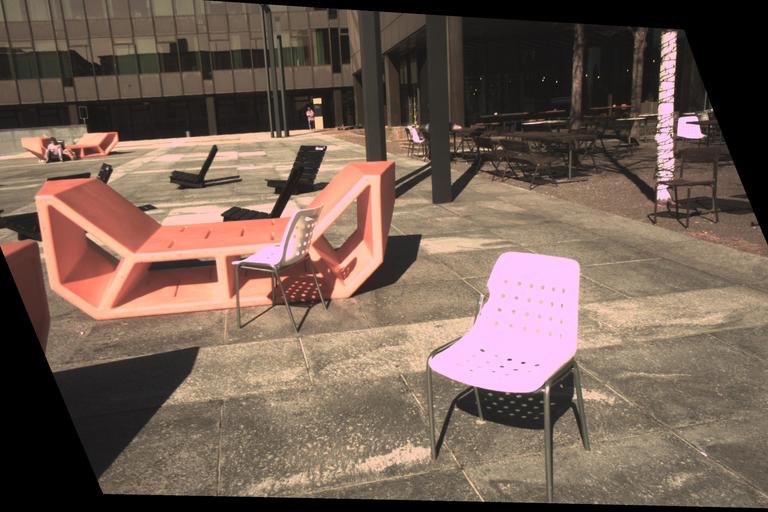} &
         \includegraphics[width=0.295\columnwidth]{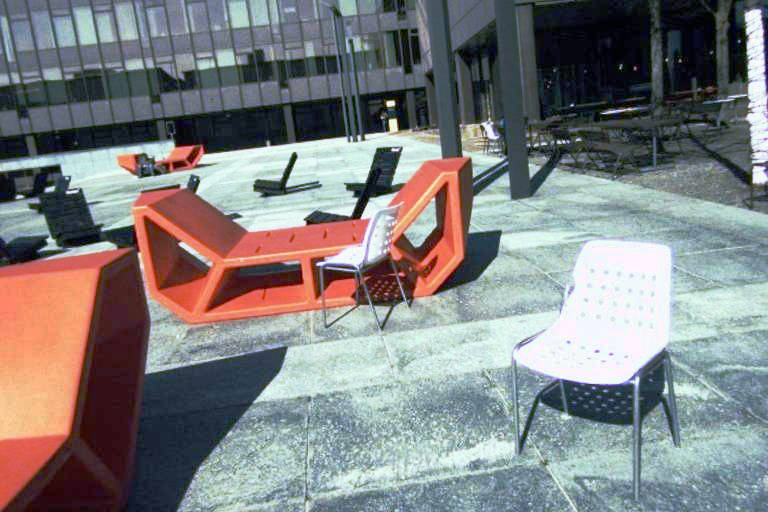} &
         \includegraphics[width=0.295\columnwidth]{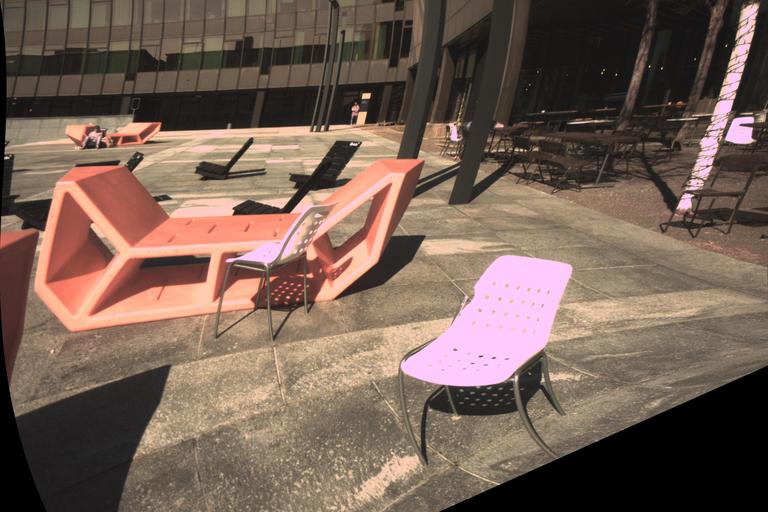} &
         \includegraphics[width=0.295\columnwidth]{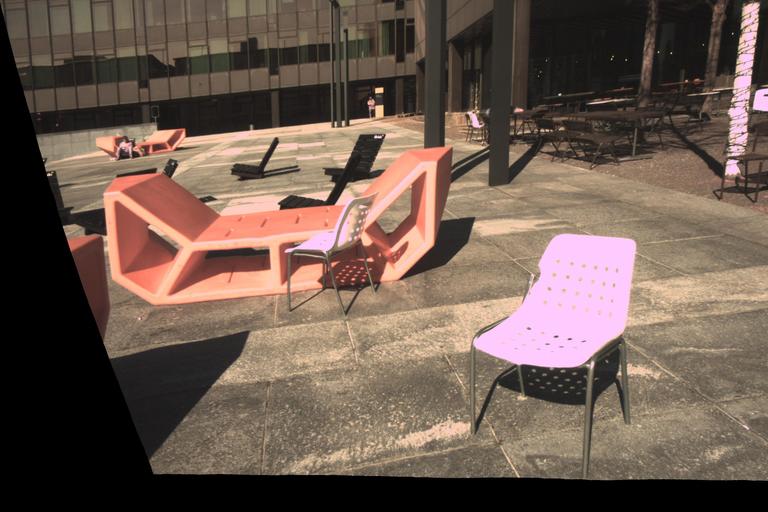} &
         \includegraphics[width=0.295\columnwidth]{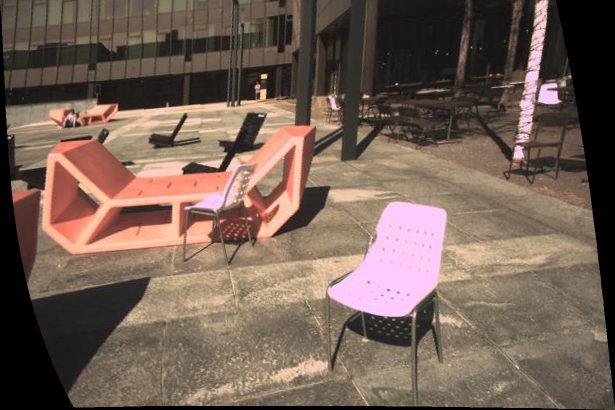} \\
         \includegraphics[width=0.295\columnwidth]{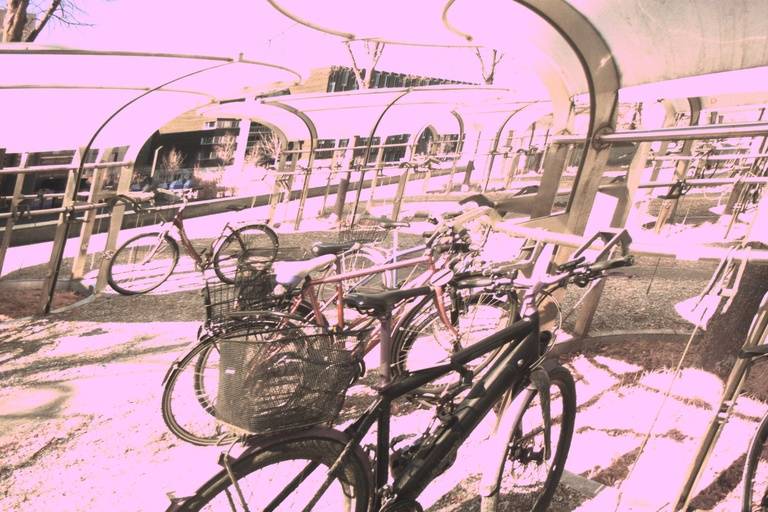} &
         \includegraphics[width=0.295\columnwidth,angle=-180,origin=c]{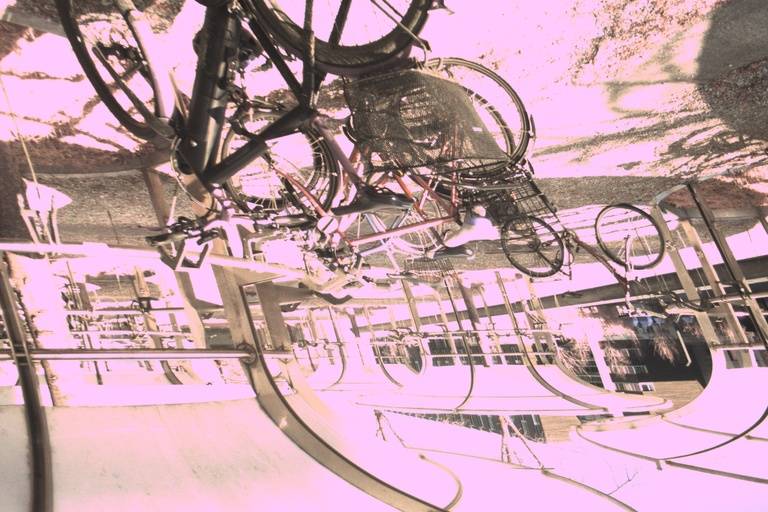} &
         \includegraphics[width=0.295\columnwidth]{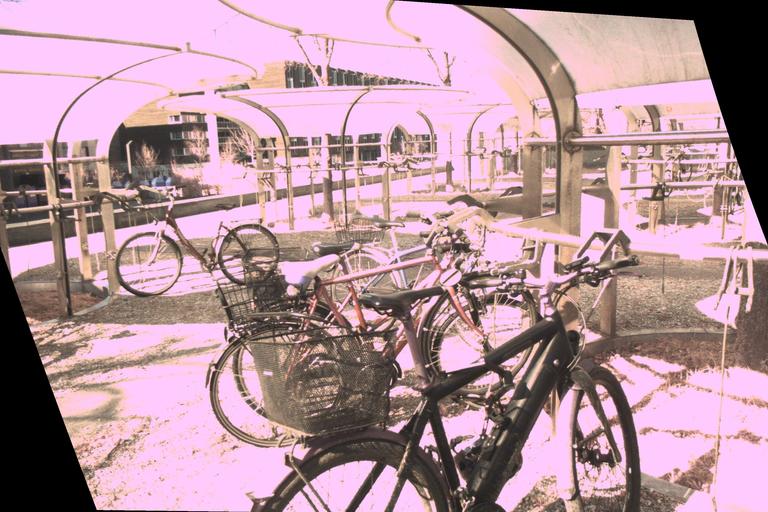} &
         \includegraphics[width=0.295\columnwidth]{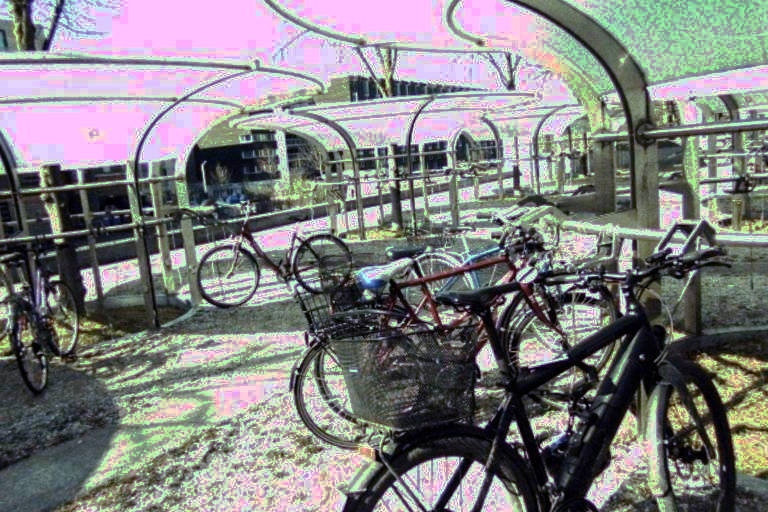} &
         \includegraphics[width=0.295\columnwidth]{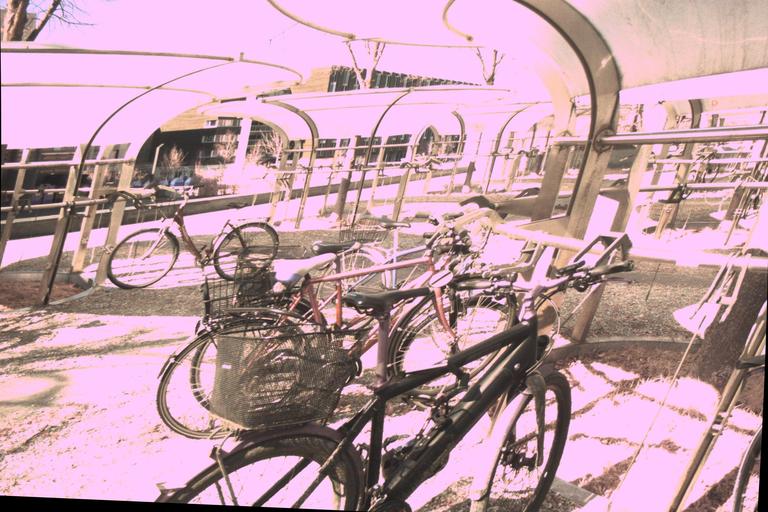} &
         \includegraphics[width=0.295\columnwidth]{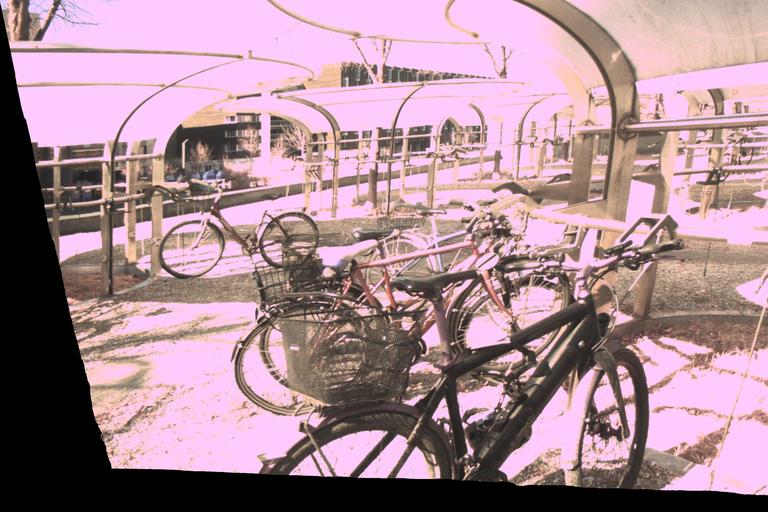} &
         \includegraphics[width=0.295\columnwidth]{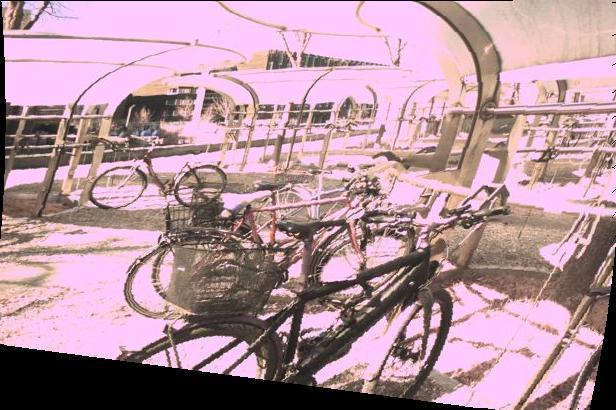} 
    \end{tabular}
        \vspace{-0.625em}
    \caption{ Camera undergoing a rotational motion. Significant RS distortion can be successfully removed using the proposed camera setup. Competing methods for RS image correction~\cite{rengarajan_bows_2016,rengarajan_unrolling_2017,lao_robust_2018} provide visibly worse results.
    \vspace{.75em}}
    % The fourth column shows the result after estimating the rotation $\omega$ and undistorting the input RS image. The sixth column shows undistortion using $\omega$ estimated only from RS input 1 and the cropped image in the third column. This simulates the case when one camera is $2\times$ zoomed, a typical case in smartphones. 
    % Note that the difference in colors between RS and GS images is caused by the RS cameras lenses being without infrared filter.}
    \label{fig:real_images_rot}
\end{figure*}

\begin{figure*}
    \centering
    \renewcommand{\tabcolsep}{0.2pt}
    \renewcommand{\arraystretch}{0.2}
    \begin{tabular}{ccccccc}
    RS input 1 & RS input 2 & OURS & GS ground truth & Undist.\ w.~\cite{rengarajan_bows_2016} & Undist.\ w.~\cite{zhuang_rolling-shutter-aware_2017} & Undist.\ w.~\cite{vasu_occlusion-aware_2018}\\
         \includegraphics[width=0.295\columnwidth]{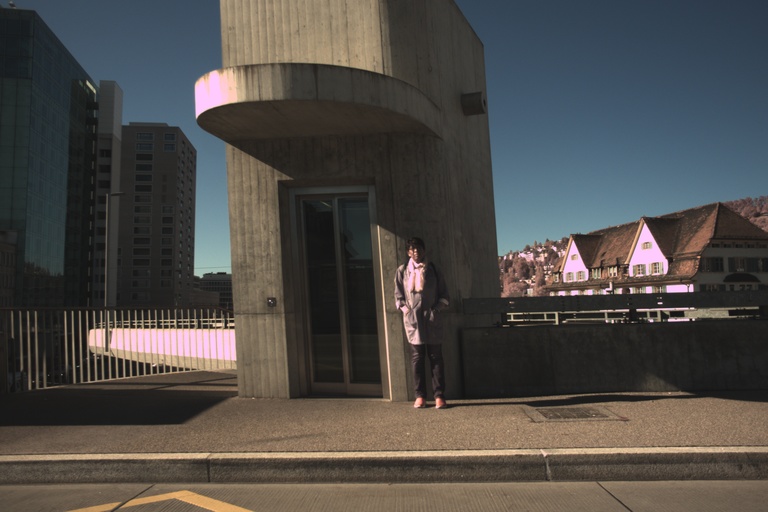} &
         \includegraphics[width=0.295\columnwidth,angle=-180,origin=c]{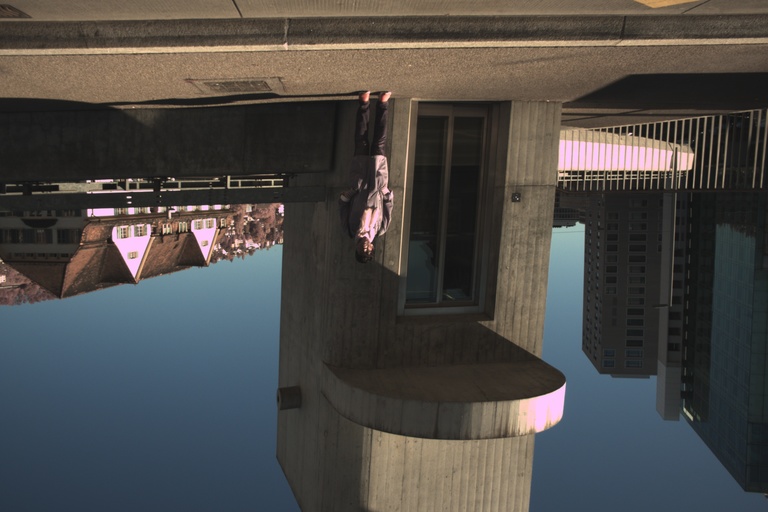} &
         \includegraphics[width=0.295\columnwidth]{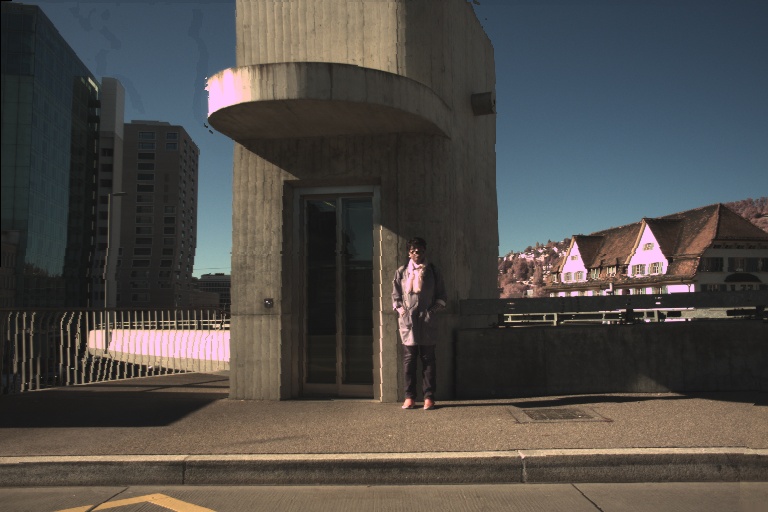} &
         \includegraphics[width=0.295\columnwidth]{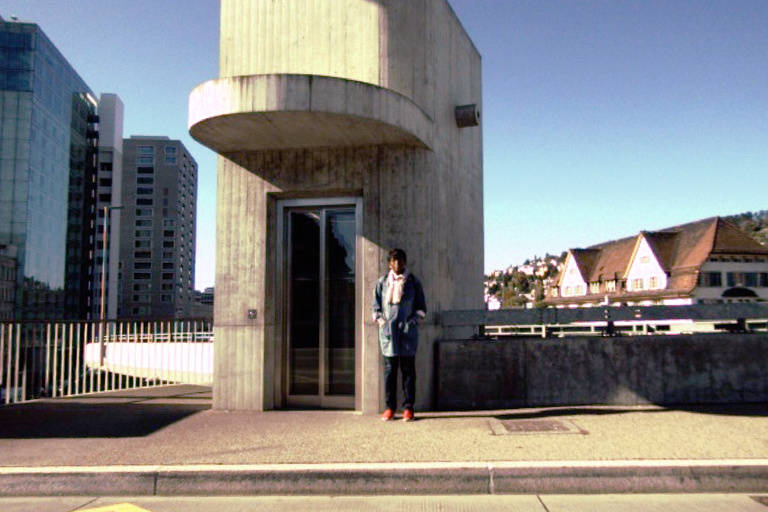} &
         \includegraphics[width=0.295\columnwidth]{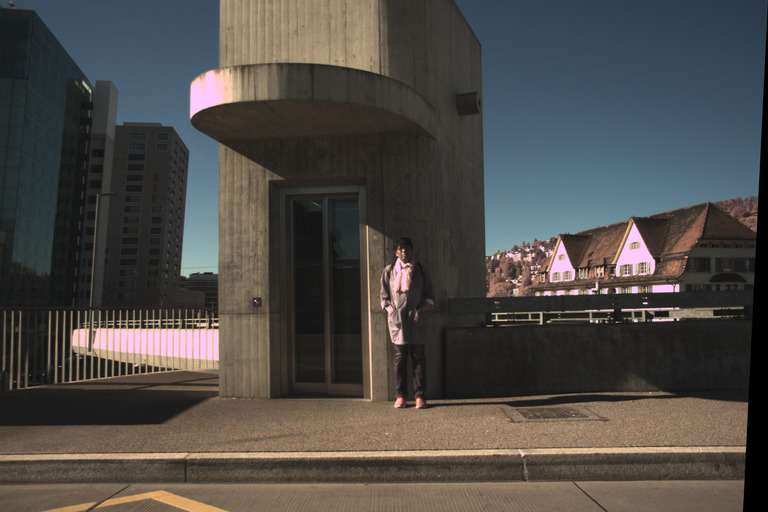} &
         \includegraphics[width=0.295\columnwidth]{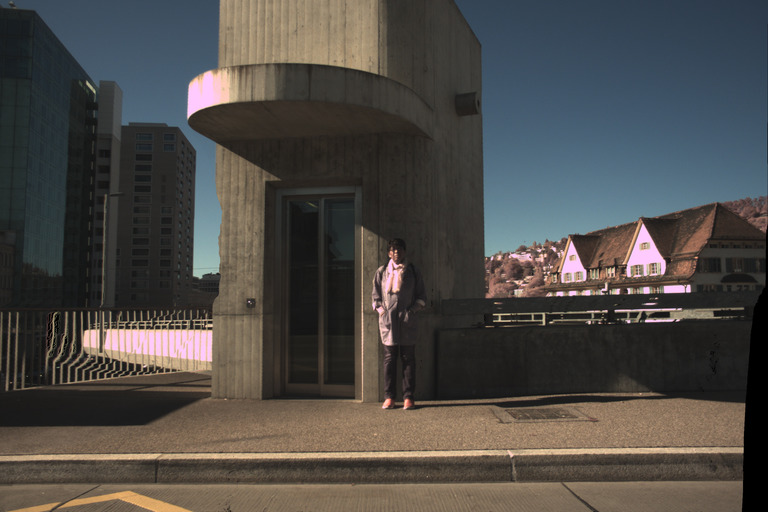} &
         \includegraphics[width=0.295\columnwidth]{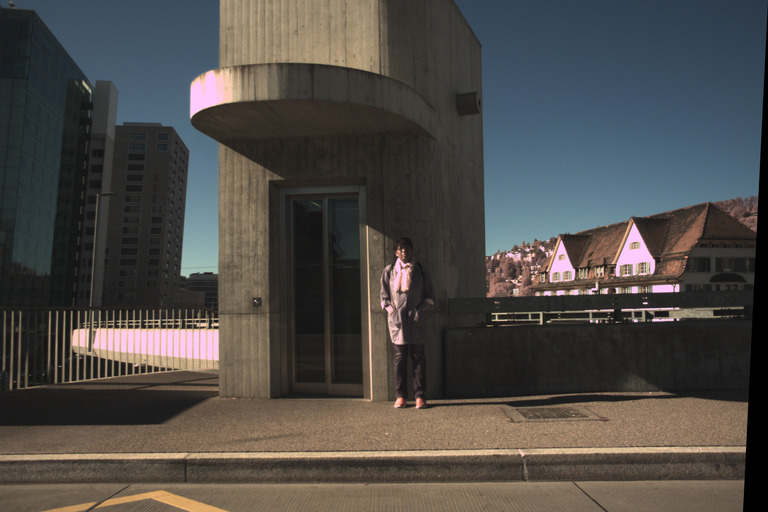} \\
         \includegraphics[width=0.295\columnwidth]{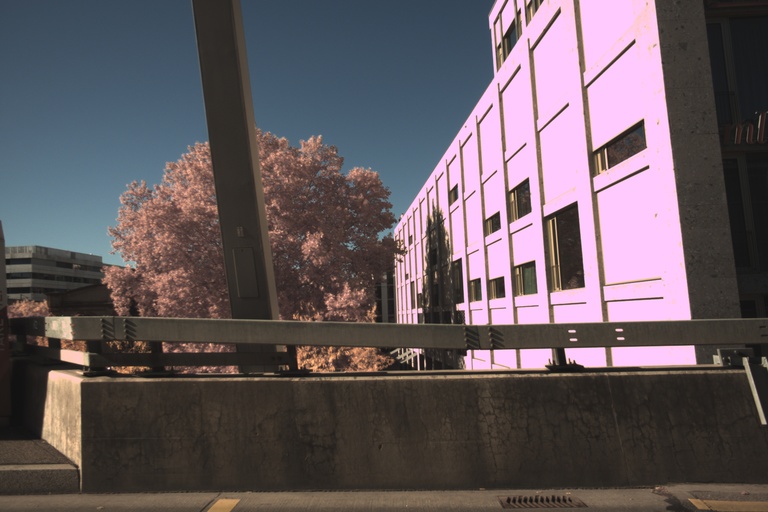} &
         \includegraphics[width=0.295\columnwidth,angle=-180,origin=c]{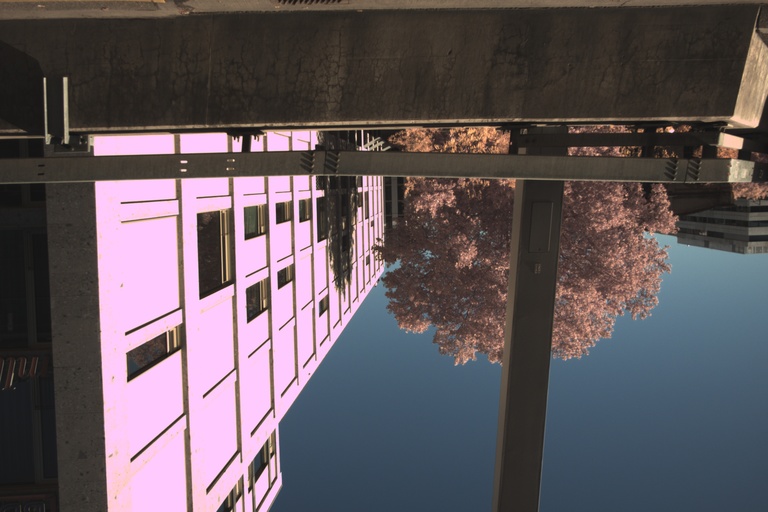} &
         \includegraphics[width=0.295\columnwidth]{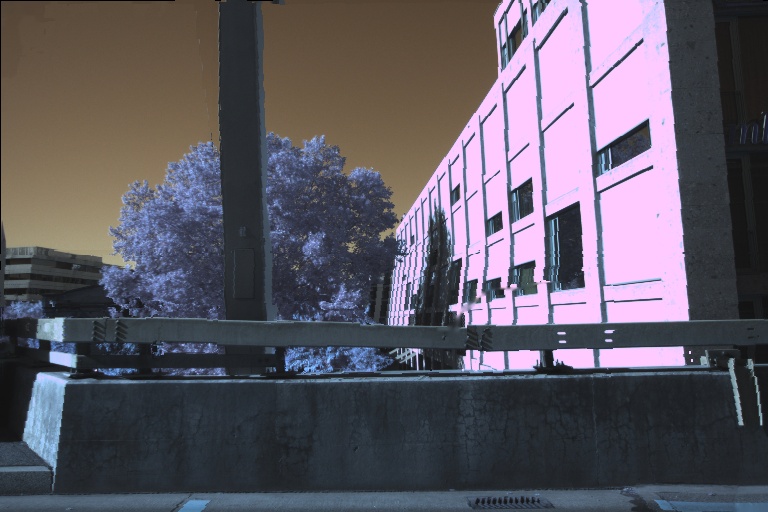} &
         \includegraphics[width=0.295\columnwidth]{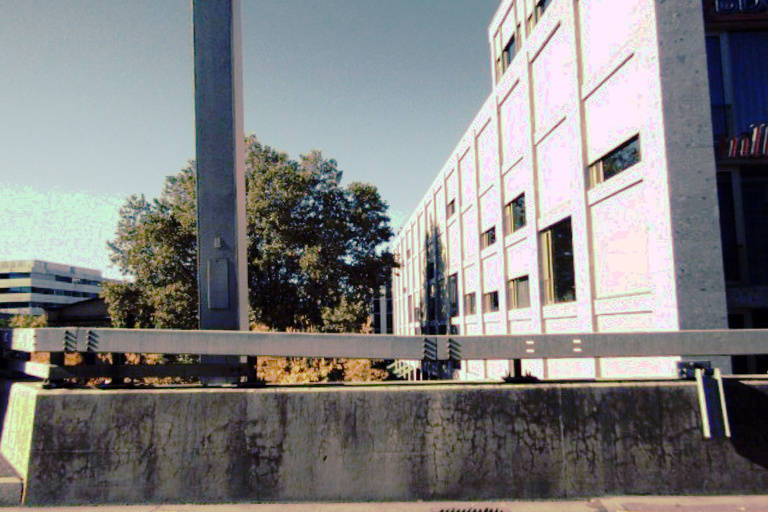} &
         \includegraphics[width=0.295\columnwidth]{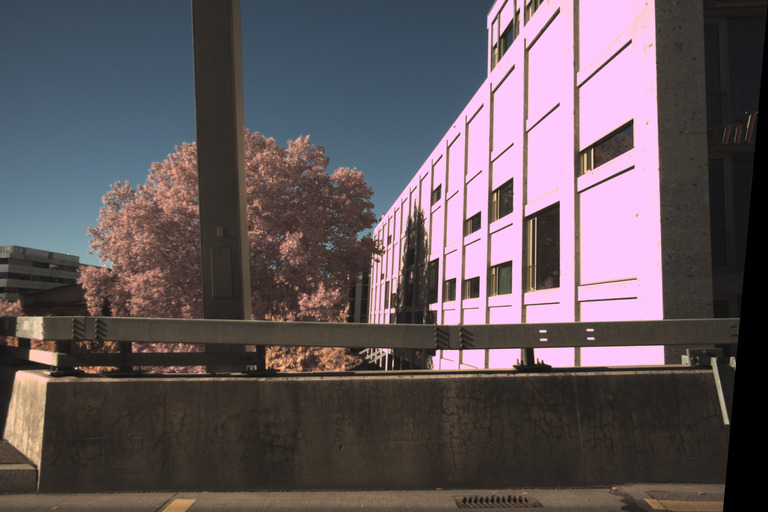} &
         \includegraphics[width=0.295\columnwidth]{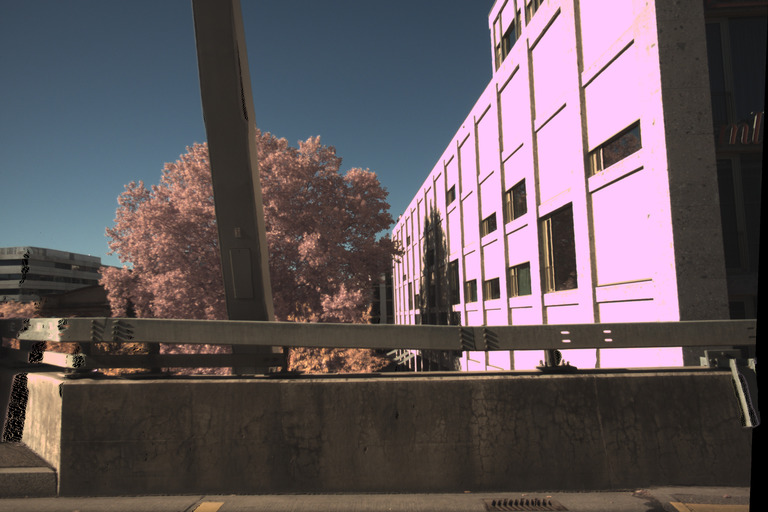} &
         \includegraphics[width=0.295\columnwidth]{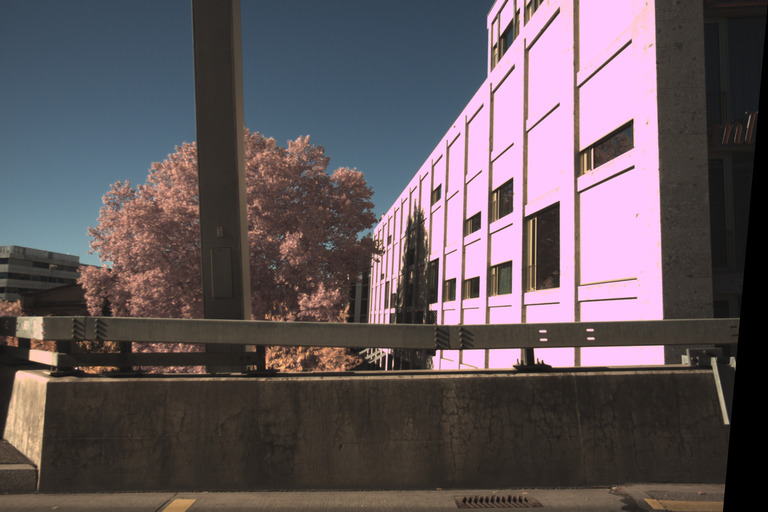} \\
         \includegraphics[width=0.295\columnwidth]{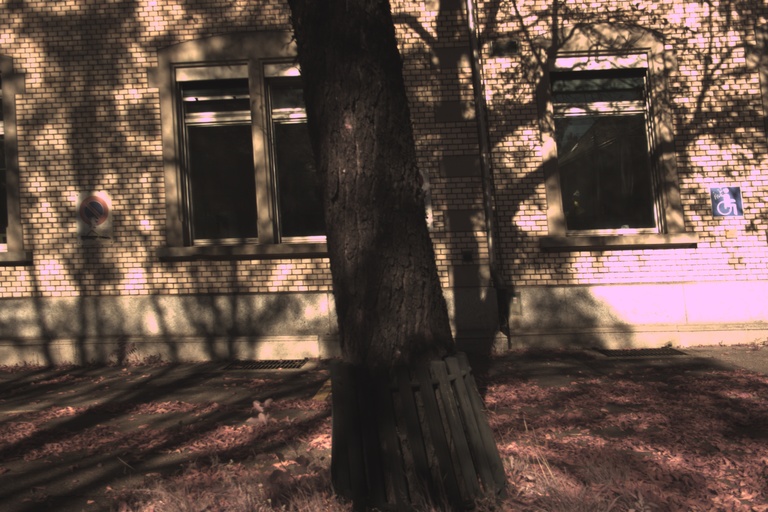} &
         \includegraphics[width=0.295\columnwidth,angle=-180,origin=c]{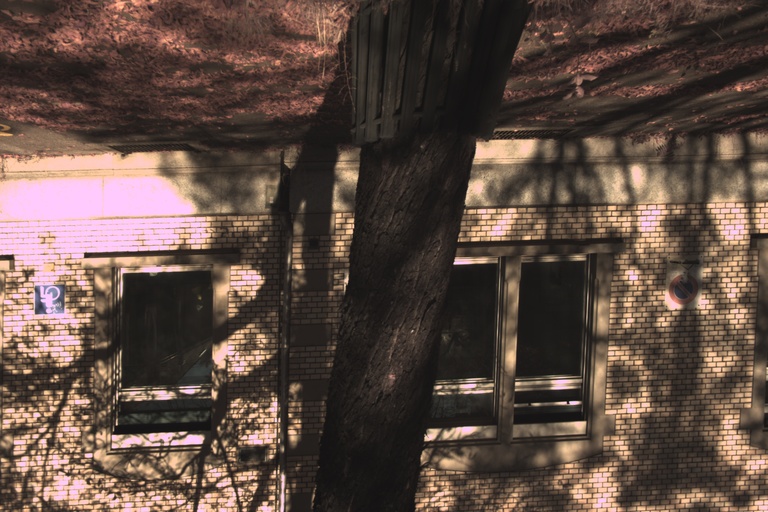} &
         \includegraphics[width=0.295\columnwidth]{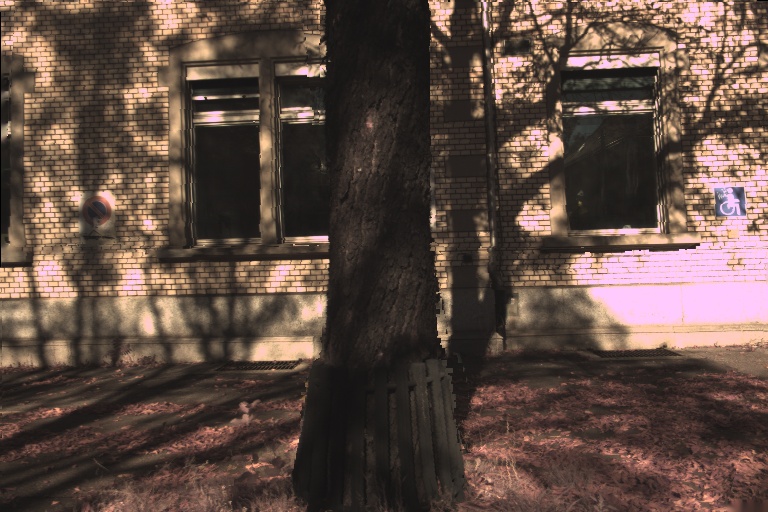} &
         \includegraphics[width=0.295\columnwidth]{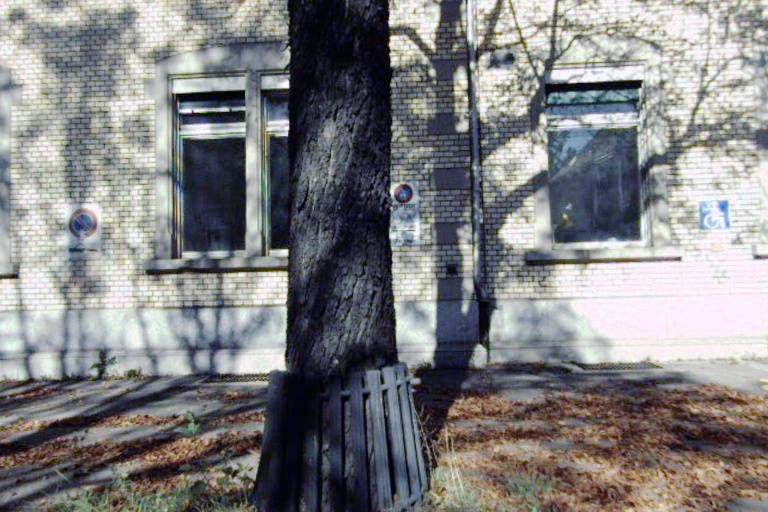} &
         \includegraphics[width=0.295\columnwidth]{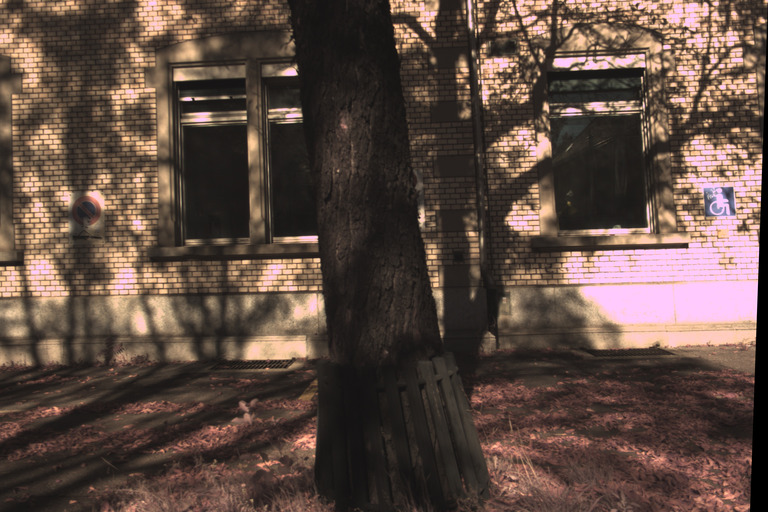} &
         \includegraphics[width=0.295\columnwidth]{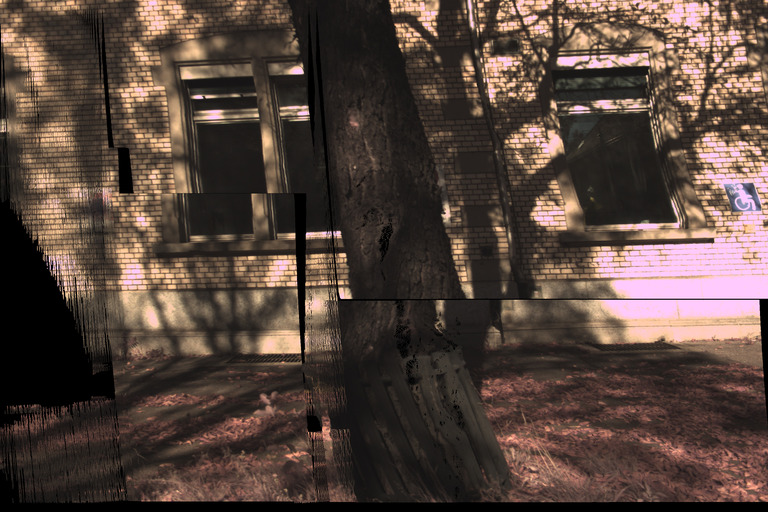} &
         \includegraphics[width=0.295\columnwidth]{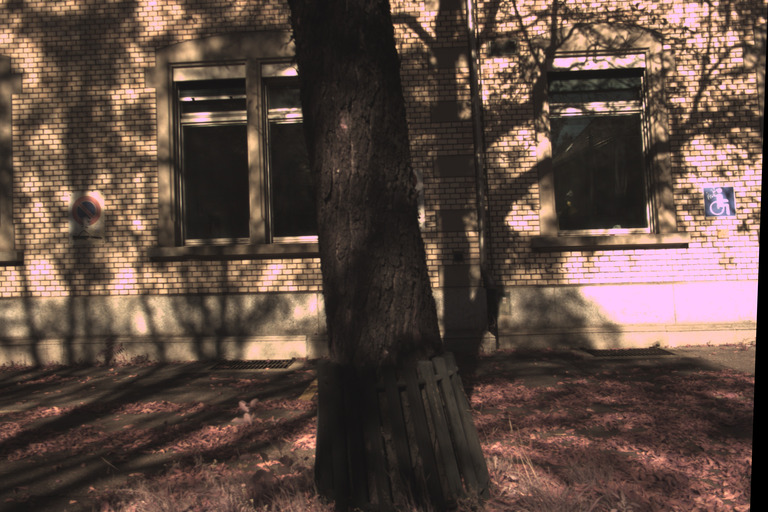} \\
    \end{tabular}
    \vspace{-0.625em}
    \caption{Camera undergoing translational motion. Images are undistorted pixel-wise, using optical flow obtained with~\cite{sun_pwc-net:_2017}.}
    % The fourth column shows the result after estimating the rotation $\omega$ and undistorting the input RS image. The sixth column shows undistortion using $\omega$ estimated only from RS input 1 and the cropped image in the third column. This simulates the case when one camera is $2\times$ zoomed, a typical case in smartphones. 
    % Note that the difference in colors between RS and GS images is caused by the RS cameras lenses being without infrared filter.}
    \label{fig:real_images_tr}
\end{figure*}

We captured images of various scenes, with significant RS distortions from a variety of different camera motions. The angular velocity in the rotation experiments was between 8 and 15 degrees per frame or 240-450$^\circ$/s. In the translation experiments, the car was moving with 30-50 km/h and, since the camera was hand-held, there was also a non-negligible rotational component. The correspondences between images were either matched SIFT features, or taken from the optical flow in the case of translational motion, see Sec.~\ref{sec:undist}. Although the proposed camera configuration is new and there are no other algorithms suited to handle such data, it is interesting to see the results of existing RS undistortion algorithms. See examples in Fig.~\ref{fig:real_images_rot} for rotation and Fig.~\ref{fig:real_images_tr} for translation, where we compare our undistorted GS images with those of most recent competing methods~\cite{rengarajan_bows_2016,rengarajan_unrolling_2017,lao_robust_2018,vasu_occlusion-aware_2018,zhuang_rolling-shutter-aware_2017}. 

We used RANSAC with a fixed number of 200 iterations, which proved to be enough due to the small minimal set for the $\omega$ solver, respectively the dense optical flow with low outlier fraction for the $\omega t$ solver.
Note that we compare each method only in relevant scenarios, e.g.,~\cite{rengarajan_unrolling_2017,lao_robust_2018}  work under the assumption of pure rotation and therefore are not able to handle translation properly;~\cite{zhuang_rolling-shutter-aware_2017} requires a baseline and thus handles only translation;~\cite{rengarajan_bows_2016,vasu_occlusion-aware_2018,Zhuang_2019_CVPR} should be able to handle both cases, the results of~\cite{vasu_occlusion-aware_2018,rengarajan_unrolling_2017} were unsatisfactory for rotation, so we do not present them.

Compared to existing methods, our results demonstrate robustness to the motion type as well as the scene content. The proposed camera setup allows us to correct various types of distortion with small residual error compared to the corresponding GS image. For rotation~\cite{lao_robust_2018} in some cases provides satisfactory results (Fig.\ref{fig:real_images_rot}, rows 1 and 3), but it fails when there are too few straight lines in the scene (row 6). ~\cite{rengarajan_bows_2016} almost always fails and~\cite{Zhuang_2019_CVPR}, although trained on 6DOF motion data, only produced usable result in rows 3,5 and 6 with rotation around the $y$-axis. 

\begin{figure}[t]
    \centering
    \includegraphics[width=0.32\columnwidth]{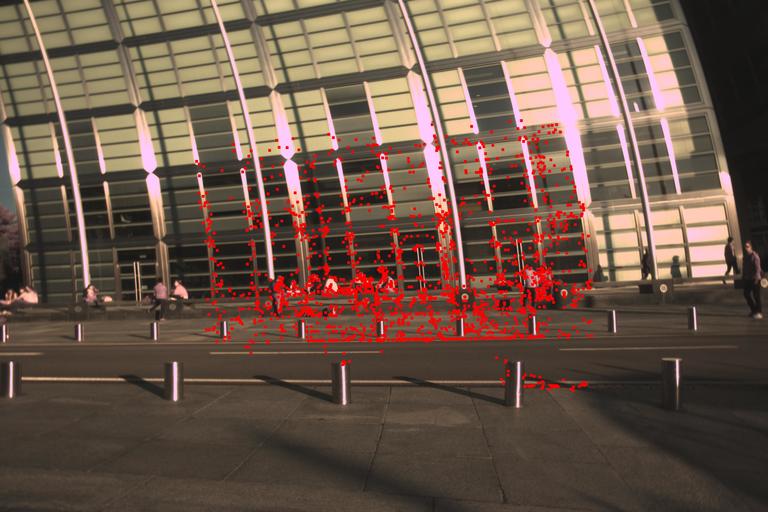}
    \includegraphics[width=0.32\columnwidth,angle=-180,origin=c]{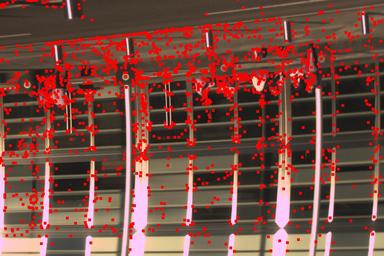}
    \includegraphics[width=0.32\columnwidth]{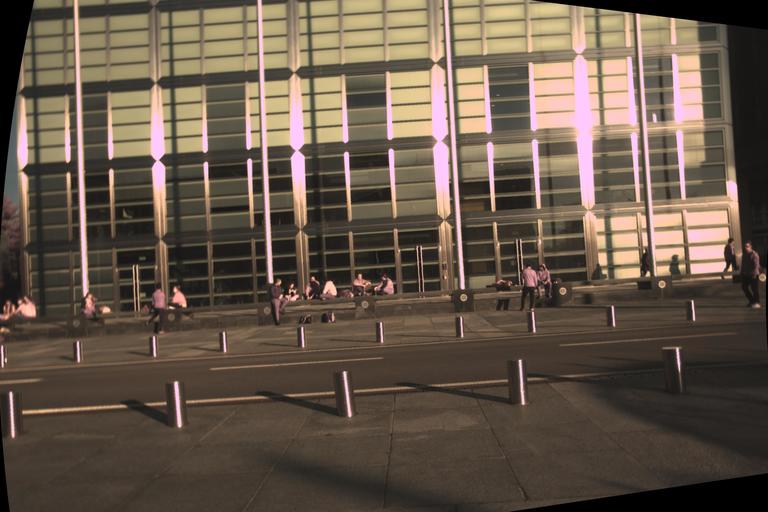}
            \vspace{-0.25em}
    \caption{Example of rotation undistortion ($\boldsymbol\omega$ solver). Correspondences are displayed in red. One camera has narrower FOV. Although parts of the wide-angle view have no correspondences, they are undistorted correctly.}
    \label{fig:crop}
\end{figure}

In Fig.~\ref{fig:eyecatcher} we show a sample with very strong, real RS effect. Even in this situation our method produces a near-perfect GS image, whereas competing methods fail. Furthermore, in Fig.~\ref{fig:crop} we demonstrate that even using a sub-window of the second image, one is able to recover the motion correctly and undistort the full image. This suggests that a standard smartphone configuration with a wide-angle and a zoom camera can be handled.

\begin{figure}[t]
    \centering
    \renewcommand{\tabcolsep}{0.5pt}
    \renewcommand{\arraystretch}{0.5}
    \begin{tabular}{ccc}
        RS input 1 & RS input 2 & Features from $\omega t$  \\
        \includegraphics[width=0.32\columnwidth]{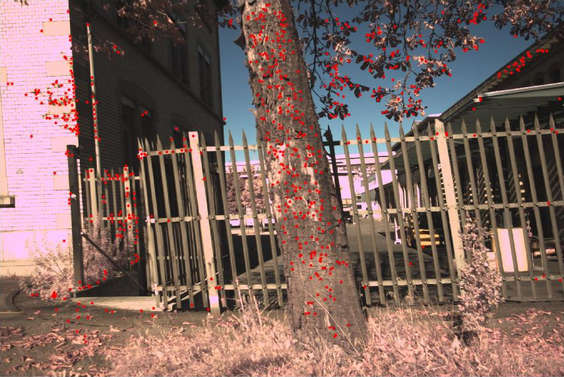} & 
        \includegraphics[width=0.32\columnwidth]{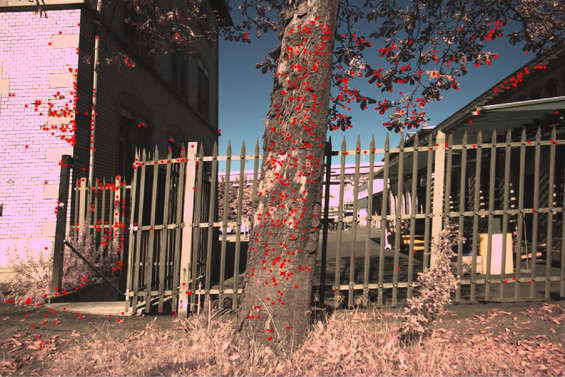} & 
        \includegraphics[width=0.32\columnwidth]{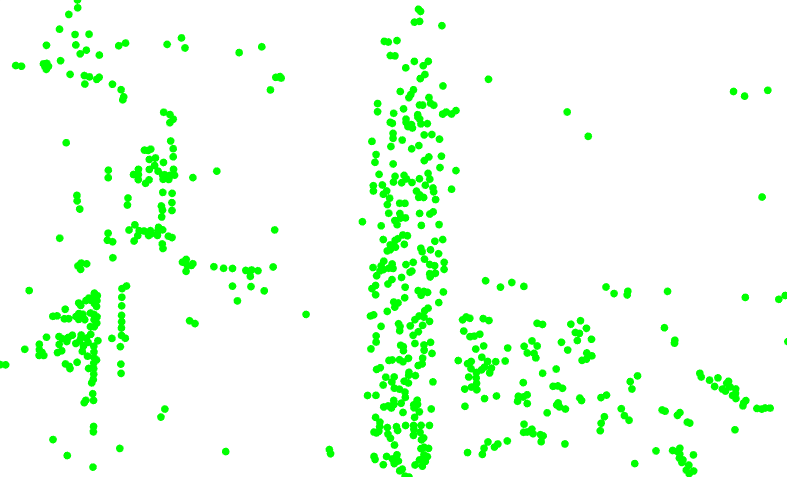}
    \end{tabular}
    \includegraphics[width=\columnwidth]{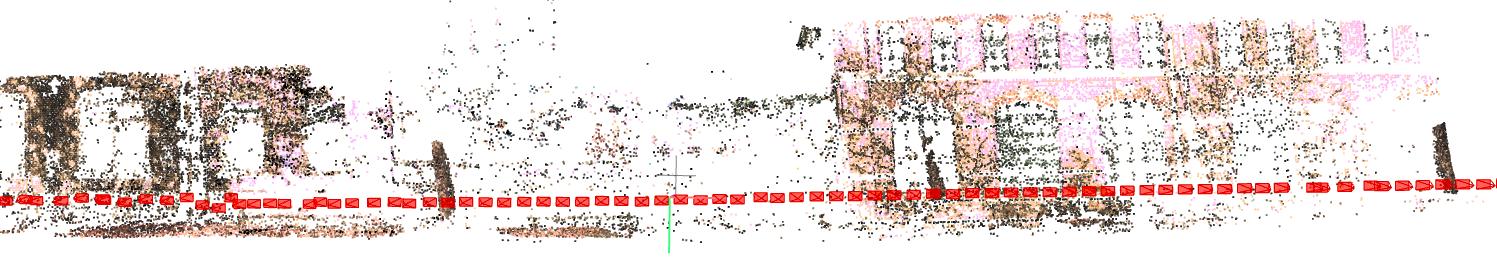}
    \includegraphics[width=\columnwidth]{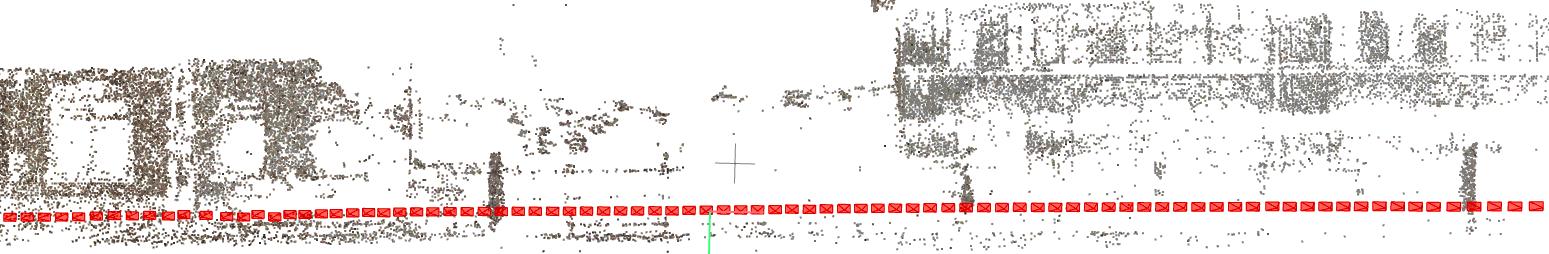}
    \includegraphics[width=\columnwidth]{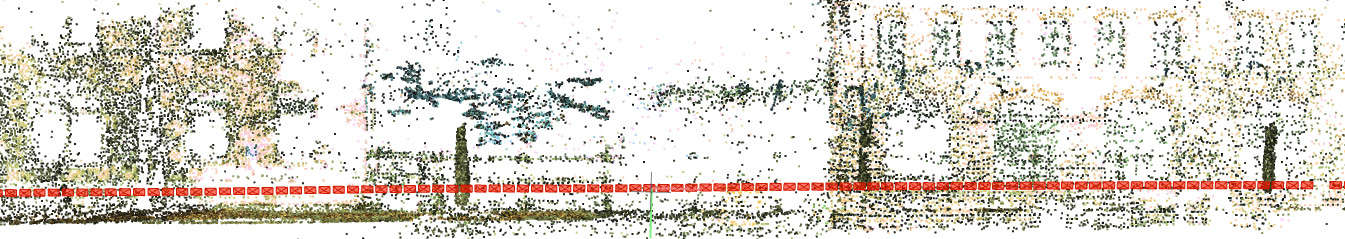}
    \caption{GS-equivalent sparse features from RS images. The corrected features can be further used, e.g, feeding them to an SfM pipeline yields a better reconstruction (middle) than feeding in the raw RS features (top). At the bottom is the reconstruction from a real GS camera.\vspace{1em}}
    \label{fig:trans_sparse}
\end{figure}

\begin{figure}[t]
    \centering
    \includegraphics[width=0.32\columnwidth]{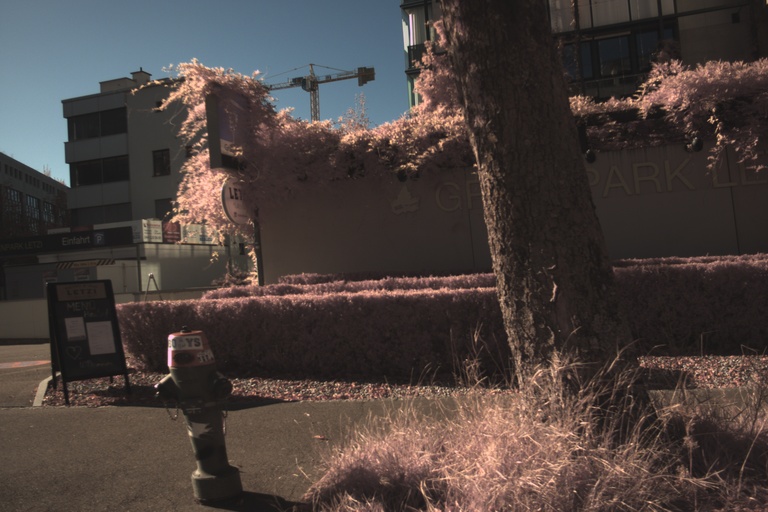}
    \includegraphics[width=0.32\columnwidth,angle=-180,origin=c]{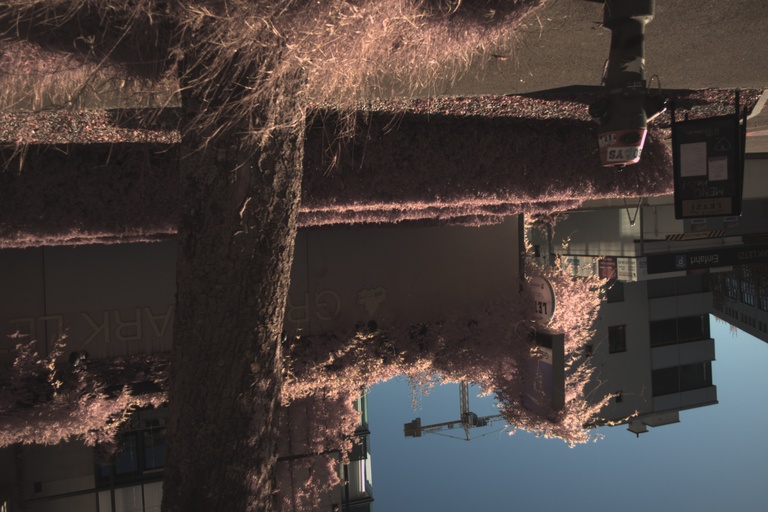}
    \includegraphics[width=0.32\columnwidth]{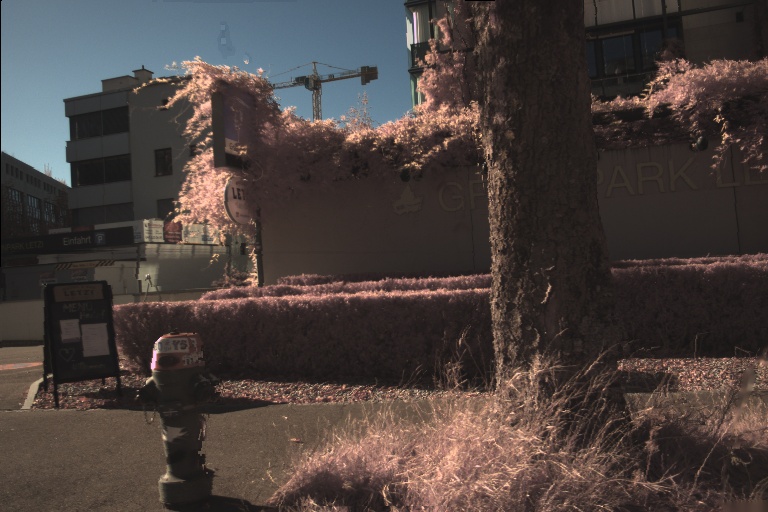} \\
    \includegraphics[width=0.32\columnwidth]{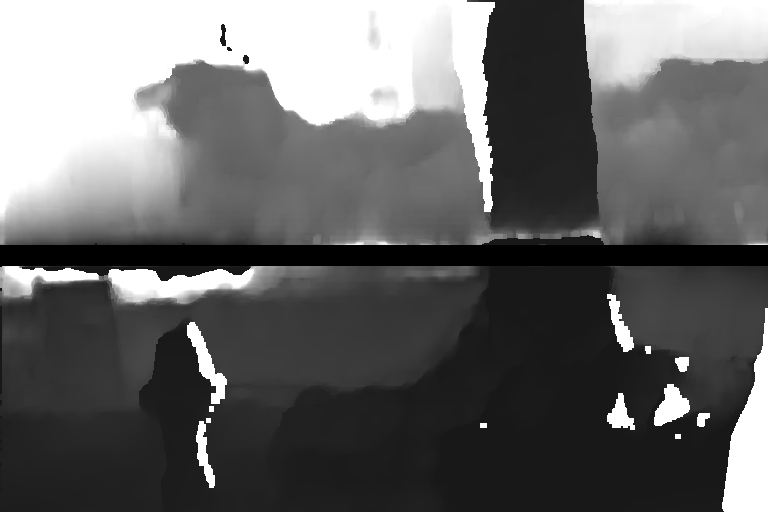}
    \includegraphics[width=0.32\columnwidth]{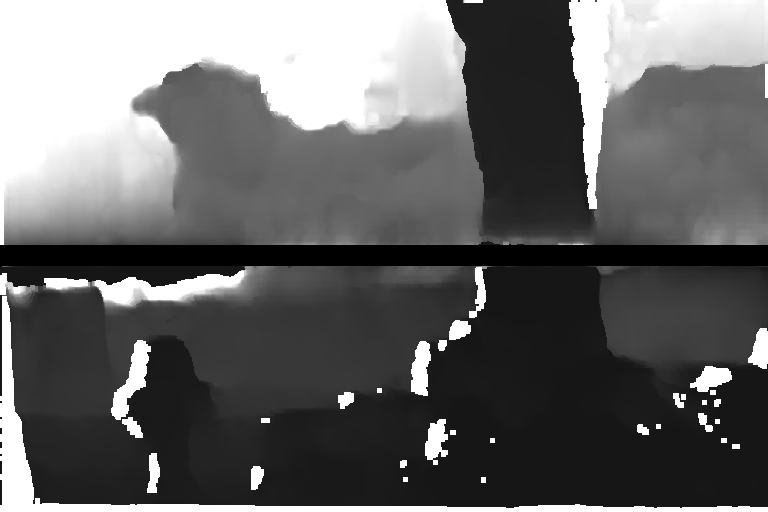}
    \includegraphics[width=0.32\columnwidth]{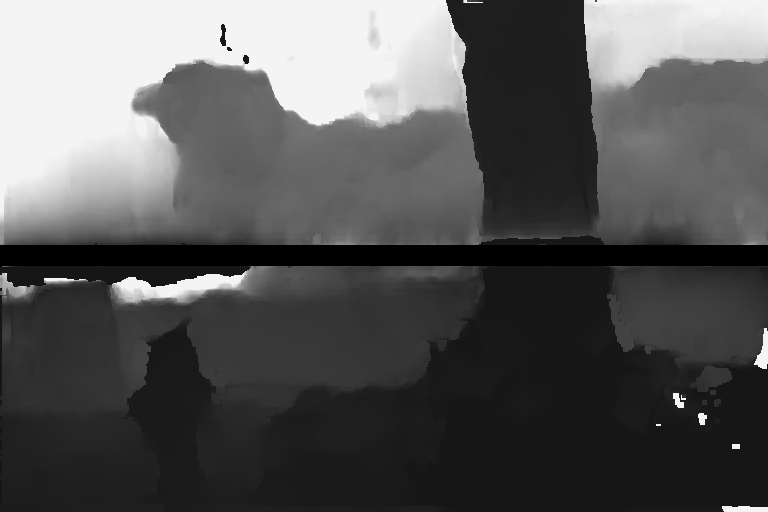}
            \vspace{-0.25em}
    \caption{Depth maps. The top row shows the two input images (left) and the resulting undistorted image (right). The bottom row shows the depth maps created from both input images (left) and the final fused depth map (right).}
    \label{fig:depth}
\end{figure}

For translational motion, undistorting the entire image is in general a hard problem~\cite{vasu_occlusion-aware_2018}, as it requires pixel-wise correspondences as well as occlusion handling, c.f.\ Sec.~\ref{sec:undist} and Appendix~\ref{sec:a_undist_tr}. The results in Fig.~\ref{fig:real_images_tr} show that with our method translation distortion, including occlusions, can be compensated as long as good pixel-wise correspondences can be obtained.
On the contrary, \cite{rengarajan_bows_2016} struggled to compensate the building (row 2) and the tree (row 3) ; \cite{zhuang_rolling-shutter-aware_2017} works in some situations (row 1) and fails in others (row 2 and 3); \cite{vasu_occlusion-aware_2018} often provides good results (row 1 and 3), but sometimes compensates the motion only partially, and also exhibits post-processing artefacts (row 2). We also tried the recent method~\cite{Zhuang_2019_CVPR}, but the authors do not provide the code or the trained model, and our re-implementation trained on 6DOF data provided worse results than all other methods in all cases, so we do not show them.
Note that \cite{zhuang_rolling-shutter-aware_2017,vasu_occlusion-aware_2018} require two, respectively three consecutive frames for which the exact (relative) trigger times as well as the exact shutter read-out speed must be known.

We show further outputs that can be obtained with our method, besides undistorted images. One interesting possibility is to output undistorted (sparse) feature points, which can be fed into an unmodified SfM pipeline. Figure~\ref{fig:trans_sparse} shows an example of SIFT features extracted from RS images, recorded from a car in motion and corrected with the model $\boldsymbol\omega\mathbf{t}$.
Feature point on both the background and the tree were successfully included as inliers and undistorted.
Figure~\ref{fig:trans_sparse} (top row) shows the result of sparse SfM reconstruction with COLMAP~\cite{schonberger_structure--motion_2016}.
One can clearly see the difference between using corrected or uncorrected feature points, especially on the trees in the foreground.

As an intermediate product during translation correction, we obtain depth maps, see Fig.~\ref{fig:depth}. While the camera rig undergoes translation, we obtain depth with lower latency than from consecutive frames, since we are using the inter-row baseline rather than the inter-frame baseline.
The price to pay is that our stereo baseline diminishes towards the central image rows, such that depth estimates become less accurate (and are impossible for the exact middle row).

\vspace{0.25em}
\noindent \textbf{Choice of solver.}
The $\omega$ solver is well suited for slow-moving (e.g., hand-held) devices where translation is insignificant, and for distant scenes with shallow depth range (e.g., drones).
The $t_{xy}$ solver suits scenarios with pure, fast translation (e.g., side-looking cameras on straight roads or rails), and can in the presence of small rotations still be used, trading some accuracy for higher robustness.
For fast, general motion we found the 6DOF solver to perform significantly better, as expected. \\
Another possible scenario which we unfortunately did not manage to test is the case where the camera rig stands still and there are objects moving in the scene, as in surveillance cameras. There, especially the $t_{xy}$ solver could provide a fast and efficient RS correction and depth estimation. 

\section{Conclusion}
\noindent We present a novel configuration of two RS cameras that is easy to realise in practice, and makes it possible to accurately remove significant RS distortions from the images: by mounting two cameras close to each other and letting the shutters roll in opposite directions, one obtains different distortion patterns. Based on those, we derive algorithms to compute the motion of the two-camera rig and undistort the images. Using the corrected geometry, we can perform SfM and compute depth maps equivalent to a GS camera. Our derivations show that similar constructions are in principle also possible when there is a significant baseline between the cameras. Hence, conventional stereo rigs, for instance on robots or vehicles, could in the future also benefit from opposite shutter directions.

\renewcommand{\thesection}{\Alph{section}}
\setcounter{section}{0}

\section{Detailed derivations}
\label{sec:a_details}
\noindent This section contains a more detailed derivation of the results presented in section \ref{sec:translation}.
\subsection{Translation in all axes}
\label{sec:a_txyz}
\noindent In this subsection, we present detailed derivation of the system of equations~\ref{eq:rs_twocam_linsys_txyz}, which describes the case of a general translational motion with constant velocity. In this case we have the translational velocity described by a vector $\V{t} = \mat{ccc}{t_x & t_y & t_z}^\top$ and there is no rotational velocity, \ie $\M{R}_\omega(\alpha) = \M{I}$. Substituting these in equation~\ref{eq:rs_proj_2cam_simplest} we obtain
\begin{equation}
    \begin{array}{rcl}
       \lambda_i\V{u}_{i}  & = & \left[\M{I} \mid \mat{c}{t_x \\ t_y \\ t_z}v_i\right]\V{X}_i \\ \lambda_i^\prime\V{u}^\prime_{i} & = & \left[\M{R}_r \mid \M{R}_r\mat{c}{t_x \\ t_y \\ t_z}v^\prime_{i}\right]\V{X}_i 
%       \M{R}_r^\top\lambda_i^\prime\V{u}^\prime_{i} & = & \left[\M{I} \mid \mat{c}{t_x \\ t_y \\ t_z}v^\prime_{i}\right]\V{X}_i
    \end{array}\;.
     \label{eq:rs_proj_2cam_txyz}
\end{equation}
%We are interested in finding the projections that correspond to a global shutter camera. 
We are interested in a global shutter image that would be captured by a GS camera in a place of our camera pair, i.e., the first equation of~\eqref{eq:rs_proj_2cam_txyz} minus the translational motion
\begin{equation}
    \begin{array}{rcl}
       \lambda_i\V{u}_{gi}  & = & \left[\M{I} \mid \V{0} \right]\V{X}_i.
    \end{array}
    \label{eq:gs_ekvivalent}
\end{equation}
By multiplying the second matrix equation in~\eqref{eq:rs_proj_2cam_txyz} from the left with $\M{R}_r^{\top} = \mbox{diag}([-1,\,-1,\,1,])$ and by subtracting this equation from the first one, we obtain equations~\ref{eq:rs_twocam_linsys_txyz} in section~\ref{sec:translation}.

\subsection{Translation in the $xy$-plane}
\label{sec:a_txy}
\noindent Considering only the translation in the $xy$-plane, we have a translational velocity vector $\V{t} = \mat{ccc}{t_x & t_y & 0}^\top$ and equations~\ref{eq:rs_twocam_linsys_txyz} become
\begin{equation}
    \mat{c}{t_x \\ t_y \\ 0}v_i-\mat{c}{t_x \\ t_y \\ 0}v^\prime_{i} = \mat{c}{u_{i} \\ v_i \\ 1}\lambda_i-\mat{c}{-u^\prime_{i} \\ -v^\prime_{i} \\ 1}\lambda^\prime_i\;,
    \label{eq:rs_twocam_linsys_tx_ty}
\end{equation}
By subtracting the equation corresponding to the second row in~\eqref{eq:rs_twocam_linsys_tx_ty} from the equation corresponding to the first we obtain a system of two equations in three unknowns $t_x$, $t_y$ and $\lambda_i$
\begin{equation}
    \mat{ccc}{
    v_i - v^\prime_{i} & 0 &  -u_{i}-u^\prime_{i} \\
    0 & v_i - v^\prime_{i} &  -v_i-v^\prime_{i}
    }
    \mat{c}{t_x\\t_y\\ \lambda_i}
    =
    \V{0}\;.
    \label{eq:tx_ty_linsys}
\end{equation}
For a single correspondence this is a homogeneous system of linear equations of rank two (unless $v_i=v^\prime_{i}$). Solving~\ref{eq:tx_ty_linsys} brings us to equation~\ref{eq:rs_twocam_linsys_tx_ty} in section~\ref{sec:translation}.
\subsection{Translation in the $x$-axis}
\label{sec:a_tx}
\noindent For the motion only along the camera $x$-axis we have $\V{t} = \mat{ccc}{t_x & t_y & 0}^\top$ and equations~\ref{eq:rs_twocam_linsys_txyz} become
\begin{equation}
    \mat{c}{t_x \\ 0 \\ 0}v_i-\mat{c}{t_x \\ 0 \\ 0}v^\prime_{i} = \mat{c}{u_{i} \\ v_i \\ 1}\lambda_i-\mat{c}{-u^\prime_{i} \\ -v^\prime_{i} \\ 1}\lambda^\prime_i.
    \label{eq:rs_twocam_linsys_tx}
\end{equation}
From the third row we immediately see that $\lambda_i=\lambda^\prime_i$, and thus the second row yields $v_i=-v^\prime_{i}$. From the first row we see that 
\begin{equation}
    \begin{array}{rcl}
    t_x v_i - t_x v^\prime_{i} & = & (u_{i}+u^\prime_{i})\lambda_i \\
    t_x v_i + t_x v_i & = & (u_{i}+u^\prime_{i})\lambda_i \\
    t_x & = & \frac{u_{i}+u^\prime_{i}}{2v_i}\lambda_i,
    \end{array}
    \label{eq:tx_lambda}
\end{equation}
which gives us a direct relation between the perspective depth $\lambda_i$ of the 3D point $\V{X}_i$ and the translational velocity $t_x$. We are, again, interested in a global shutter image that would be captured by a GS camera in a place of our camera pair, i.e., the first equation of~\ref{eq:rs_proj_2cam_txyz} minus the translational motion and we obtain
\begin{equation}
    \lambda_i\V{u}_{gi} = \lambda_i\V{u}_{i}-\V{t}v_i = \lambda_i\V{u}_{i}-\mat{c}{t_x \\ 0 \\ 0}v_i\;.
\end{equation}
Substituting $t_x$ from equation~\eqref{eq:tx_lambda} leads to
\begin{equation}
    \begin{array}{rcl}
    \lambda_i\V{u}_{gi} & = &\lambda_i\V{u}_{i}-\mat{c}{\frac{u_{i}+u^\prime_{i}}{2v_i}\lambda_i \\ 0 \\ 0}v_i \\
    \V{u}_{gi} & = & \mat{c}{\frac{u_{i}+u^\prime_{i}}{2}\\v_i\\1}
    \end{array}\;,
\end{equation}
which is the result presented in section~\ref{sec:translation}, suggesting that we can simply interpolate the $x$-coordinate of the corresponding projections to obtain a GS equivalent one. 
\section{6DOF solver with known baseline}
\label{sec:a_baseline_solver}
\noindent As promised, we present an extension of the solution in section~\ref{sec:6dof_problem} which solves the case when there is a known, fixed baseline between the cameras. Although we did not use this solution for the results in the real experiments, since the baseline between the cameras was negligible compared to the scene distance, it could be useful for systems with larger baselines such as stereo setups on robots, cars, etc. We can augment the camera matrices that appear in equations~\ref{eq:rs_proj_2cam_simplest} in section~\ref{sec:problem_formulation} by a known baseline $\V{b}$ as
\begin{equation}
    \begin{array}{rcl}
        \M{P}(v_{i})&=&\left[\M{R}_\omega(v_{i}) \mid \V{t}\,v_{i}\right] \\
        \M{P}^\prime(v^\prime_{i})&=&\left[\M{R}_r\M{R}_\omega(v^\prime_{i}) \mid v^\prime_{i}\,\M{R}_r\V{t} + \M{R}_r\V{b}\right]
    \end{array}
    \label{eq:baseline}
\end{equation}
and, analogously to the solution in section~\ref{sec:6dof_problem}, transform them so the first camera is identified with the world coordinate system, obtaining
\begin{equation}
    \begin{array}{rcl}
        \M{P}&=&\left[\M{I} \mid \V{0}\right] \\
        \M{P}^\prime\left(v_{i},v^\prime_{i}\right) &=&[\M{R}(v_i,v^\prime_{i}) 
        \mid v^\prime_{i}\M{R}_r\V{t} + \M{R}_r\V{b} - v_{i}\M{R}(v_{i},v^\prime_{i})\V{t}],
   \end{array}
    \label{eq:baseline2}
\end{equation}
where $\M{R}(v_{i},v^\prime_{i})=\M{R}_r\M{R}_\omega(v^\prime_{i})\M{R}_\omega(v_{i})^\top$. The essential matrix in equation~~\ref{eq:rs_2cam_essential} will now become
\begin{equation}
    \M{E}(v_i,v_i^\prime) = \left[\V{t}(v_i,v_i^\prime)\right]_\times \M{R}(v_{i},v^\prime_{i})\;,
    \label{eq:baseline_E}
\end{equation}
where $\V{t}(v_i,v^\prime_i) = v^\prime_{i}\M{R}_r\V{t} + \M{R}_r\V{b}  - v_{i}\M{R}(v_{i},v^\prime_{i})\V{t}$. Note that we are now also solving for the scale of $\V{t}$ with respect to the baseline, cannot use the parameterization $\V{t} = \mat{ccc}{1-x & x & y}$ and therefore need six correspondences to solve for the six unknowns. 

Similarly to the no-baseline case, we use the hidden variable trick. Hiding the translation $\V{t}$, we get equations in only the rotation parameters. Applying the generator from~\cite{larsson_efficient_2017} we generate a solver for this system as well. The solver performs Gaussian elimination on a $15 \times 15$ matrix and solves a $20 \times 20$ eigenvalue problem.

\begin{figure}
    \centering
    \includegraphics[width=0.9\columnwidth]{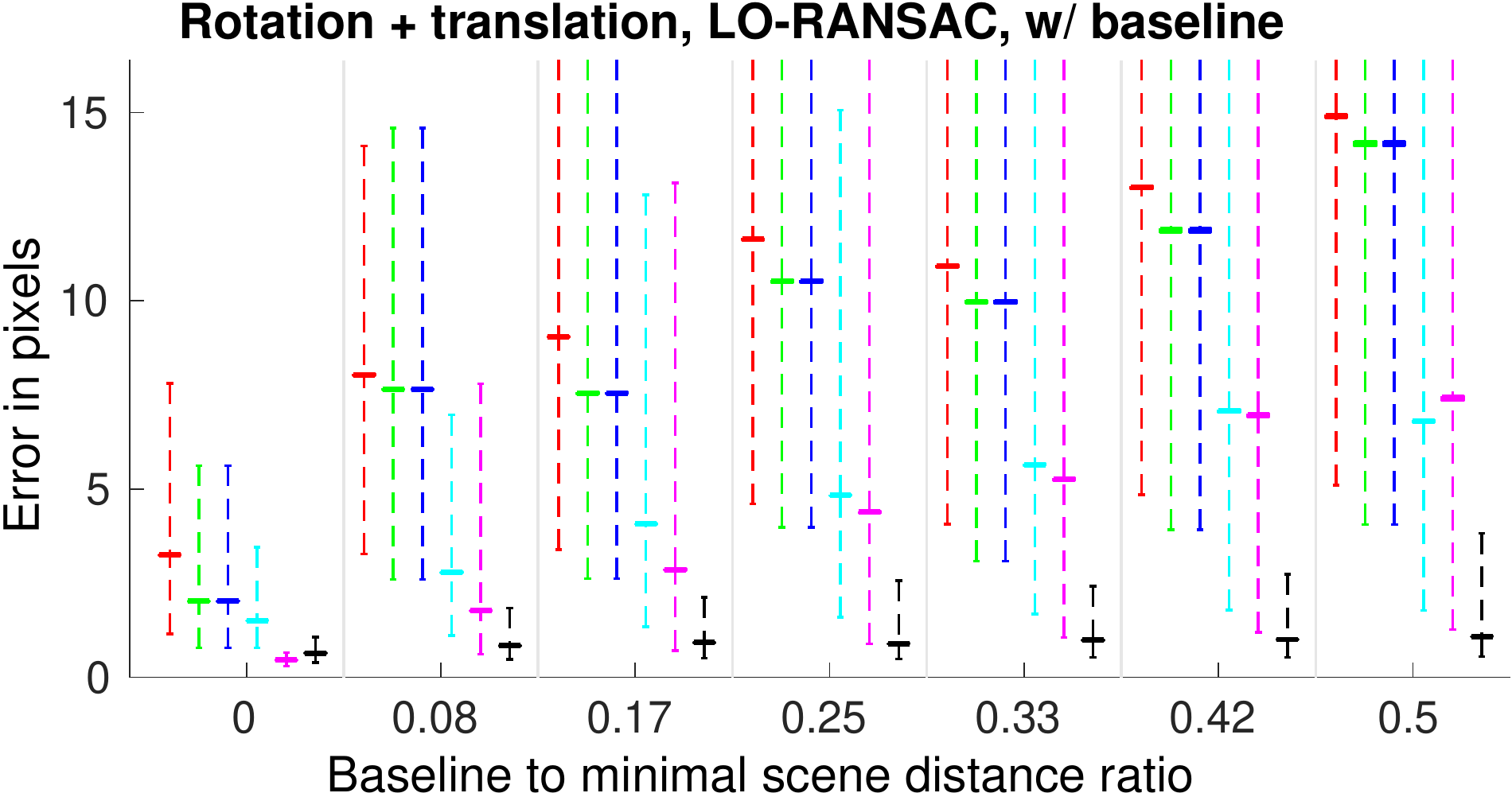}
    \caption{The solvers that do not consider a baseline ({\color{red} interpolation}, {\color{green} txy}, {\color{blue} txyz}, {\color{cyan} $\omega$}, {\color{magenta} $\omega$t}) gracefuly degrade in performance as the actual baseline in the data increases, whereas the 6DOF solver that considers a baseline (black) provides stable performance in terms of the pixel error of the undistorted correspondences with respect to the global shutter equivalent correspondences.}
    \label{fig:increasing_baseline}
\end{figure}

In Fig.~\ref{fig:increasing_baseline} we compare the solvers presented in section~\ref{sec:solutions} on data where cameras have an increasing baseline to the solution with known baseline provided here. One can see that the performance of the baseline solver provides stable performance over the increasing baseline. 

\section{Combining the undistorted images}
\label{sec:a_combine}
\begin{figure}
    \centering
    \includegraphics[width=0.47\columnwidth]{figs/re_rot11_1.jpg}
    \includegraphics[width=0.47\columnwidth,angle=-180,origin=c]{figs/re_rot11_2.jpg} \\
    \includegraphics[width=0.47\columnwidth]{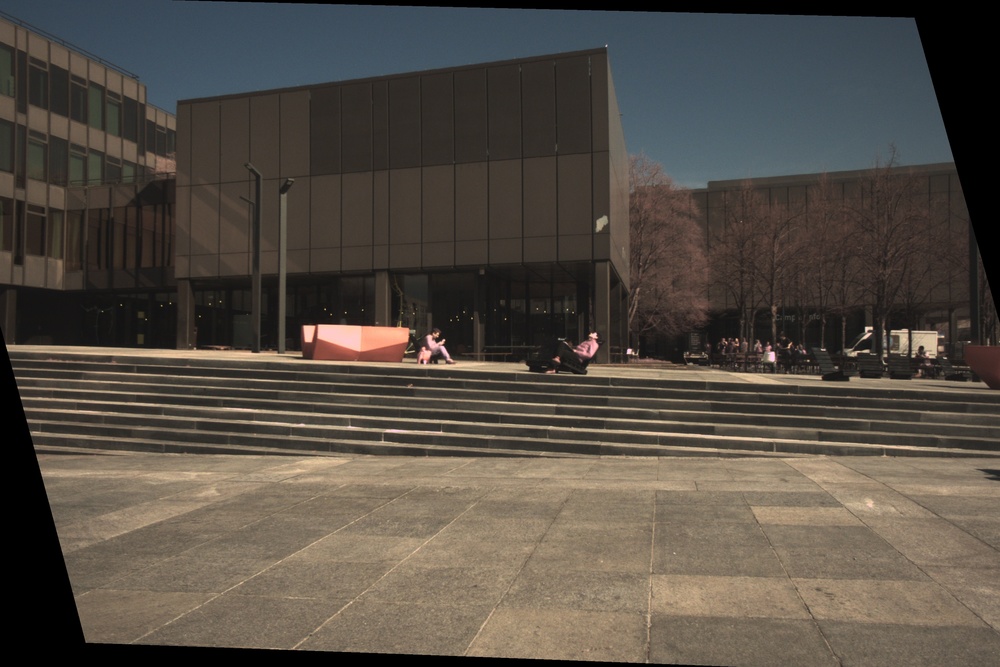}
    \includegraphics[width=0.47\columnwidth]{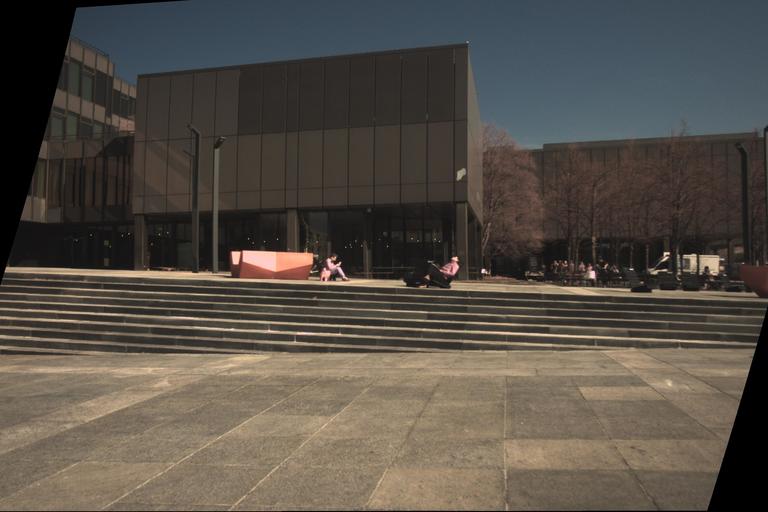}\\
    \includegraphics[width=0.952\columnwidth]{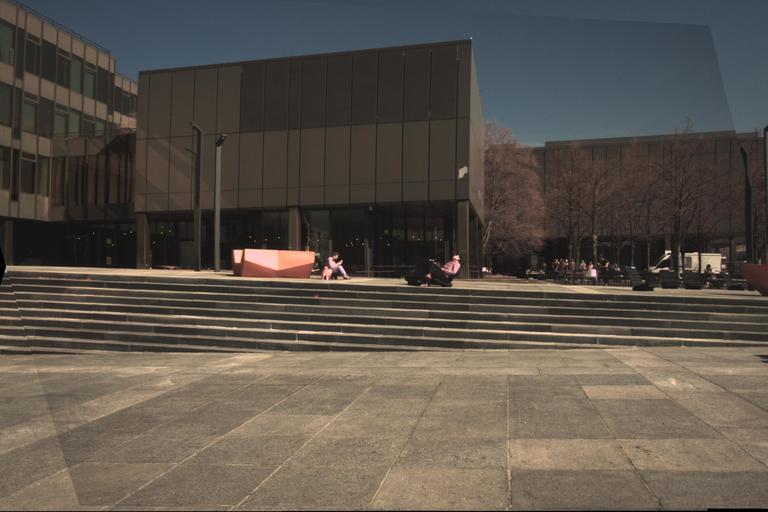}\\
    \includegraphics[width=0.47\columnwidth]{figs/re_rot7_1.jpg}
    \includegraphics[width=0.47\columnwidth,angle=-180,origin=c]{figs/re_rot7_2.jpg} \\
    \includegraphics[width=0.47\columnwidth]{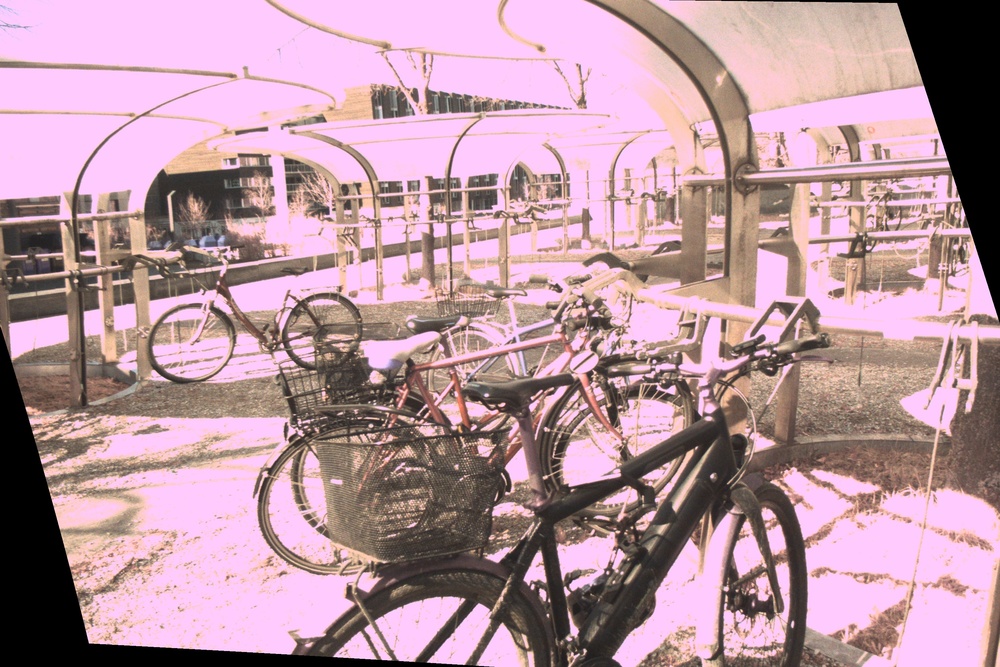}
    \includegraphics[width=0.47\columnwidth]{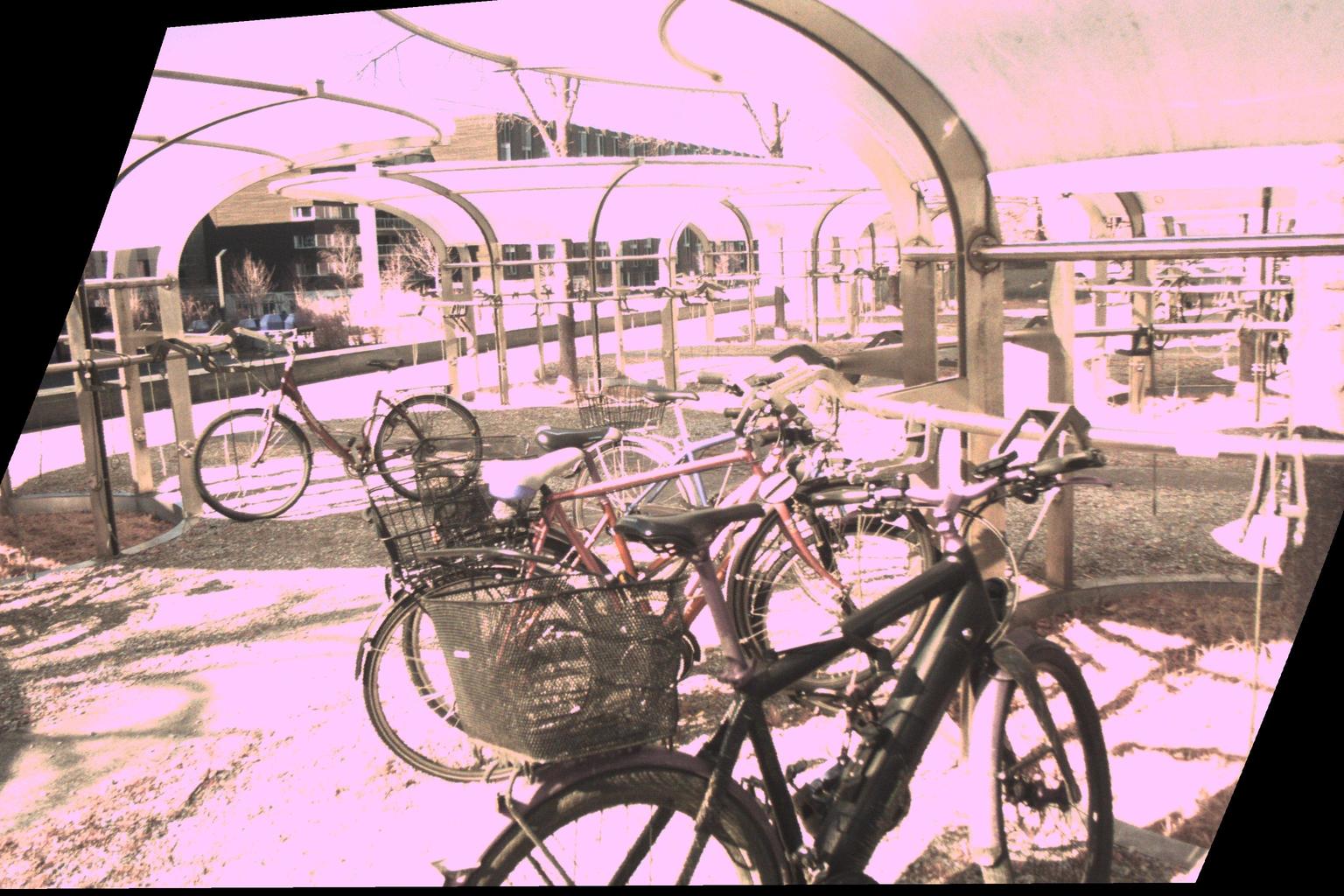}\\
    \includegraphics[width=0.952\columnwidth]{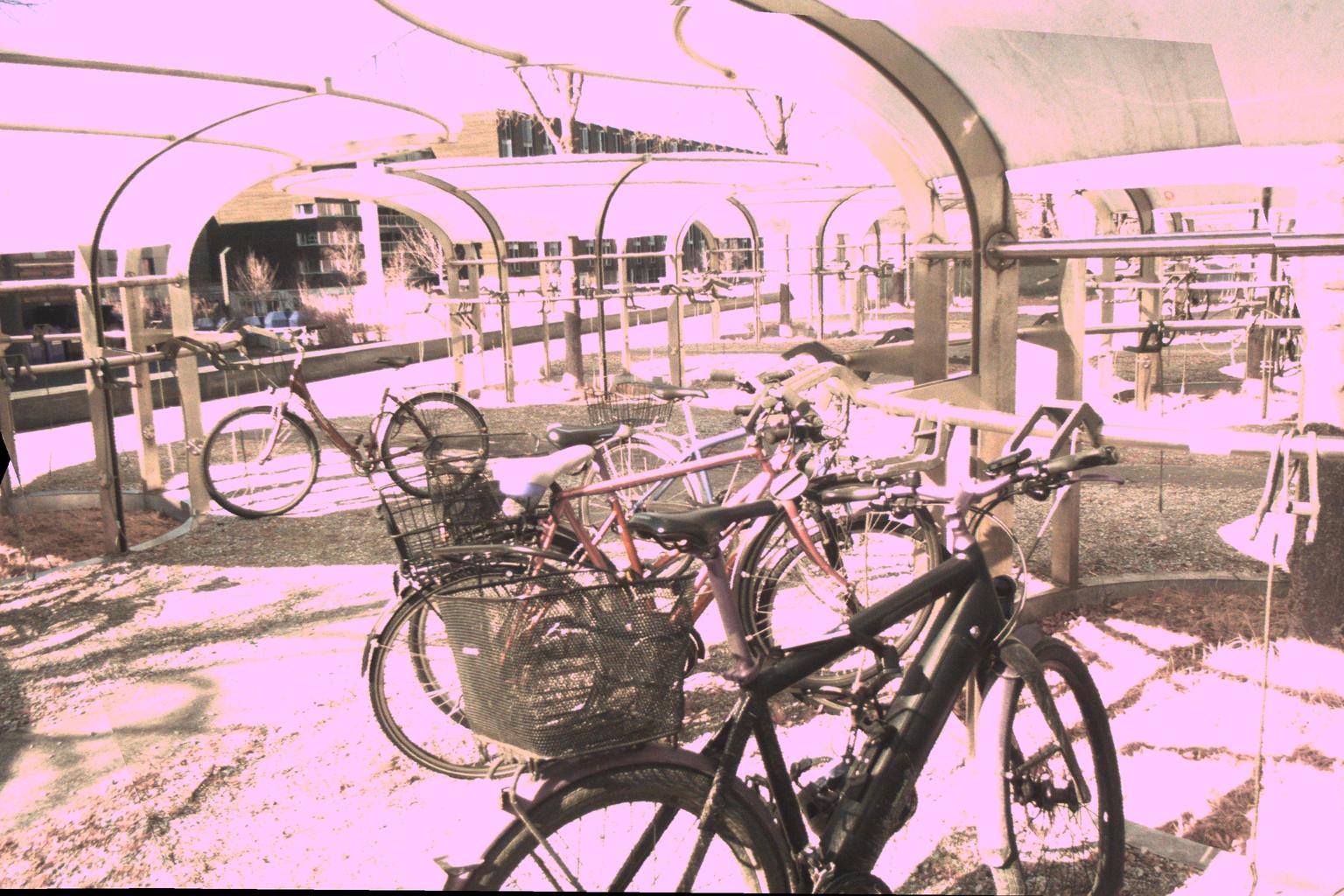}
    \caption{Combining the two undistorted outputs. Displayed are the input images (top), warped results (middle) and the combined result (bottom).}
    \label{fig:undist_combined}
\end{figure}

\noindent As described in section~\ref{sec:undist}, when performing the undistortion of images distorted by rotational motion we actually obtain two undistorted images by warping each of the inputs to the virtual global shutter image plane. Each input image covers a part of a scene content that is not visible in the other image. Therefore, it makes sense to combine both undistorted images to obtain a more complete image. In Fig.~\ref{fig:undist_combined} you can see several examples of the distorted inputs, the undistorted images and the combined output. Note that the images were only warped to the same image plane using the rotational velocity computed by $\omega$ solver and optimized using non-linear least squares refinement step minimizing equation~\ref{eq:refinement_w} in section~\ref{sec:undist}, no other image processing was performed. One can notice small discontinuities at the boundaries, which could be improved by additional image processing software suited for image blending. 

\section{Correcting images distorted by translation}
\label{sec:a_undist_tr}
\vspace{0.5em}
\noindent \textbf{Pixel-wise correspondences}

\begin{figure}
    \centering
    \includegraphics[width=\columnwidth]{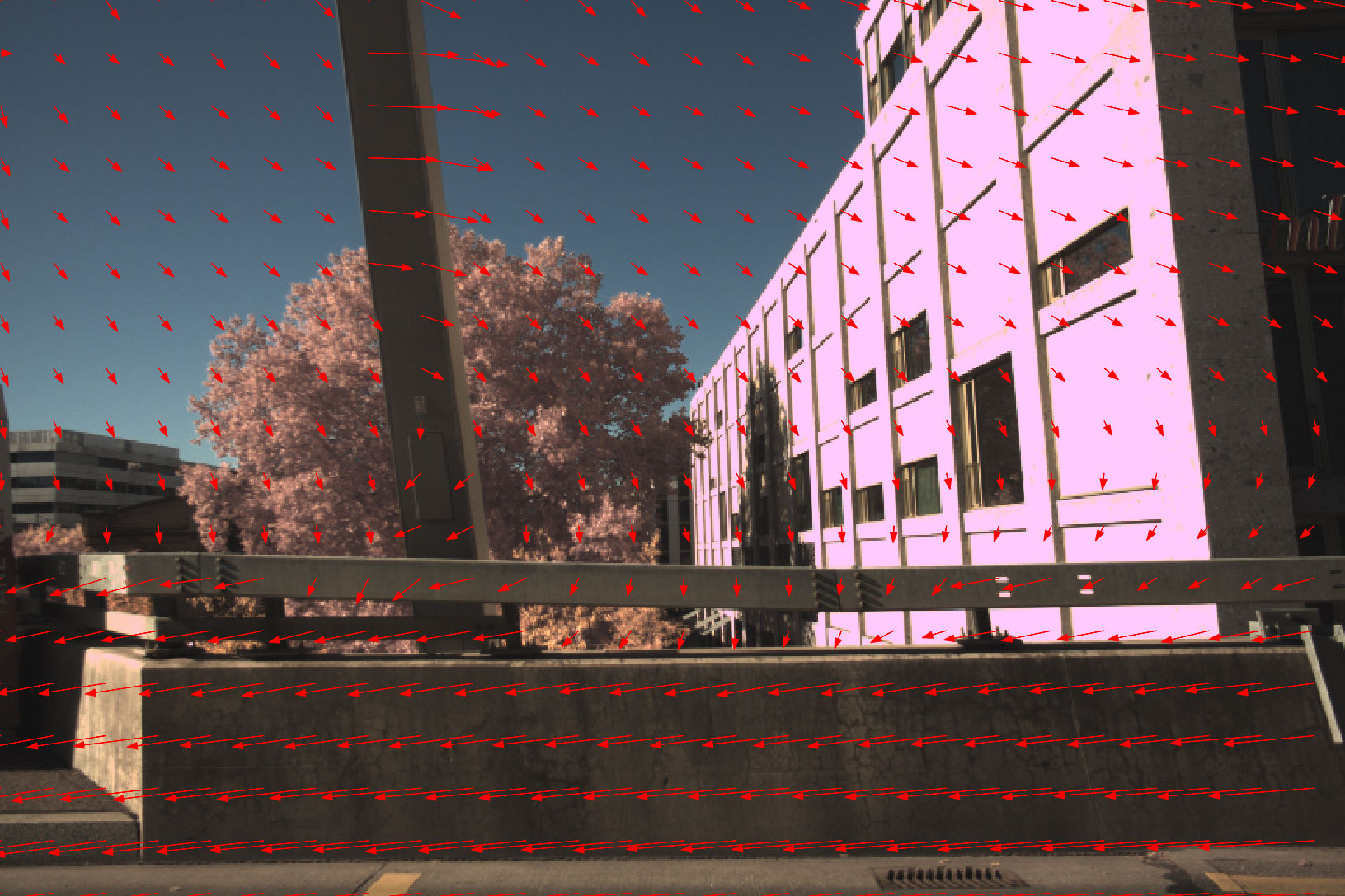}
    \caption{Flow estimated using PWC-net~\cite{sun_pwc-net:_2017}.}
    \label{fig:flow}
\end{figure}

First, the pixel-wise correspondences must be found for both images. We found PWC-net~\cite{sun_pwc-net:_2017} to work the best for these purposes. We compare the flow from the first image to the second as well as the flow from the second image to the first and filter out the flow vectors that are not consistent, see Fig.\ref{fig:flow}. Since we know that the flow is caused by the difference in time of capture, that is zero for the middle row and increases towards the top and bottom of the image we can also filter the flow further by introducing a threshold on the maximum flow based on the distance from the middle row of the image.

\vspace{0.5em}
\noindent \textbf{Motion estimation}

Second step is computing the 6DOF motion parameters. Although, in principle we could use just the txy or even tx solver to solve for the translation if we knew the motion was purely translational we did not resort to such constraint, since our data was coming from a handheld camera travelling in a car and rotation could be present as well. 

We have two choices of correspondences to use - sparse matches coming from, e.g., SIFT features or dense matches coming from the optical flow. Although sparse correspondences are usually more reliable, because they come from distinct image features and are usually sub-pixel precise, we found that they are in some cases not sufficient for correct motion estimation, since the parts of the image that determine the motion - e.g.\ the pole in Fig.~\ref{fig:flow} might contain only little texture and provide only few features. The correspondences coming from the flow are therefore a better candidate for a robust solution.

\vspace{0.5em}
\noindent \textbf{Choosing good correspondences for RANSAC}

RANSAC is necessary for filtering mismatches. We found that to estimate correct motion parameters, the pixel-wise correspondences obtained from optical flow should not be used without proper pre-selection. To determine the motion, some correspondences carry more information than others. E.g.\ in our case, the correspondences around the middle row (where the temporal displacement is small) carry only little information and have low signal to noise ratio. 

Another issue is that some types of motion cause very similar distortions to others, e.g.\ translation in x causes the same distortions as rotation around y if the scene is planar. Therefore, our motion estimation is prone to the presence of a dominant plane, similar to the estimation of epipolar geometry~\cite{ChumWM05}. In our scenario, we found that for a general scene it is important to choose a balanced set of correspondences that contain both large displacements on the foreground objects as well as the small displacements that appear towards the top or bottom of the image (around the middle row of the image all displacements will be small), because only using those we can distinguish between translational and rotational motion. 

If, e.g. in the example in Fig.~\ref{fig:flow} we would select a uniform representation of correspondences across the entire image, the correspondences on the slanted pole would be dominated by correspondences from other parts of the image in RANSAC, even though they are very important for determining the correct motion parameters.

\vspace{0.5em}
\noindent \textbf{Depth maps and occlusion masks}

\begin{figure}
    \centering
    \includegraphics[width=0.47\columnwidth]{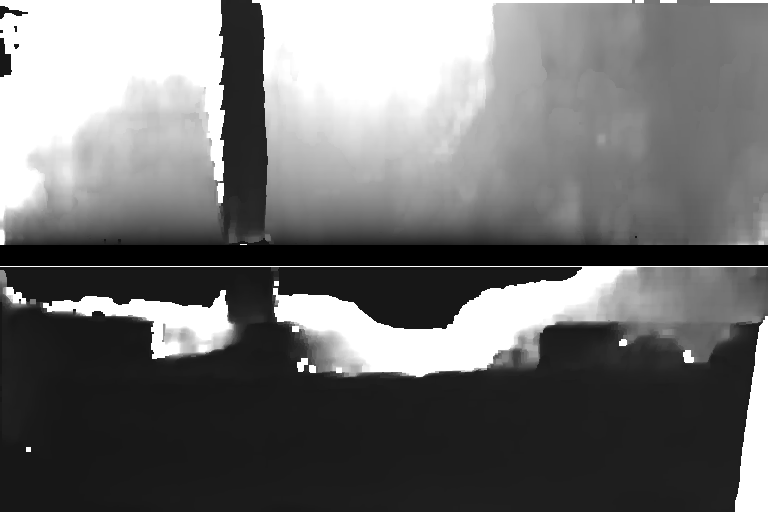}
    \includegraphics[width=0.47\columnwidth]{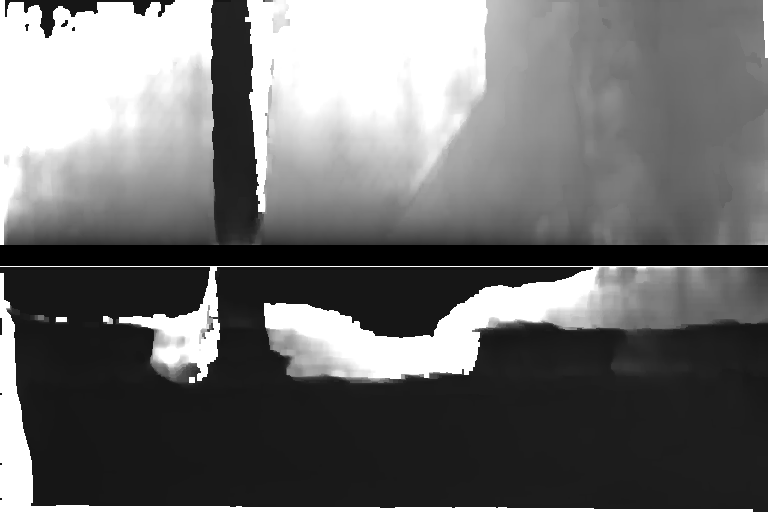}
    \caption{Depth maps projected in a virtual GS image plane, each created using one direction of the flow. Darker means closer. Notice the wrongly estimated flow in the bottom half of the image creates errors in the depth estimation.}
    \label{fig:depthmaps}
\end{figure}

After computing the motion parameters, we can proceed to computing the pixel-wise depth from the flow. We compute two depth maps, one for each flow, see Fig.~\ref{fig:depthmaps}. The depth is computed in the following way. Using the motion parameters $\omega$ and $\V{t}$ we create for each correspondence $\V{u}_i \leftrightarrow \V{u}^\prime_i$ the camera matrices corresponding to their rows, i.e. $\M{P}(v_i)$ and $\M{P}(v^\prime_i)$. We then use these projection matrices and the corresponding image points to triangulate a 3D point $\V{X}_i$, which we project into a virtual GS camera coordinate system by $\M{P}\V{X} = \mat{cc}{\M{I} & \V{0}} = \V{X} = \mat{ccc}{X_1 & X_2 & X_3}^\top$. The depth $X_3$ is then projected at the corresponding location in the image plane, i.e. $\mat{cc}{X_1/X_3 & X_2/X_3}$. A small number of rows around the middle of the image is filled with zeros as we consider the depth there to be too unreliable and we just use interpolation of the input images in this region.

\begin{figure}
    \centering
    \includegraphics[width=0.47\columnwidth]{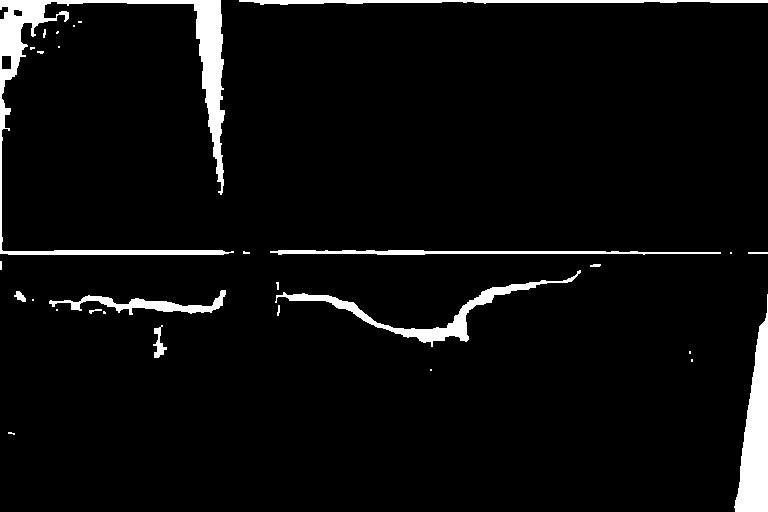}
    \includegraphics[width=0.47\columnwidth]{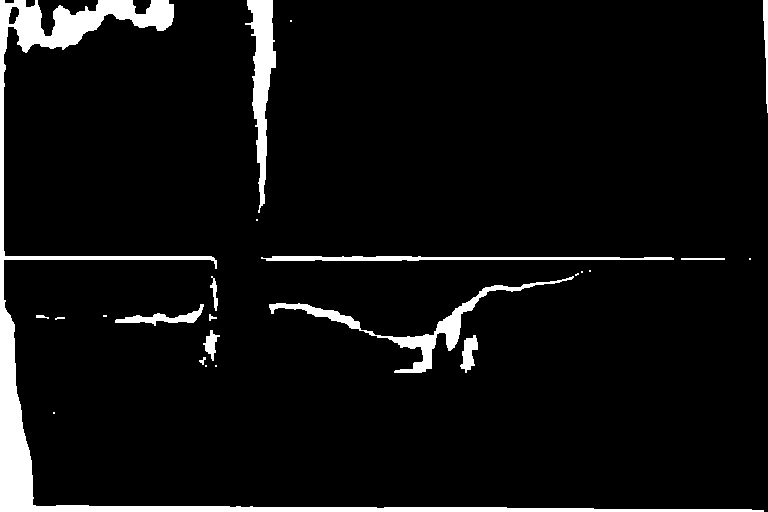}
    \caption{Occlusion masks for both input images computed based on the flow. Based on those we decide from which image we take pixels in areas that are occluded. White areas mean we do not fill the pixels at this location in the final image from the respective input image.}
    \label{fig:occlusion_mask}
\end{figure}

By comparing the depth assigned to each pixel from the first and from the second flow we determine the occlusion mask, which tells us which image has pixel value that should be assigned to this location, see Fig.~\ref{fig:occlusion_mask}. This is important, since the flow is also estimated for pixels that lie in occluded regions and therefore don't actually have a correspondence and those should be filled with values from the image in which the occluding object was not present. If the mask is zero for given location, it means that this location in the final image should not be filled with pixels coming from the corresponding input image. White areas in figure~\ref{fig:occlusion_mask} depict the areas with zeros.

\begin{figure}
    \centering
    \includegraphics[width=\columnwidth]{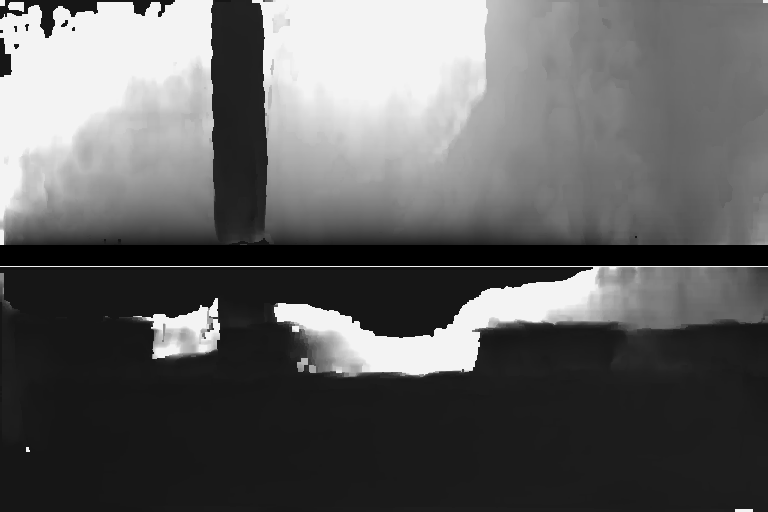}
    \caption{Final depth map fused from the two in Fig.\ref{fig:depthmaps}. Notice that in the lower half of the image, errors in the estimated flow cause errors in the depth estimation. In the middle region the depth is undetermined.}
    \label{fig:depthmapfused}
\end{figure}

Before the final undistortion we fuse the two depth maps, taking the closer values from each, see figure~\ref{fig:depthmapfused}.

\vspace{0.5em}
\noindent \textbf{Creating the final image}

\begin{figure*}[tp]
    \centering
    \includegraphics[width=0.95\columnwidth]{figs/re_tr3_1.jpg}
    \includegraphics[width=0.95\columnwidth]{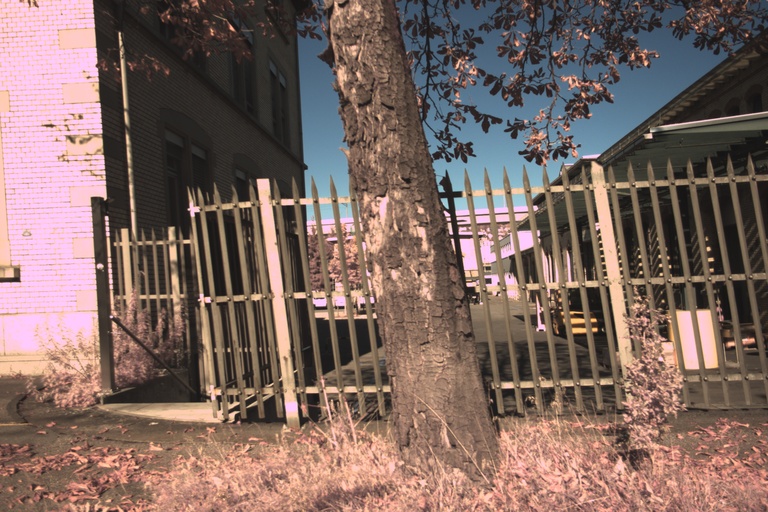} \\
    \includegraphics[width=0.95\columnwidth,angle=-180,origin=c]{figs/re_tr3_2.jpg}
    \includegraphics[width=0.95\columnwidth,angle=-180,origin=c]{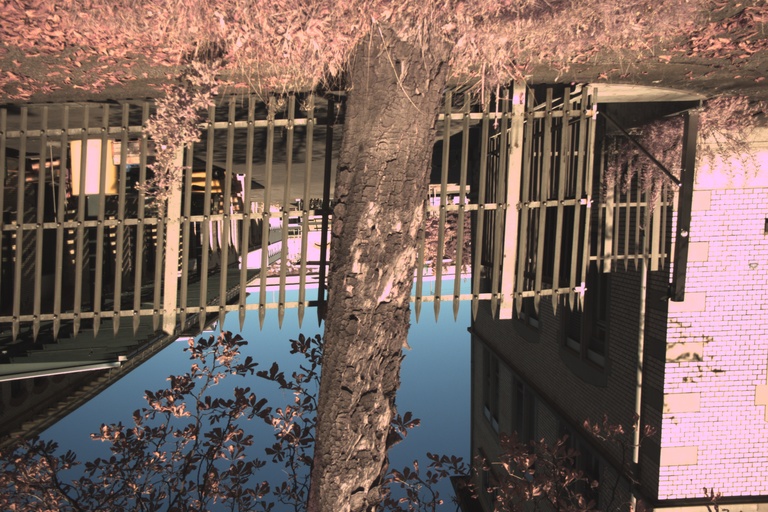}\\
    \includegraphics[width=0.95\columnwidth]{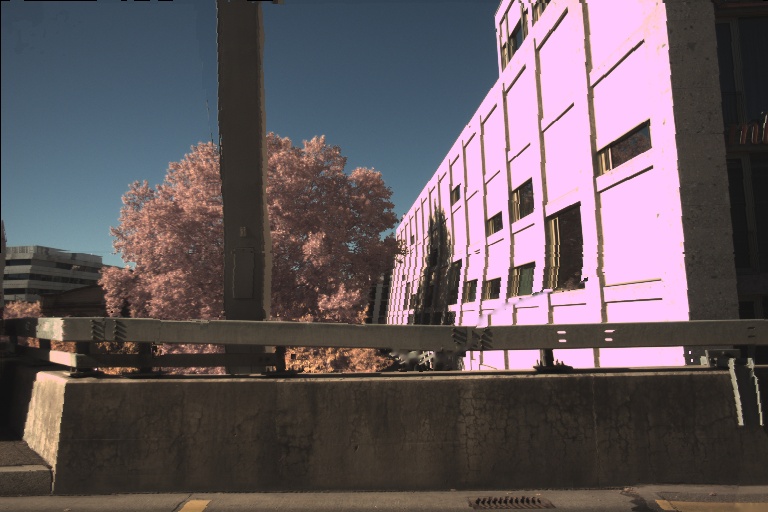}
    \includegraphics[width=0.95\columnwidth]{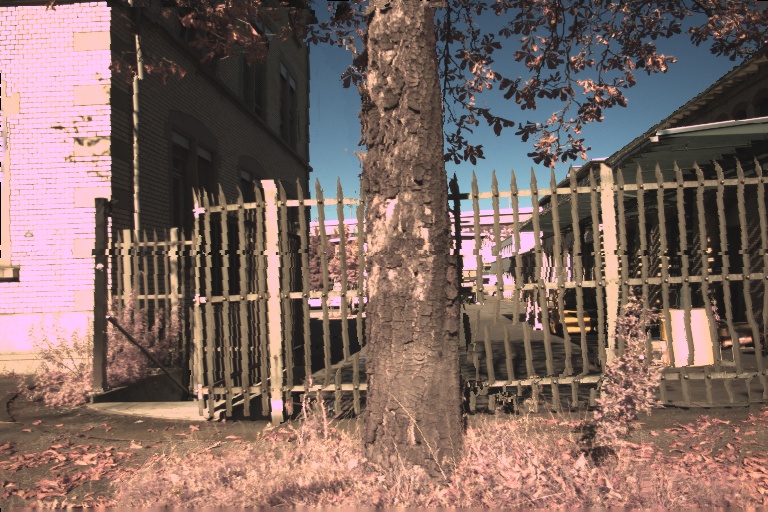} \\
    \includegraphics[width=0.95\columnwidth]{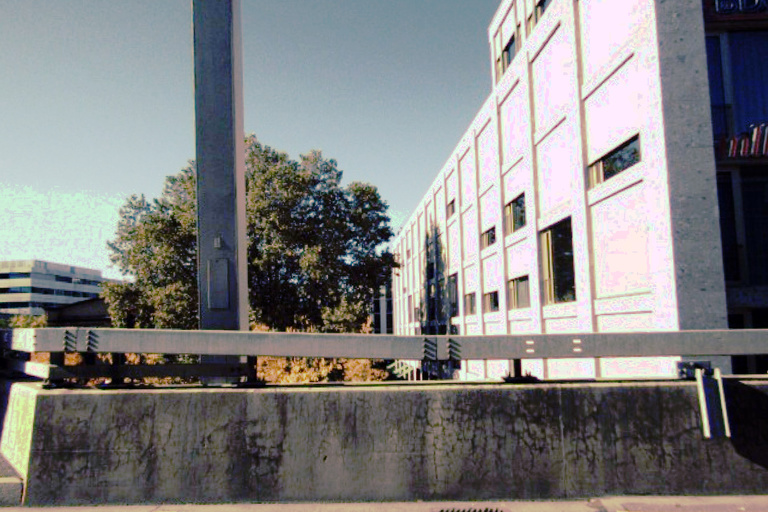}
    \includegraphics[width=0.95\columnwidth]{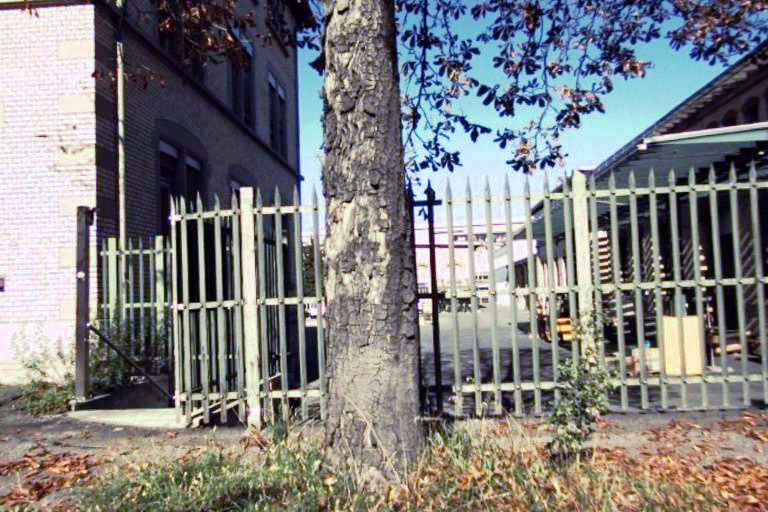}
    \caption{Undistortion examples on two image pairs, one in each column. From top to bottom - the two input images, the resulting undistorted image and the corresponding image from a GS camera.}
    \label{fig:result}
\end{figure*}

Final image is created by traversing the depth map pixel by pixel, recovering the corresponding 3D point $\V{X}$ and finding the coordinates where this point projects into either of the input images based on the occlusion mask. The pixel value is then taken from this location. The resulting image is the left column of Fig.~\ref{fig:result}.

%
% \begin{figure}[tp]
%     \centering
%     \includegraphics[width=0.9\columnwidth]{figs/re_tr6_1.jpg}
%     \includegraphics[width=0.9\columnwidth,angle=-180,origin=c]{figs/re_tr6_2.jpg}
%     \includegraphics[width=0.9\columnwidth]{figs/tr6_res.jpg}
%     \includegraphics[width=0.9\columnwidth]{figs/re_tr6_3.jpg}
%     \caption{Another example of two input images, the resulting undistorted image and the corresponding image from a GS camera.}
%     \label{fig:result2}
% \end{figure}

Note that this approach is very generic and does not assume any properties of the motion, scene or image content and the results occasionally contain artefacts due to errors in the input flow. In the places where good correspondences are provided, our methods are able to correct large RS distortion of very challenging scenes, see Fig.~\ref{fig:result} right column.

The visual quality of the result could be improved by assuming scene properties such as piece-wise planarity~\cite{vasu_occlusion-aware_2018}, segmenting the scene, applying more advanced filtering of the flow and depth maps and further post-processing steps. This was, however, not the purpose of this work. 

\clearpage

{\small
\bibliographystyle{ieee_fullname}
\bibliography{main}
}

\clearpage

\end{document}